\documentclass[pdflatex,sn-standardnature]{sn-jnl}
\jyear{2025}
\usepackage{setspace}
\usepackage{lineno}
\usepackage{graphicx}
\usepackage{multirow}
\usepackage{amsmath,amssymb,amsfonts}
\usepackage{amsthm}
\usepackage{mathrsfs}
\usepackage[title]{appendix}
\usepackage{xcolor}
\usepackage{textcomp}
\usepackage{manyfoot}
\usepackage{booktabs}
\usepackage{algorithm}
\usepackage{algorithmicx}
\usepackage{algpseudocode}
\usepackage{listings}
\usepackage{textcomp}
\usepackage{array}
\usepackage{threeparttable}
\usepackage{float}
\usepackage{caption}
\usepackage{multirow}
\usepackage{makecell}
\usepackage{adjustbox}
\renewcommand{\figurename}{Fig.}

\begin{document}

%% Title
\title[Article Title]{Dynamical errors in machine learning forecasts}

%% Authors
\author[1]{\fnm{Zhou} \sur{Fang}}\email{zhou.fang@u.nus.edu}

\author*[1,2]{\fnm{Gianmarco} \sur{Mengaldo}}\email{mpegim@nus.edu.sg}

% Affiliations
\affil[1]{\orgname{Department of Mechanical Engineering, National University of Singapore}, \\ \country{9 Engineering Drive 1, Singapore, 117575}}
\affil[2]{\orgname{Department of Mathematics (by courtesy), National University of Singapore}, \\ \country{10 Lower Kent Ridge Road, Singapore, 119076}}

%TC:ignore
\abstract{
In machine learning forecasting, standard error metrics such as mean absolute error (MAE) and mean squared error (MSE) quantify discrepancies between predictions and target values. However, these metrics do not directly evaluate the physical and/or dynamical consistency of forecasts, an increasingly critical concern in scientific and engineering applications.

Indeed, a fundamental yet often overlooked question is whether machine learning forecasts preserve the dynamical behavior of the underlying system. Addressing this issue is essential for assessing the fidelity of machine learning models and identifying potential failure modes, particularly in applications where maintaining correct dynamical behavior is crucial.

In this work, we investigate the relationship between standard forecasting error metrics, such as MAE and MSE, and the dynamical properties of the underlying system. 
To achieve this goal, we use two recently developed dynamical indices:  the instantaneous dimension ($d$), and the inverse persistence ($\theta$).
Our results indicate that larger forecast errors -- e.g., higher MSE -- tend to occur in states with higher $d$ (higher complexity) and higher $\theta$ (lower persistence). 
To further assess dynamical consistency, we propose error metrics based on the dynamical indices that measure the discrepancy of the forecasted $d$ and $\theta$ versus their correct values. 
Leveraging these dynamical indices-based metrics, we analyze direct and recursive forecasting strategies for three canonical datasets -- Lorenz, Kuramoto-Sivashinsky equation, and Kolmogorov flow -- as well as a real-world weather forecasting task. 
Our findings reveal substantial distortions in dynamical properties in ML forecasts, especially for long forecast lead times or long recursive simulations, providing complementary information on ML forecast fidelity that can be used to improve ML models.
% Furthermore, we demonstrate that DI-based error metrics can effectively analyze real-world problems, enabling the identification of weather states with high dynamical errors. 
% Overall, our work provides a powerful framework for assessing the dynamical consistency of ML forecasts, offering valuable insights for improving future ML models.
}


\maketitle
%TC:endignore

%%%%%%%%%%%%%%%%%%%%%%%%%%%%
% 1. INTRODUCTION
%%%%%%%%%%%%%%%%%%%%%%%%%%%%
\section{Introduction}
\label{sec:introduction} 

%% Background on traditional forecasting
Forecasting, the process of making predictions about future states of a system based on past and present information, is closely related to dynamical systems~\cite{lorenz1963deterministic}. 
The latter are systems that evolve in time according to some rules, namely ordinary or partial differential equations, that are commonly derived from first principles~\cite{strogatz2018nonlinear}. 
The resulting equation-based models provide a rigorous mathematical representation of the system behavior, and their solution is usually approximated via conventional numerical methods, including spectral, finite difference, finite element and spectral element methods (see e.g.,~\cite{maday1989spectral,hughes2000finite,karniadakis2005spectral,leveque2007finite,quarteroni2008numerical}).  
This equation-based approach has proven extremely successful, yielding accurate and actionable solutions across different disciplines, including weather and climate science~\cite{bauer2015quiet} and engineering~\cite{mengaldo2021industry}, among many others.

% ML forecasting 
The emergence of machine learning (ML) has led to a paradigm shift in forecasting, with researchers and practitioners increasingly leveraging these data-driven methods as alternatives to traditional equation-based models. 
Unlike traditional approaches that explicitly leverage physical laws (i.e., our knowledge about the system) in the form of ordinary or partial differential equations, ML models learn complex patterns directly from data; this often without explicitly enforcing the underlying governing equations. 
ML models have demonstrated the ability to achieve accurate predictions for both canonical dynamical systems~\cite{pathak2018model} and real-world applications, such as weather~\cite{lam2023learning,bi2023accurate,bodnar2024aurora,price2025probabilistic} and climate~\cite{wang2024orbit,watt2024ace2,wang2025condensnet}.

Despite significant progress, several key challenges remain. 
In particular, ML models -- including neural networks -- often struggle to accurately capture fine-scale structures in long-term predictions~\cite{chakraborty2025binned}. 
Additionally, they can exhibit instability or unphysical behavior, limiting their reliability in applications where high-fidelity is paramount.

More fundamentally, ML models often function as black boxes, making it difficult to assess whether they adhere to established physical principles encoded in equation-based models (e.g.,~\cite{ben2024rise,chattopadhyay2023long}). 
To address these limitations, various promising strategies have emerged, including physics-informed ML approaches~\cite{karniadakis2021physics}, which weakly embed partial differential equations (PDEs) into the model, and explicit physical constraints that enforce the conservation of key physical quantities~\cite{wang2025condensnet}.

However, little to no attention has been given to evaluating the physical and/or dynamical fidelity of ML forecasts, other than looking at traditional error metrics such as mean squared error, and its variants~\cite{hyndman2006another,botchkarev2018performance}. 

In this work, we propose error metrics that directly evaluate the physical fidelity of ML forecasts from a dynamical perspective. 
The proposed error metrics leverage local dynamical indices (DI) derived from recent advances in dynamical systems theory. 
DI have provided a mathematically rigorous and purely data-driven framework for analyzing local (also referred to as instantaneous) dynamical properties of complex systems~\cite{lucarini2016extremes}. 
This framework consists of two dynamical indices: (i) the local dimension $d$ that provides information on the system's dynamical complexity, and (ii) the inverse persistence $\theta$ which describes how fast the trajectory leaves the current state. 
Several works have shown that $d$ and $\theta$ can provide useful dynamical and physical insights in many disciplines, including atmospheric sciences~\cite{faranda2017dynamical,dong2024multiscale}, 
oceanography~\cite{liu2021dynamical}, and fluid mechanics~\cite{messori2020dynamical}.
Dynamical indices have also been recently applied to identify the differences between simulated and real slow earthquakes, showing that current numerical models may not suffice to describe the dynamical complexity of natural observations~\cite{gualandi2024similarities}. 
More recent developments introduced a new predictability metric for dynamical systems, based on the the DI framework~\cite{dong2024revisiting}.

Since differences in dynamical indices indicate discrepancies in dynamical properties, we use the proposed DI-based error metrics as a quantitative measure of dynamical consistency for ML forecasts.

Evaluating machine learning forecasts using these error metrics reveals that ML models produce larger forecast errors in regions characterized by higher dimension and lower persistence.
This underscores the expected potential limitations in capturing complex and fast dynamics. 
Although the predicted system mean dimension and persistence closely resemble those of the true system, the dynamical error grows substantially with recursive predictions, indicating a decline in dynamical stability and dynamical fidelity.

The proposed dynamical metrics can complement existing and standard error metrics, providing a purely data-driven and model agnostic way of assessing dynamical fidelity of ML forecasts.

%%%%%%%%%%%%%%%%%%%%%%%%%%%%%%%%%%%
% RESULTS
%%%%%%%%%%%%%%%%%%%%%%%%%%%%%%%%%%%
\section{Results}
\label{sec:results}

\subsection{Error metrics and data}
\label{sec:results-data}

\begin{figure}[H]
    \centering
    \includegraphics[width=1.0\textwidth]{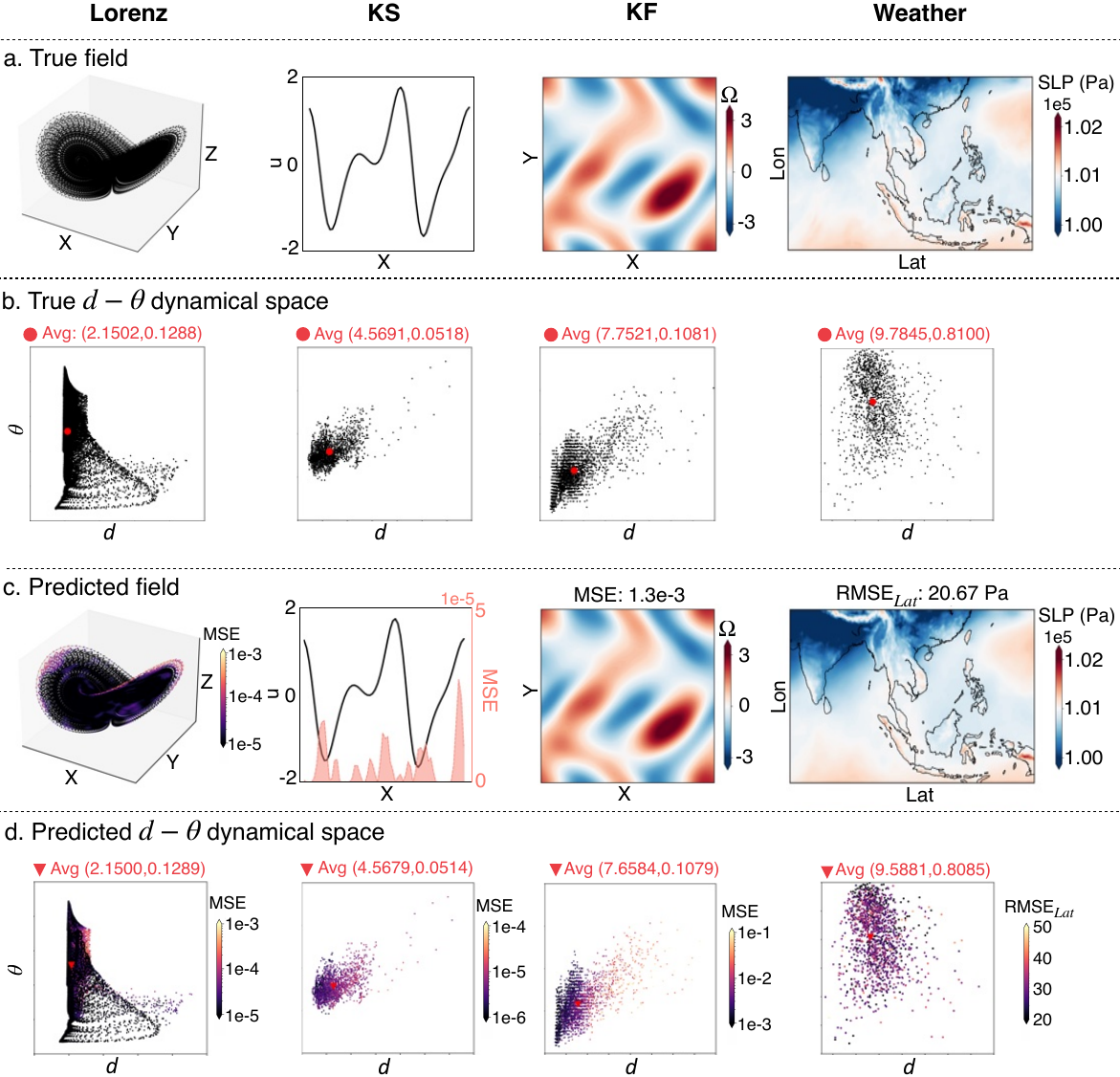}
    \caption{\textbf{Overview of datasets.} 
    Panel (a): Ground truth solution for each dataset, used as `true data' for ML learning. 
    Panel (b): Dynamical space of true data, where each point represents a data snapshot. The coordinates $d$ and $\theta$ are dynamical indices that describe the dynamical properties of each state. The mean values of the indices are highlighted with red circle and text.
    Panel (c): ML forecast solution, accompanied by standard forecast errors, namely MSE (and RMSE for the weather dataset). 
    Panel (d): Dynamical space of ML forecasts. Each forecast state is plotted at corresponding $d$ and $\theta$, colored by the forecast error. The average dynamical indices of the forecasts are marked with red triangle and text. 
    }
    \label{fig:Fig1}
\end{figure}

We outline the proposed analysis approach on three different canonical systems, namely the Lorenz-63 model (referred simply to as Lorenz dataset hereafter), the Kuramoto-Sivashinsky equations (KS), and the Kolmogorov flow (KF), and on a real-world problem, namely weather forecasting (referred simply to as weather dataset hereafter), where we focus on the mean sea level pressure (SLP).
These are depicted in Fig.~\ref{fig:Fig1}, where Fig.~\ref{fig:Fig1}a illustrates the Lorenz dataset, and a snapshot of the KS, KF, and weather datasets. 
Fig.~\ref{fig:Fig1}b, shows the $d-\theta$ dynamical space, where $d$ and $\theta$ are the two dynamical properties that we are measuring, and that are introduced in section~\ref{sec:methods-indices}. 
Each point in Fig.~\ref{fig:Fig1}b represents a time snapshot, and the shape of the point clouds characterizes the dynamical properties of the system, with distinguishable differences across the different systems considered.  
Fig.~\ref{fig:Fig1}c presents sample predictions of each system, along with the corresponding mean squared error (MSE). 
Fig.~\ref{fig:Fig1}d is the dynamical space of the predicted states, colored by their MSE errors. 
A detailed description of each dataset is provided in section~\ref{sec:data_and_method}.

% \textcolor{red}{GM: add machine learning models considered, and mention about hyperparameter optimization. The machine learning models considered range from ...}
The ML models considered are the prevailing architectures that practitioners are using for a range of ML forecasting tasks, namely Convolutional Neural Networks (CNN), Long Short-Term Memory neural networks (LSTM)~\cite{sak2014long}, Transformers ~\cite{vaswani2017attention,dosovitskiy2020image}, and the Graph Neural Networks (GNN)~\cite{kipf2016semi}. 
To ensure a fair comparison, we perform hyperparameter optimization for each architecture and input length using the Tree-structured Parzen Estimator (TPE) sampling algorithm~\cite{watanabe2023tree} implemented in the open-source package Optuna~\cite{akiba2019optuna}.
We additionally adopted two state-of-the-art models used for weather forecasting applications, namely the Transformer-based Pangu-Weather model~\cite{bi2023accurate} and the GNN-based GraphCast model~\cite{lam2023learning}.
We measure the performance of each ML forecast using traditional error metrics, namely MSE (and its variants), that is
\begin{equation}
\mathrm{MSE} = \frac{1}{N_t}\frac{1}{N_s}\sum_{i=1}^{N_t}\sum_{i=1}^{N_s}(\hat{y} - y)^2,
\label{eq:mse-y}
\end{equation}
where $\hat{y}$ is the ML predicted solution, and $y$ is the true value (i.e., the target of the ML task), $N_t$ is the number of time samples, and $N_s$ represents the number of space samples (e.g., spatial locations). 
We then measure the MSE for the dynamical indices $d$ and $\theta$, that is
\begin{subequations}
\begin{align}
    \mathrm{MSE}_{d}      &= \frac{1}{N_t}\sum_{i=1}^{N_t}(\hat{d} - d)^2 \label{eq:di-d} \\
    \mathrm{MSE}_{\theta} &= \frac{1}{N_t}\sum_{i=1}^{N_t}(\hat{\theta} - \theta)^2, \label{eq:di-theta}
\end{align}
\label{eq:mse-d-theta}
\end{subequations}
where $\hat{d}, \hat{\theta}$ are the ML predicted dynamical indices, and $d, \theta$ are their true values (i.e., the true dynamical indices of the system). 
In Eq.~\eqref{eq:mse-d-theta}, we do not have dependence on the space dimension, as the space component is contracted when calculating the DI, $d$ and $\theta$, as reported in section~\ref{sec:methods-indices}, where the interested reader can find more details.

Similar to MSE, $\mathrm{MSE}_d$ and $\mathrm{MSE}_\theta$ quantify the magnitude of prediction errors as positive values, making them suitable for both sample-wise and statistical evaluation, such as computing averages over entire datasets.
For tasks where the sign of the dynamical error carries physical significance, we introduce error metrics based on simple DI differences (briefly DID) as a sample-wise diagnostic tool, that is
\begin{subequations}
    \begin{align}
        \mathrm{DID}_{d}      &= \hat{d} - d\label{eq:did-d} \\
        \mathrm{DID}_{\theta} &= \hat{\theta} - \theta \label{eq:did-theta}.
    \end{align}
    \label{eq:did}
\end{subequations}
DID preserves the sign of the forecasted $d$ and $\theta$ values relative to the ground truth, thereby capturing whether the predicted dynamical indices are over- or underestimated.

For each test case, we consider both direct single-step forecasts (section~\ref{sec:results-direct}), and recursive ones (section~\ref{sec:results-recursive}), as these are the two main forecasting workflows commonly adopted in machine learning. 
We also provide a more in-depth analysis of the weather dataset in section~\ref{sec:results-weather}, to show how these indices and index-based metrics can be useful in the context of practical real-world applications.

\subsection{Dynamical errors in direct forecasts}
\label{sec:results-direct}

\begin{figure}[H]
    \centering
    \includegraphics[width=0.75\linewidth]{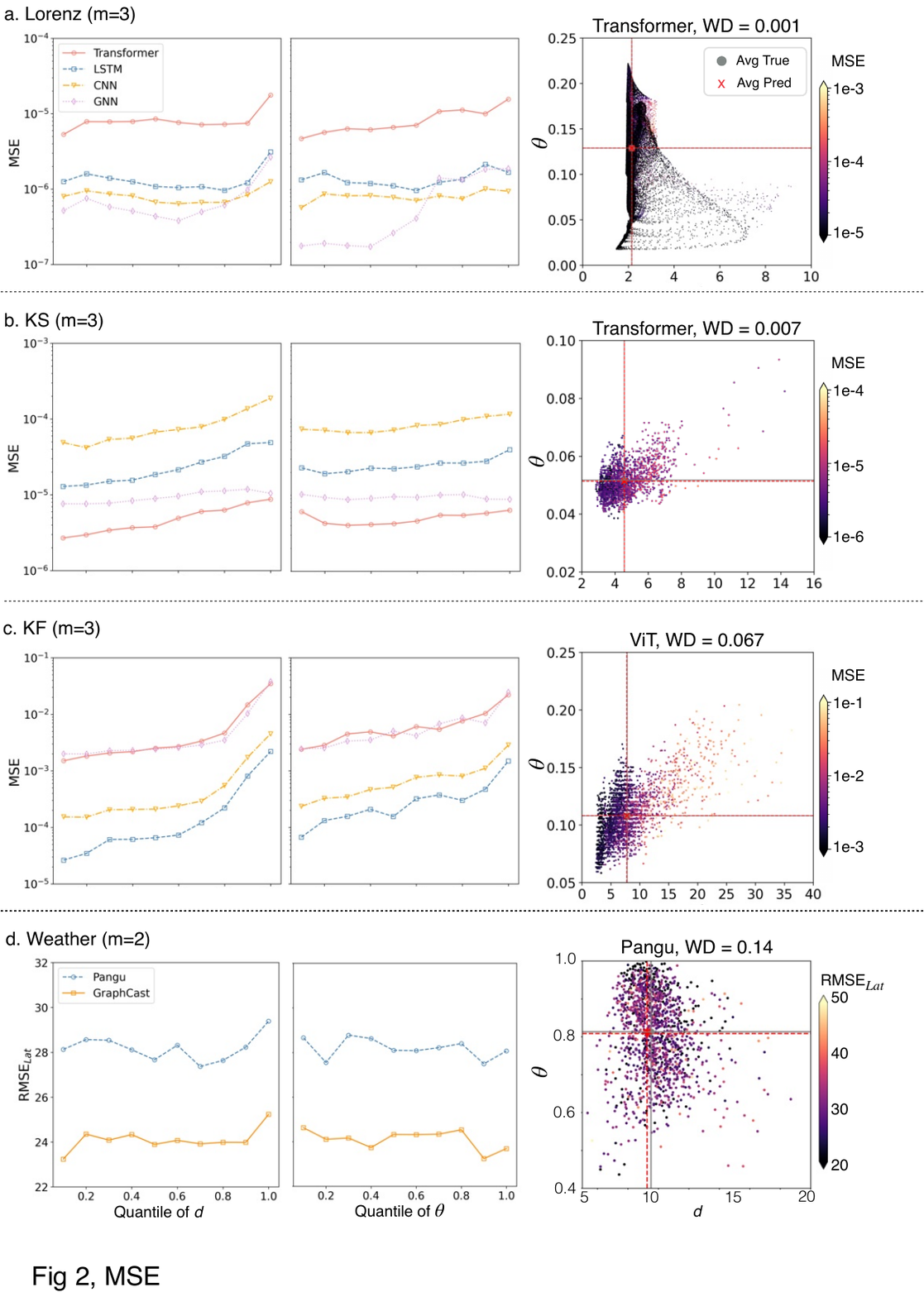}
    \caption{\textbf{Relationship between forecast error and dynamical indices (1-step time lead; direct forecasts).}
    Each panel represents one dataset (where $m$ is the input length used). 
    The left/middle columns show mean MSE (and RMSE for the weather datasset) vs.\ quantiles of $d$ (left) and $\theta$ (middle), with forecasts grouped into 10 bins. 
    The right column shows the $d-\theta$ space of each forecast colored by MSE (and RMSE for the weather dataset) forecast error, alongside average true/predicted indices. 
    At the top of each plot in the right column, we report the Wasserstein Distance (WD), that measures differences in these $(d,\theta)$ distributions; smaller WD indicates a closer match.
    }
    \label{fig:Fig2}
\end{figure}

In Fig.~\ref{fig:Fig2}, we show how the values of $d$ and $\theta$ (introduced in section \ref{sec:methods-indices}) relate to the behavior of standard error metrics, namely MSE as calculated in Eq.~\eqref{eq:mse-y}, for direct forecasts with a lead time of 1 time step. 
\noindent The input length used is 3 time steps for the canonical datasets (i.e., Lorenz, KS and KF), and 2 time steps for the weather dataset, as the ML forecasts for the latter are directly taken from WeatherBench2~\cite{rasp2023weatherbench} (a widely used benchmark for ML weather forecasting).

Fig.~\ref{fig:Fig2}a shows results for Lorenz, Fig.~\ref{fig:Fig2}b for KS, Fig.~\ref{fig:Fig2}c for KF, and Fig.~\ref{fig:Fig2}d for the weather dataset. 
The first column of Fig.~\ref{fig:Fig2} depicts MSE as a function of the $d$ quantile, while the second column as a function of the $\theta$ quantile. 
The third column depicts the $d-\theta$ space, colored by MSE, where we also report the Wasserstein Distance (WD) as the title of each plot for one of the models used for each dataset.
Notably, MSE tends to be higher for high values of $d$ and $\theta$ for all canonical datasets.
% (except in the case of Weather, that does not display this behavior for $\theta$ -- noting that this dataset has a limited amount of time snapshots available, aspect that may affect the results). 
In other words, higher complexity (high $d$) and low persistence (high $\theta$) are predictors of high MSE, for the analyzed datasets. 
Indeed, this behavior is also true if we were to consider other standard error metrics -- see Supplementary Information section~\ref{si:errors-vs-quantile}.

High $d$ and $\theta$ generally correlate with higher MSE, yet the MSE--quantile patterns differ.
For the Lorenz dataset, MSE increases for $d>0.8$ and $\theta>0.6$, with a plateau for quantiles between 0.2 and 0.8.
In contrast, KS and KF exhibit more monotonic trends, flatter in the KS case (especially for Transformer and GNN).
For the weather dataset (lead time: 6\,h, WeatherBench2~\cite{rasp2023weatherbench}), fewer time snapshots yield flatter $\theta$ behavior.
GraphCast shows a plateau for $d$ (0.2--0.9) before a steep rise, whereas Pangu-Weather fluctuates, increasing consistently only for $d>0.7$.
These differences arise from (i) the diagonally-shaped $d$--$\theta$ space, where high complexity and low persistence may jointly increase forecast difficulty, and (ii) the 6\,h interval capturing regular yet not persistent daytime fluctuations, weakening the correlation between $\theta$ and error.

Turning to the $d-\theta$ space in the third column of Fig.~\ref{fig:Fig2}, we observe how the average dynamical properties for the one-step direct forecasts are similar to the true value. 
Yet, the distribution of the $d-\theta$ space mirrors what is observed in terms of the MSE and $d-\theta$ quantile distributions: higher values of $d$ and $\theta$ are associated with larger MSE.
% Interestingly, the WD exhibits larger values for KS than for Lorenz, although the overall MSE of KS is smaller than Lorenz. 
% This may indicate that, while the MSE performance is better for KS, its dynamical consistency is worse, an aspect that is critically important for physics-driven applications. 

The results for lead time of 1 time step generalize to longer lead times, as shown in section~\ref{si:lead_forecast}, where we conduct experiments with lead times of 10, 20, 30, 40 time steps across the three canonical datasets (with the corresponding Lyapunov time (LT) or time unit (TU) values shown in the figures). 
Our findings indicate a similar monotonic increase in error for KS versus $d$ and $\theta$, even for a large lead (3.0 LT). 
Additionally, the Lorenz and KF datasets exhibit a plateau-and-rising-tail error pattern vs DI quantile. 

Results for mean absolute error (MAE, as defined in Eq.~\eqref{eq:mae_metrics}a), normalized mean absolute error (NMAE, as defined in Eq.~\eqref{eq:nmae_metrics}a) and normalized mean square error (NMSE, as defined in Eq.~\eqref{eq:nmse_metrics}a) are reported in Supplementary Information section~\ref{si:errors-vs-quantile}. 
Results for larger lead times and different input lengths are presented in Supplementary Information section~\ref{si:lead_forecast} and ~\ref{si:inputs}, respectively.
These results exhibit similar trends with those presented in this section, further supporting the findings.

For detailed forecast error values, we refer the readers to Extend Data Tab.~\ref{tab:ml_metrics_mse}, and to Supplementary Tab.~\ref{si-tab:ml_metrics_nmse} in Supplementary Information section~\ref{si:ml_stats}, showing MSE, $\mathrm{MSE}_d$, and $\mathrm{MSE}_\theta$ and their normalized variants, namely NMSE, $\mathrm{NMSE}_d$, and $\mathrm{NMSE}_\theta$ (defined in section \ref{sec:methods-indices}).
We find that the model with the lowest MSE does not necessarily exhibit the highest dynamical consistency, as indicated by $MSE_d$ and $MSE_\theta$.
Indeed, MSE and related metrics might not be sufficient to detect unphysical behavior, calling for metrics with physical insights -- see for instance the realm of weather applications~\cite{bracco2024machine,bonavita2024some}.

\begin{figure}[H]
    \centering
    \includegraphics[width=0.9\linewidth]{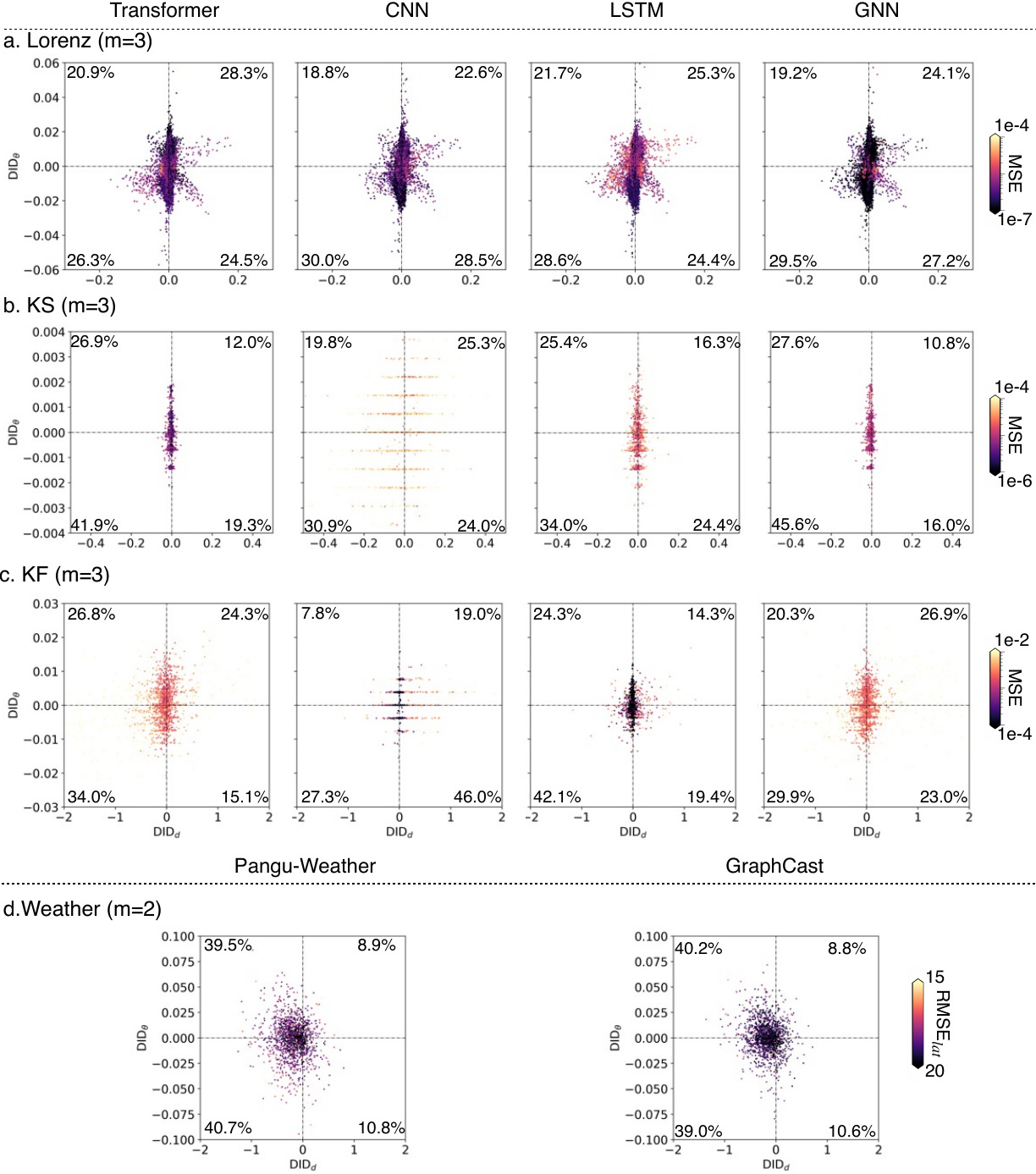}
    \caption{\textbf{DID of Lorenz, KS and KF, and weather dataset.}
    In each subplot, the $x$-axis represents $\mathrm{DID}_d$ and the $y$-axis represents $\mathrm{DID}_\theta$. The percentage of points falling into each quadrant is displayed at the corresponding corner. Points are colored according to their MSE}
    \label{fig:did_direct}
\end{figure}

In Fig.~\ref{fig:did_direct}, we show the DID results of the canonical datasets and the weather dataset.
For each forecasted state, we calculate $\mathrm{DID}_d$ and $\mathrm{DID}_\theta$ according to Eq.~\eqref{eq:did} (also reported in section~\ref{sec:methods-indices}).
The value of DID, as visualized in the $\mathrm{DID}_d-\mathrm{DID}_\theta$ space, provides information on whether $d$ and $\theta$ is overestimated or underestimated (i.e. a positive $\mathrm{DID}_d$ indicates the forecasted $d$ is larger than the true value). 
The points are colored by MSE of the corresponding state.
We find that DID and MSE are sometimes positively correlated.
For instance, in KF, points located at the edge of the cluster -- indicating large errors in either the positive or negative direction -- tend to exhibit higher MSE.
However, this relationship does not hold universally.
In the Lorenz system, for example, points near the origin can have relatively high MSE, while certain states with large dynamical errors may not be reflected in the MSE values.
In the weather dataset, DID analysis reveals a systematic underestimation of $d$, suggesting that the forecasted system exhibits reduced dynamical complexity compared to the ground truth.

% For detailed forecast error values, we refer readers to the summary of error statistics presented in Table~\ref{tab:ml_metrics_mse}, located in Supplementary Information section~\ref{si:ml_stats}.

% \textcolor{red}{GM: integrate the following text here. Results in the SI should always be discussed together after results in the main have been fully discussed. }

\subsection{Dynamical errors in recursive forecasts}
\label{sec:results-recursive}

\begin{figure}[H]
    \centering
    \includegraphics[width=0.8\linewidth]{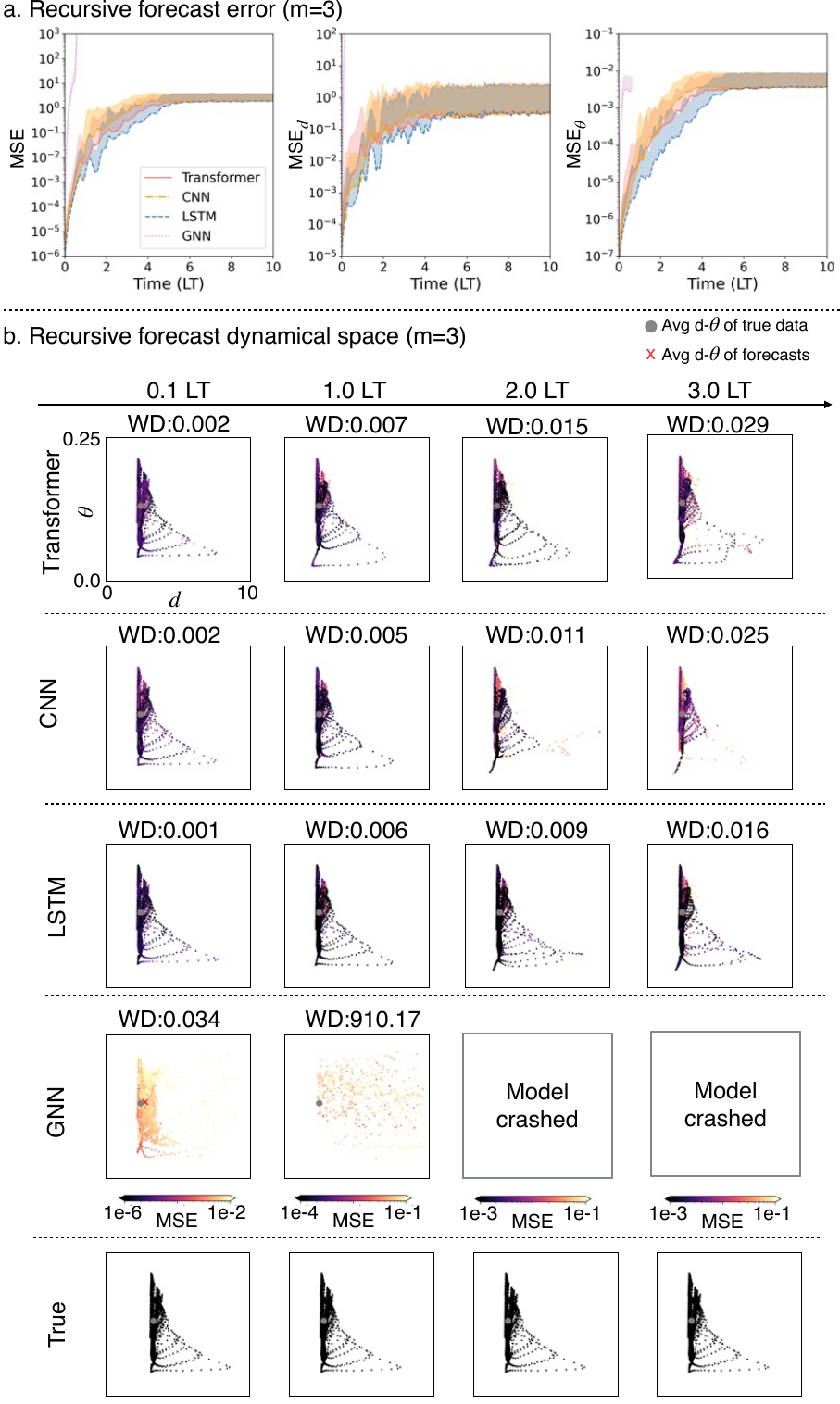}
    \caption{\textbf{Error and dynamical space of Lorenz recursive forecast.} 
    Panel (a): Forecast error vs recursive forecast time in terms of Lyapunov time (LT). 
    The shaded area represents the standard deviation of forecasts starting from 5000 initial states. 
    Panel (b): $d-\theta$ space of the 5000 trajectories at forecast time 0.1 LT, 1.0 LT, 2.0 LT and 3.0 LT. 
    The horizontal and vertical coordinates are $d$ and $\theta$, respectively.
    The mean value of the indices and WD is annotated on the figure. 
    The GNN $d-\theta$ spaces for 2.0 LT and 3.0 LT are not plotted since the model crashed.}
    \label{fig:Fig3_lorenz}
\end{figure}

% \subsubsection{Canonical datasets}
In Fig.~\ref{fig:Fig3_lorenz}, we analyze recursive forecast errors for the Lorenz dataset across the different ML models considered, using input length $m$ equals to 3 time steps. 
In particular, Fig.~\ref{fig:Fig3_lorenz}a shows the MSE (left column), and the dynamical errors MSE$_{d}$ and MSE$_{\theta}$ (middle and right columns, respectively) vs time. 
The definition of the dynamical error metrics has been briefly introduced in equations~\eqref{eq:di-d} and \eqref{eq:di-theta}, with more details provided in section~\ref{sec:methods-indices}. 
The $x$-axis of each plot in Fig.~\ref{fig:Fig3_lorenz}a reports the recursive forecast time normalized by Lyapunov time of the system, as introduced in section~\ref{sec:time_scale}.
% (Lyapunov time for Lorenz and KS, and forcing period for KF). 
As the forecast time increases, MSE, MSE$_d$ and MSE${_\theta}$ also increase for all models, albeit with different slopes.
More specifically, Transformer, CNN, and LSTM are able to generate stable forecasts for long recursive time horizons (i.e., 10 LT), with bounded MSE and dynamical errors. 
GNN, despite similar performance compared to other models for lead time of 1 time step (as shown in Fig.~\ref{fig:Fig2}, Extended Data Tab.~\ref{tab:ml_metrics_mse}) and Supplementary Tab.~\ref{si-tab:ml_metrics_nmse}), is sensitive to error accumulation and finally crashes after around 1 LT forecast time, indicating poor stability for longer forecast horizons.
Fig.~\ref{fig:Fig3_lorenz}b visualizes the evolution of the $d-\theta$ space at recursive forecast times of 0.1, 1.0, 2.0, and 3.0 LT, colored by MSE. 
For comparison, each column shares the same color bar.
Indeed, we observe how the $d-\theta$ space becomes progressively distorted, with WD increasing for forecasts farther into the future (i.e., longer recursive forecast time). 
Specifically, Transformer, CNN, and LSTM maintain the shape of $d-\theta$ space with a long recursive time, with partial loss of dynamics (e.g., the bottom part of CNN from 2.0 LT onward is lost compared to both earlier predictions and the ground truth). 
In contrast, the $d-\theta$ space of GNN forecasts spreads rapidly by 1 LT and totally loses the true structure, before the model ultimately crashes.
Extended Data Fig.~\ref{fig:Fig3_ks}, \ref{fig:Fig3_kf}, and~\ref{fig:fig3_weather} share the same information as Fig.~\ref{fig:Fig3_lorenz}, but for the KS, KF and weather dataset, respectively. 

For KS and KF, the forecast error increases and saturates for most models with increasing recursive time, as shown in Extended Data Fig.~\ref{fig:Fig3_ks}a and Extended Data Fig.~\ref{fig:Fig3_kf}a.
This trend is consistent with the behavior observed for the Lorenz system.
Extended Data Fig.~\ref{fig:Fig3_ks}b and Extended Data Fig.~\ref{fig:Fig3_kf}b reveal a significant distortion of dynamical properties after 1 and 2 LT of forecast time, respectively. 
This distortion is particularly significant for the GNN model in the KS dataset, and for the CNN, LSTM, and GNN models in the KF dataset. 
In these cases, the point clouds in the $d-\theta$ space exhibit a noticeable displacement as a whole, typically towards higher $d$ values.
This behavior can be attributed to the fact that forecasts with large errors tend to deviate from the historical data in phase space, possibly falling into regions close to the borders of the attractor. Dimension $d$ may reveal an anomalously high value in these regions, as reported in literature~\cite{platzer2025density,faranda2017dynamical}.
Specifically for the weather dataset, as shown in Extended Data Fig.~\ref{fig:fig3_weather}a, the forecast error continues to increase over the 40-step forecast horizon, without signs of saturation. 
In Fig.~\ref{fig:fig3_weather}b, the predicted 
$d-\theta$ space becomes increasingly compact and shrinks in coverage as forecast time grows, indicating a decline in dynamical consistency and a loss of richness in dynamics.
In Supplementary Information section~\ref{si:recursive}, we show the same nine figures for other standard error metrics calculated on the three canonical datasets, with findings consistent with the ones presented in this section.

\begin{figure}[H]
    \centering
    \includegraphics[width=0.9\linewidth]{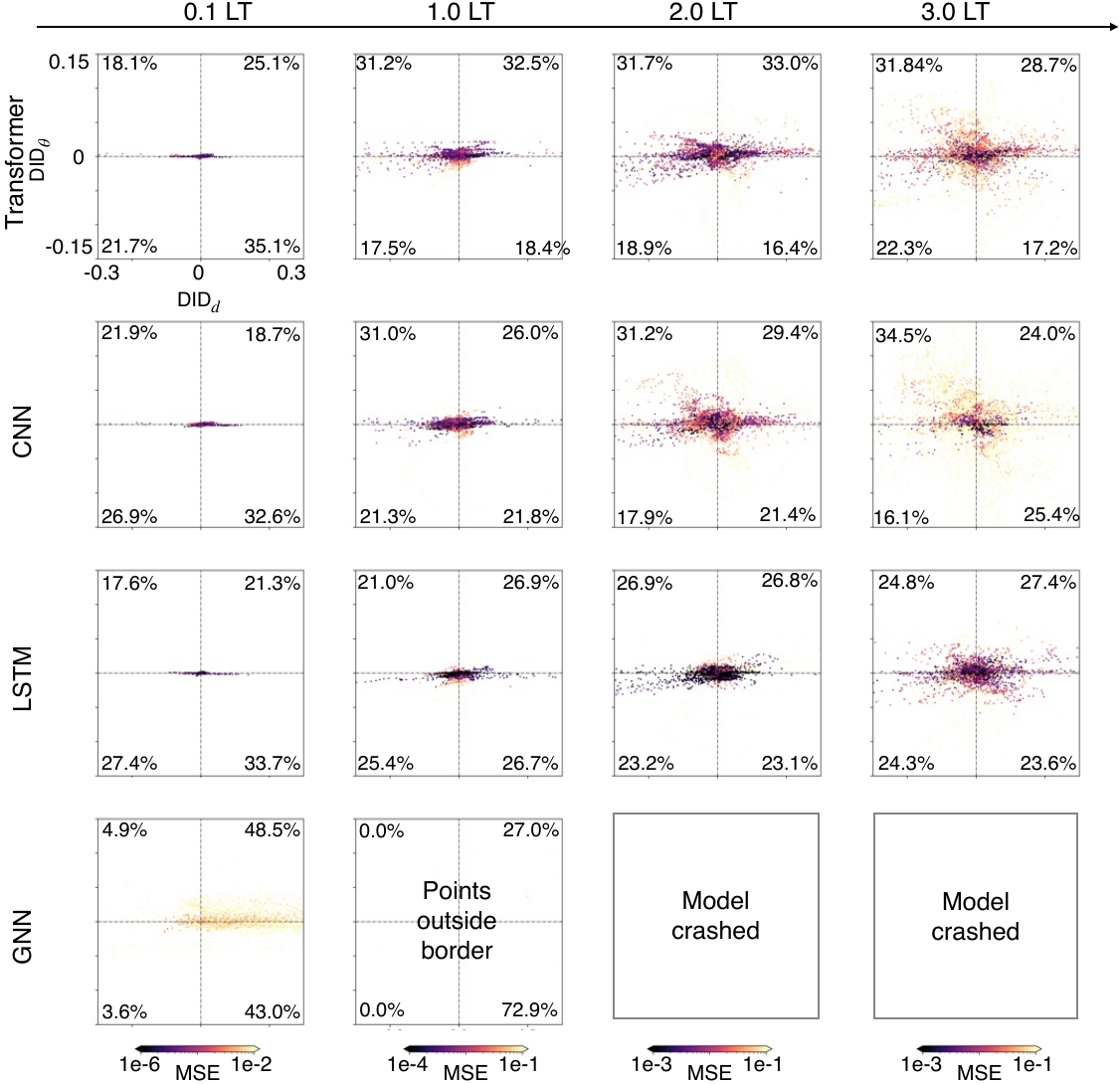}
    \caption{\textbf{DID of recursive forecast for Lorenz.} The timeline at the top indicates the recursive forecast time. Each row of subplots corresponds to a distinct machine learning architecture. In each subplot, the $x$-axis represents $\mathrm{DID}_d$ and the $y$-axis represents $\mathrm{DID}_\theta$, with consistent axis ranges across all subplots. The percentage of points falling into each quadrant is displayed at the corresponding corner. Points are colored according to their MSE.}
    \label{fig:recursive_did_lorenz}
\end{figure}

Additionally, the DID results for both canonical and weather datasets are presented in Fig.~\ref{fig:recursive_did_lorenz}, Extended Data Fig.~\ref{fig:recursive_did_ks}, \ref{fig:recursive_did_kf} and~\ref{fig:recursive_did_weather}.
As forecast time increases, we observe a progressive dispersion of points in the $\mathrm{DID}d$–$\mathrm{DID}\theta$ space across all four datasets, indicating reduced dynamical consistency, confirming previous findings.
In the Lorenz system, the probability of under- or overestimating the dynamical indices is relatively uniform compared with other datasets, with the percentage of points lying in each quadrant closer to 25\%.
In contrast, for the KS and the KF dataset, both $d$ and $\theta$ tend to be severely overestimated across nearly all model architectures, particularly at longer recursive forecast time.
For the weather dataset, $d$ tends to be systematically underestimated, indicating a decreased dynamical complexity of forecasts relative to the true system.

\subsection{Further insights on real-world weather dataset}
\label{sec:results-weather}

\begin{figure}[H]
    \centering
    \includegraphics[width=0.9\linewidth]{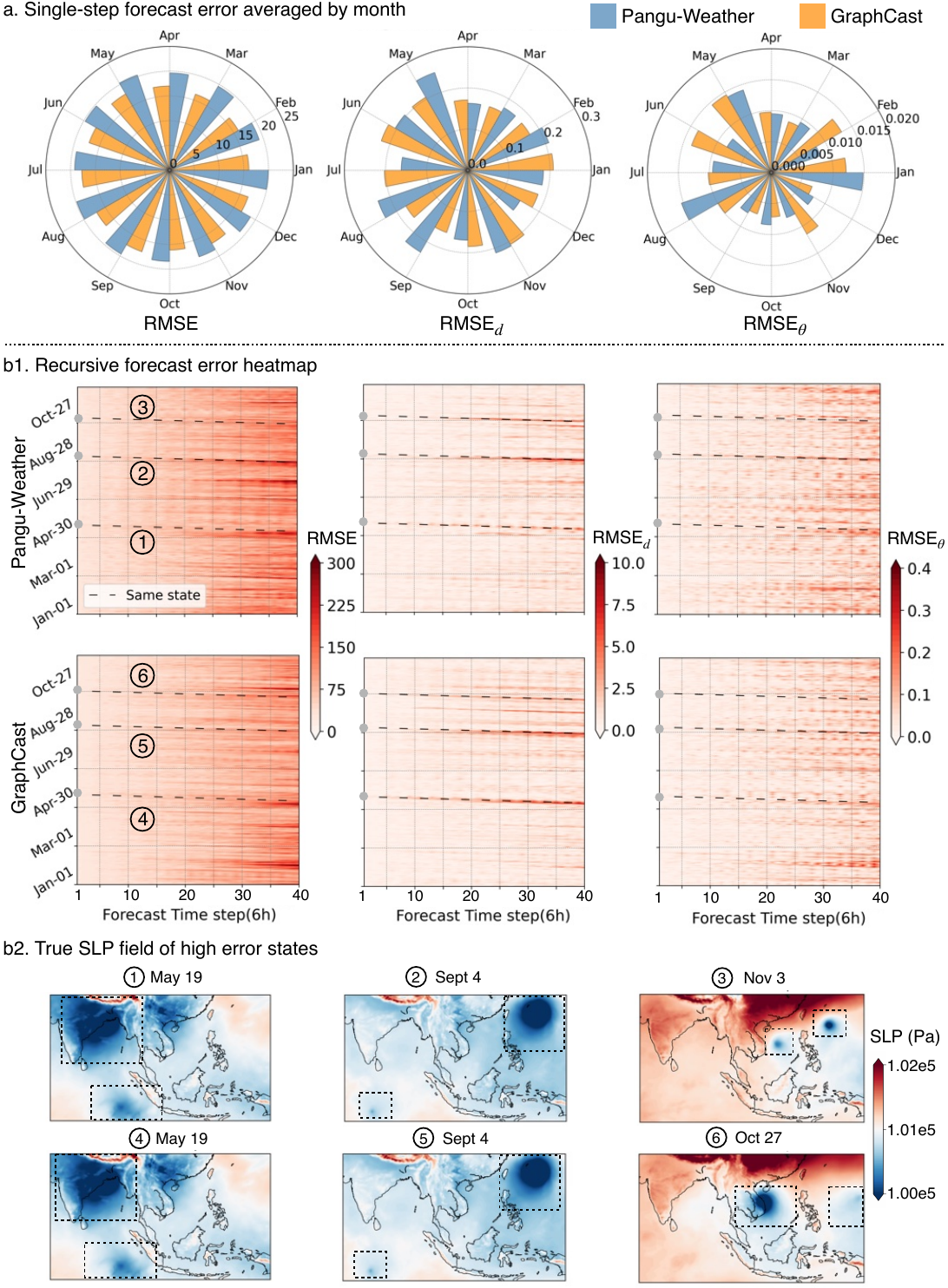}
    \caption{\textbf{Further analysis on direct 1-step and recursive weather forecast.} Panel (a): Direct 1-step forecast error averaged over calender months.
    Panel (b): RMSE heatmap of Pangu-Weather and GraphCast. 
    The $x$-axis denotes recursive forecast steps and the $y$-axis the forecast starting date. 
    Left column of panel (a-b): RMSE; Middle and right columns: RMSE$_d$ and RMSE$_\theta$. 
    Panel (c): Six SLP fields from the true data, corresponding to the six high-error states indicated by the dashed line and numbered circles in the heatmaps.}
    \label{fig:Fig4_weather}
\end{figure}

In Fig.~\ref{fig:Fig4_weather}a, we further analyze the direct 1-step weather forecasts generated by Pangu-Weather and GraphCast. 
The left column shows the latitude-weighted RMSE (defined in Eq.~\eqref{eq:rmse_lat}, hereafter referred to as RMSE) averaged over each month, for the year 2020. 
Notably, both models exhibit relatively uniform RMSE throughout the year, though GraphCast consistently reports lower RMSE values.
The middle and right columns show the dynamical errors based on $d$ and $\theta$, with substantial variation across different months. 
For instance, Pangu-Weather forecasts exhibit higher RMSE$_d$ in May, August, and September, and higher MSE$_\theta$ in January, May, and August;  GraphCast shows larger RMSE$_d$ in June and November, and higher RMSE$_\theta$ for May and June.
% %
% \begin{figure}[H]
%     \centering
%     \includegraphics[width=0.9\linewidth]{figures/Fig4_weather.pdf}
%     \caption{\textbf{Monthly mean error for single-step weather forecast.} Left: Latitude-weighted RMSE; Middle and right: dynamical error metrics MSE$_d$ and MSE$_\theta$.}
%     \label{fig:Fig5_weather}
% \end{figure}
% %

Fig.~\ref{fig:Fig4_weather}b1 and ~\ref{fig:Fig4_weather}b2 shows the forecast errors for 40-step recursive weather forecasts, and 6 featured SLP fields. 
The forecasts start at 6:00 am each day with a time step of 6 hours.
Fig.~\ref{fig:Fig4_weather}b1 display the heatmaps of recursive forecast error, for Pangu-Weather (upper row) and GraphCast (lower row), respectively. 
The left column shows the forecast errors in terms of RMSE, and RMSE$_d$ and RMSE$_\theta$ are reported in the middle and right columns. 
The $y$-axis displays the forecast starting date, and the $x$-axis represents the recursive time step. 
The oblique dashed line in the figure denotes the same physical time in different forecasts, e.g., the 5th step forecasts starting from Mar 12 (predicting 6:00 am Mar 13) and the 1st step forecast starting at Mar 13 (also predicting 6:00 am Mar 13) correspond to a same target state. 
Notably, we find that there are several states that reveal a higher error than others, especially obvious for RMSE$_d$.
These high error states, highlighted by the same-time line and numbered circles, cannot be simply attributed to the error accumulation in the recursive forecasts, as they align with the same slope as the oblique line representing identical target times.
Instead, this indicates that some states might be intrinsically more challenging for ML models to forecast, which is consistent with our previous findings. 
The SLP fields corresponding to the six high-error states identified are shown in Fig.~\ref{fig:Fig4_weather}b2.
We find that these high-error states are commonly accompanied by several low-pressure regions, suggesting a possible relation to multi-cyclone systems.

In the Supplementary Information section ~\ref{si:cyclone_did} we present the same heatmap but using $\mathrm{DID}_d$ and $\mathrm{DID}_\theta$ metrics, with the same six high-error states highlighted.
Typically, states 1, 2, 4, 5, and 6 correspond to overestimation of $d$, whereas state 3 is associated with an underestimation of $d$.
This observation suggests that there might be different underlying mechanisms contributing to the high forecast errors observed across these states.
%

% \textcolor{red}{GM: can you bring the tropical cyclone figure here and discuss it. Thank you.}

\section{Discussion}
\label{sec:discussion}
% Data generated by dynamical systems reveal diverse dynamics, reflecting the underlying physical laws that govern the behaviors. 
% Capturing these dynamics is important, especially in applications where dynamical fidelity is critical, such as weather and climate forecasting.
Dynamical systems exhibit diverse behaviors, intrinsically reflecting the underlying physical principles and governing equations. 
Accurately capturing these dynamical characteristics is essential, particularly for applications requiring high physical fidelity, such as numerical weather prediction and climate modeling.
In this work, we leverage two dynamical indices (DI), namely $d$ and $\theta$ (see section~\ref{sec:methods-indices}) to capture these dynamical characteristics and assess the dynamical consistency of machine learning forecasts. 
To this end, we introduce DI-based error metrics (detailed in section~\ref{sec:method_metrics}) and investigate three canonical benchmark datasets (Lorenz, KS, and KF) as well as a real-world weather forecasting dataset.

We first show that the dynamical properties of the datasets analyzed are linked to the performance of ML models, based on the direct 1-step forecasting results shown in Fig.~\ref{fig:Fig2}. 
Specifically, data with higher $d$ (higher complexity) and $\theta$ (less persistent) tend to have larger ML forecast errors, as measured using standard error metrics, such as MSE and MAE, among others.
This indicates that those states with high $d$ and $\theta$ may either have lower intrinsic predictability, or they are simply less visited, resulting in fewer training samples available. 
To further evaluate dynamical consistency, we introduce $d$ and $\theta$-based dynamical error metrics (defined in section~\ref{sec:method_metrics}) to measure the differences in dynamics between the forecasted and true states, using the full set of historical information from the training data as a reference.
By comparing standard error metrics with dynamical error metrics, as presented in Extend Data Tab~\ref{tab:ml_metrics_mse} and Fig.~\ref{fig:Fig2}, we show that a lower MSE does not necessarily imply a smaller dynamical error, highlighting the need for incorporating dynamical consistency metrics into the evaluation framework. 
% \textcolor{red}{GM: cite the related result -- i.e., Fig. or Tab.}.

For recursive forecasts, we observe that both standard and dynamical error metrics increase with forecast time, accompanied by a progressive distortion in the $d-\theta$ dynamical space, as reported in Fig.~\ref{fig:Fig3_lorenz}, Extended Data Fig.~\ref{fig:Fig3_ks} and Extended Data Fig.~\ref{fig:Fig3_kf}.
By analyzing the evolution of $d–\theta$ space, we identify three distinct failure modes in recursive machine learning forecasts. 
The first mode corresponds to scenarios involving incomplete dynamical representation, characterized by regions of the forecast trajectories being absent in the $d–\theta$ space. This absence indicates that certain intrinsic dynamical features are not adequately captured by the predictive model.
The second mode is characterized by a systematic shift in the $d–\theta$ space. 
This mode is more likely to occur when the predicted states deviate significantly from the true states (i.e., longer recursive forecast time), usually towards an overestimation of $d$, as discussed in section~\ref{sec:results-recursive}. 
% \textcolor{red}{GM: I am unclear about overestimation -- is it overestimation or underestimation? usually towards an overestimation of $d$.}
The third mode reflects a breakdown in the $d–\theta$ space structure, where previously clustered point clouds become widely dispersed. 
All these failure modes reflect losses of dynamical consistency in the forecasts and highlight potential failure in the underlying ML models.

Additionally, in the weather forecast, our proposed dynamical metrics effectively identify states with high dynamical inconsistency, as shown in Fig.~\ref{fig:Fig4_weather}.
These states are not fully captured by the standard error metrics commonly adopted, such as the RMSE. 
Further analysis of these states reveals a potential link to multi-cyclone patterns, underscoring the additional diagnostic value offered by the approach proposed.

In summary, we identified a systematic relationship between regions of high forecast error and higher $d$ and $\theta$, which correspond to greater dynamical complexity and lower persistence, respectively. 
By leveraging DI and proposing DI-based error metrics, we analyzed the dynamical behavior of ML forecasts, offering a new perspective on their consistency of the underlying dynamics. 
This work highlights the importance of dynamical evaluation in forecasting tasks and opens new avenues for assessing and improving ML models from a dynamical standpoint.
It also constitutes the basis for future research, including: (i) enhancing the dynamical consistency of ML forecasts; (ii) incorporating dynamical information to guide or constrain model behavior; and (iii) investigating the causal relationship between dynamical deviations and forecast errors, for identifying early indicators of model failure.

%TC:ignore
\section{Methods}
\label{sec:methods}

% \textcolor{red}{GM: the whole methods is written in elementary english, and in many parts not coherent. Please address asap.} 

\subsection{Dynamical indices}
\label{sec:methods-indices}
Dynamical indices (DI) are quantitative measures used to characterize the fundamental properties of dynamical systems, such as complexity, persistence, and predictability. They offer valuable insights into the behavior of the system and help understand nonlinear, time-dependent processes.

Dynamical indices (DI) can be computed either globally or locally. 
In this work, we focus exclusively on local dynamical indices, which characterize the transient, state-dependent properties of a dynamical system at each point in the phase space (the space of the observables). 
Unlike global indices, which quantify average, time-invariant properties of the attractor (the space of the trajectories followed by the dynamical system over time) as a whole and thus fail to capture region-specific variations, local dynamical indices offer detailed insights into the trajectory evolving behavior across different regions of the attractor. 
Hence, they can be used to estimate the difference in terms of dynamical properties between ML forecasts and the underlying true values.

A brief overview of the definitions of the two dynamical indices, namely instantaneous dimension $d$ and inverse persistence $\theta$, is provided below. 
For full mathematical derivations and theoretical background, we refer the reader to~\cite{faranda2017dynamical,faranda2024statistical}.\\

\noindent \textbf{Instantaneous Dimension $d$.}
The instantaneous dimension $d$ at a given state $\zeta$ quantifies the effective degrees of freedom locally~\cite{datseris2023estimating}.
% , and is derived from the return probability of trajectories in its vicinity.
It is derived from the probability that the system’s trajectory $x(t)$ revisits a neighborhood around the specific system state $\zeta$.
To quantify the proximity of states, the negative logarithmic distance function between the trajectory and the target state $\zeta$ is defined as:
\begin{equation}
    g(x(t)) = -\log(\delta(x(t),\zeta)),
\end{equation}
where $\delta(x(t),\zeta)$ is the distance between states $x(t)$ and $\zeta$.
As $x(t)$ approaches $\zeta$, $\delta(x(t),\zeta)$ tends to zero, leading to a larger $g(x(t))$.
The neighborhood of $\zeta$ is defined by selecting a threshold $g_q$ based on a high quantile $q$ (e.g., 0.98) of $g(x(t))$. The exceedance of $\zeta$, denoted by $u(\zeta)$, is then defined by:
\begin{equation}
    u(\zeta)=g(x(t)) - g_q, \quad \forall g(x(t))>g_q.
\end{equation}

The cumulative probability distribution $\mathrm{F}(u(\zeta)$ fits the exponential form of the generalized Pareto distribution (GPD):
\begin{equation}\label{eq:d}
    \mathrm{F}[u(\zeta)]\simeq \exp{\left[-\frac{u(\zeta)}{\sigma(\zeta)}\right]},
\end{equation}
where $\sigma$ is the scale parameter of the distribution, depending on the specific reference state $\zeta$. 
For each $\zeta$, the instantaneous dimension $d$ can be calculated by $d(\zeta) = 1/\sigma(\zeta)$ according to its definition~\cite{datseris2023estimating}.
The range of $d$ lies in $(0,\infty)$, with the larger values indicating higher dynamical complexity or greater degrees of freedom. \\

\noindent \textbf{Inverse persistence $\theta$.} 
Inverse persistence $\theta$ measures the persistence of a trajectory, characterizing how long a system tends to remain in the vicinity of a specific state $\zeta$ before leaving. 
This definition is closely related to the extremal index (EI), which was initially introduced to characterize the clustering of extreme events in EVT~\cite{smith1994estimating}. A higher EI indicates the extremes tend to occur more independently with fewer temporal clusters.

In the computation of $\theta$, the extremes are defined as the states extremely close to the specific reference state $\zeta$, namely those satisfying $g(x(t))>g_q$. 
When such extreme events occur more independently in time, they are less temporally clustered, resulting in shorter residence times near $\zeta$.
Shorter residence times indicate lower persistence, which corresponds to higher values of $\theta$. 
By definition, $\theta$ ranges between 0 and 1 \cite{faranda2024statistical}. \\

\noindent \textbf{DI computation using ML output.}
We treat the training dataset as the reference attractor, against which the DI of the model forecasts and their corresponding targets are computed, as illustrated in Supplementary Fig.~\ref{si-fig:calculate_di} in Supplementary Information section~\ref{si:method_illustration} .

DI can be calculated using either raw or normalized data. 
In this work, we use normalized data for the three canonical datasets, and raw (original scale) data for the weather dataset, as the latter provides more intuitive and physically meaningful interpretations for real-world tasks.
A comparison between the two approaches is provided in Supplementary Information section~\ref{normalization}, demonstrating that the proposed method remains robust under both conditions. \\

\noindent \textbf{Generalized Pareto distribution fit test.}
Fitting the GPD is a key step in calculating dynamical indices. 
A key factor influencing the quality of the fit is the number of neighbors -- equivalently, the choice of quantile $q$ -- as discussed in section~\ref{sec:methods-indices}.

In this work, we define the top 2\% of closest states (i.e. $q$=0.98) as dynamical neighbors, following the approach of~\cite{faranda2017dynamical}. 
Supplementary Fig.~\ref{si-fig:FigS11_q_sensitivity} presents a comparison of the goodness of fit across different quantile choices using the Chi-squared test, where a $p$-value greater than 0.05 indicates a statistically acceptable fit
As expected, the fit improves with increasing quantile (i.e., larger $q$), as a smaller and more similar subset of neighbors is selected, resulting in a distribution that more closely matches the GPD.

Supplementary Fig.~\ref{si-fig:FigS10_chi2} shows the goodness of fit for single-step machine learning forecasts at $q = 0.98$, evaluated across different lead times
In most cases, the fit deteriorates as the lead time increases, indicating that fewer forecasted states retain valid dynamical neighbors, likely due to higher forecast errors, which is consistent with the results in section~\ref{sec:results-direct}.

\subsection{Dynamical error metrics}
\label{sec:method_metrics}
We introduce L1- and L2-based dynamical error metrics -- MAE${_d}$, MAE${_\theta}$, MSE${_d}$, and MSE${_\theta}$ -- along with their normalized counterparts: NMAE${_d}$, NMAE${_\theta}$, NMSE${_d}$, and NMSE${_\theta}$. 
These metrics are formulated to mirror the definitions of standard error metrics (i.e., MAE and MSE), while specifically quantifying the discrepancies in dynamical properties between forecasts and target states.

In addition, we propose the dynamical indices difference (DID) as a sample-wise diagnostic metric for ML forecasts.
Unlike traditional error metrics, DID retains the sign of the forecasted $d$ and $\theta$ values relative to the ground truth, enabling the identification of under- or overestimation in dynamical indices.
% The definition is provided in the last row of Table~\ref{tab:define_metrics}.

The definitions of the standard and dynamical metrics used in this work are presented in Eq.~\eqref{eq:mse_metrics}--\eqref{eq:did_metrics}. 
In these definitions, $y$ denotes the observed quantity, $N_t$ is the number of time samples, $N_s$ is the number of spatial locations, $\mu$ represents the mean value, and $\sigma$ the standard deviation, used for normalization. In addition, quantities denoted by $\hat{g}$ are ML forecasts, while quantities denoted without the hat symbol -- i.e., $g$ -- represent the true values.
\begin{subequations} \label{eq:mse_metrics}
\begin{align}
% MSE
\mathrm{MSE} &= \frac{1}{N_t\,N_s}\sum_{i=1}^{N_t}\sum_{j=1}^{N_s} \Bigl(\hat{y}_{ij}-y_{ij}\Bigr)^2,\\
\mathrm{MSE}_d &= \frac{1}{N_t}\sum_{i=1}^{N_t} \Bigl(\hat{d}_i - d_i\Bigr)^2,\\
\mathrm{MSE}_\theta &= \frac{1}{N_t}\sum_{i=1}^{N_t} \Bigl(\hat{\theta}_i - \theta_i\Bigr)^2;
\end{align}
\end{subequations}

\begin{subequations} \label{eq:nmse_metrics}
\begin{align}
\mathrm{NMSE} &= \frac{1}{N_t\,N_s}\frac{\sum_{i=1}^{N_t}\sum_{j=1}^{N_s} \Bigl(\hat{y}_{ij}-y_{ij}\Bigr)^2}{\sigma_y^2,} \\
\mathrm{NMSE}_d &= \frac{1}{N_t}\frac{\sum_{i=1}^{N_t}\Bigl(\hat{d}_i-d_i\Bigr)^2}{\sigma_d^2},\\
\mathrm{NMSE}_\theta &= \frac{1}{N_t}\frac{\sum_{i=1}^{N_t}\Bigl(\hat{\theta}_i-\theta_i\Bigr)^2}{\sigma_\theta^2};
\end{align}
\end{subequations}

\begin{subequations} \label{eq:mae_metrics}
\begin{align}
\mathrm{MAE} &= \frac{1} {N_t\,N_s}\sum_{i=1}^{N_t}\sum_{j=1}^{N_s} \|\hat{y}_{ij}-y_{ij}\|_1, \\
\mathrm{MAE}_d &= \frac{1}{N_t}\sum_{i=1}^{N_t}\|\hat{d}_i-d_i\|_1,\\
\mathrm{MAE}_\theta &= \frac{1}{N_t}\sum_{i=1}^{N_t}\|\hat{\theta}_i-\theta_i\|_1;
\end{align}
\end{subequations}

\begin{subequations} \label{eq:nmae_metrics}
\begin{align}
\mathrm{NMAE} &= \frac{1}{N_t\,N_s}\frac{\sum_{i=1}^{N_t}\sum_{j=1}^{N_s} \|\hat{y}_{ij}-y_{ij}\|_1}{\mu_y}, \\
\mathrm{NMAE}_d &= \frac{1}{N_t}\frac{\sum_{i=1}^{N_t}\|\hat{d}_i-d_i\|_1}{\mu_d},\\
\mathrm{NMAE}_\theta &= \frac{1}{N_t}\frac{\sum_{i=1}^{N_t}\|\hat{\theta}_i-\theta_i\|_1}{\mu_\theta};
\end{align}
\end{subequations}

\begin{subequations} \label{eq:did_metrics}
\begin{align}
\mathrm{DID}_d &= \hat{d} - d,\\
\mathrm{DID}_\theta &= \hat{\theta} - \theta.
\end{align}
\end{subequations}

\subsection{Characteristic time scales}
\label{sec:time_scale}
In this work, the lead time and recursive time of the three canonical datasets are normalized by their characteristic time scales: LT for Lorenz and KS, and TU for KF.\\

\noindent\textbf{Lyapunov time (LT).}
Lyapunov exponents quantify the rate at which infinitesimally close trajectories diverge over time. The largest Lyapunov exponent serves as a measure of predictability: the larger the exponent, the lower the predictability of the system.
The Lyapunov time (LT) is defined as the inverse of the largest Lyapunov exponent. 
It represents the timescale over which the distance between two initially close trajectories increases by a factor of $e$.

Specifically, the largest Lyapunov exponent of the Lorenz 63 system is 0.906~\cite{wolf1986quantifying}, corresponding to an LT of approximately 1.1, or about 110 time steps.  
For KS system, the largest Lyapunov exponent is about 0.043~\cite{edson2019lyapunov}, leading to an LT of about 23 (92 time steps). \\

\noindent \textbf{Time unit (TU).}
Typically for Kolmogorov flow, the exact Lyapunov time is intractable. 
We instead use the period of forcing as the characteristic time unit (TU), which corresponds to 6 time steps.

\subsection{Wasserstein Distance}
Wasserstein distance (WD), also known as the optimal transport distance, measures the similarity between two probability distributions by calculating the minimum "cost" required to transform one distribution into the other. 
A smaller WD indicates a closer match between the two distributions. 
For two one-dimensional (1D) probability distributions, the Wasserstein distance can be computed as follows:

\begin{equation}
\label{eq:wd_1d}
    \mathrm{WD}_y = \underset{\gamma \in \Gamma(y_1,y_2)}{\inf}\int_{\mathbb{R}\times\mathbb{R}} \|x_1-x_2\|\mathrm{d}\Gamma(y_1,y_2).
\end{equation}
Here, $y_1$ and $y_2$ represent the probability distributions, with corresponding occurrences defined on $x_1$ and $x_2$, respectively. $\Gamma(y_1,y_2)$ denotes the joint distribution defined over the region $\mathbb{R}\times\mathbb{R}$.

To calculate the overall WD for $d$ and $\theta$, we first normalize their distributions by dividing by the respective sum, resulting in two 1D probability distributions.
Next, the individual WDs for $d$ and $\theta$ are computed using Eq.~\eqref{eq:wd_1d}, respectively.
Finally, the overall WD is calculated as follows:
\begin{equation}
\label{eq:wd}
    \mathrm{WD} = \sqrt{(\mathrm{WD}_d)^2+(\mathrm{WD}_\theta)^2}.
\end{equation}

\subsection{Datasets and machine learning setup}
\label{sec:data_and_method}

\subsubsection{Datasets}
\noindent \textbf{Lorenz 63 system (Lorenz).}
Lorenz system, mathematically described in Eq.~\eqref{eqn:lorenz}, is built as a simplified model of atmospheric convection ~\cite{lorenz1963deterministic}. 
Its simplicity and chaotic nature make it a popular benchmark for forecasting tasks. 
In this work, we generated data with $\sigma = 10$, $\rho = 28$, $\beta = 2.667$ with a time step of $dt = 0.01$ for a total of  1,000,000 time steps. 
% The largest Lyapunov exponent of the Lorenz 63 system is 0.906, corresponding to an LT of approximately 1.1, or about 110 time steps.  
%
\begin{align}
    \frac{dx}{dt} & = \sigma(y-x) \nonumber, \\
    \frac{dy}{dt} &= \rho(x-z)-y,  \label{eqn:lorenz}\\
    \frac{dz}{dt} &= xy-\beta z. \nonumber 
\end{align}
\noindent \textbf{Kuramoto–Sivashinsky equation (KS).}
The 1D KS equation is a fourth-order partial differential equation (PDE) given by Eq.~\eqref{eqn:ks}, where $u$ represents the observable, $t$ is time, and $x$ is the spatial position defined on the interval $[0,L)$. 
This equation serves as a classic spatiotemporal model, exhibiting rich dynamics and chaotic behavior.
\begin{equation}\label{eqn:ks}
    % u_t + u u_x + u_{xx} + u_{xxxx} = 0.
    \frac{\partial u}{\partial t} + u \frac{\partial u}{\partial x} + \frac{\partial^2 u}{\partial x^2} + \frac{\partial^4 u}{\partial x^4} =0.
\end{equation}
Following the configuration in~\cite{pathak2018model}, we set $L$=22 for numerical simulation, with the solution discretized over 64 equally spaced grid points in the domain $[0,L)$. 
The data was generated using a small time step $dt=0.01$s for numerical stability, with a total of 2,500,000 simulated time steps. 
We then discarded the initial 10,000 steps to allow the trajectory to stabilize, and downsampled the data to $dt=0.25$s to introduce greater variability and make the forecasting task more challenging. \\

\noindent \textbf{Kolmogorov flow (KF).}
Kolmogorov flow (KF) is a two-dimensional shear flow governed by the Navier–Stokes equations, driven by a spatially periodic Kolmogorov forcing, as shown in Eq.~\eqref{eqn:kflow}:
\begin{align}\label{eqn:kflow}
    \frac{\partial u}{\partial t} +u \cdot \nabla u &= - \nabla p + \frac{1}{Re} \nabla^2 u + \mathrm{sin}(ny)\hat{x} \\
    \nabla u & = 0.  \nonumber 
\end{align}
We adopted the same configuration as in~\cite{lin2023online}, using vorticity, defined as $\omega = \nabla \times u$, as the target variable in the regression task. 
The simulation was performed with a Reynolds number $Re=14.4$ and forcing mode $n=2$, a regime in which the flow exhibits intermittent bursting behavior. \\
% Since the exact Lyapunov time for the Kolmogorov flow is intractable, we instead use the period of forcing as the characteristic time scale, which corresponds to 6 time steps.

\noindent \textbf{Weather dataset.}
We use the fifth-generation ECMWF atmospheric reanalysis (ERA5) dataset~\cite{hersbach2020era5} as the ground truth dataset.
The data is provided with a spatial resolution of $0.25^\circ$ in both latitude and longitude, and a temporal resolution of 6 hours.
Our analysis focuses on the Indochina region, defined as $70^\circ$E–$140^\circ$E and $10^\circ$S–$30^\circ$N. 
We examine the mean sea level pressure (SLP), a key dynamical variable associated with large-scale atmospheric oscillations.

The weather forecasts analyzed in this study are generated by two state-of-the-art machine learning models: Pangu-Weather~\cite{bi2023accurate} and GraphCast~\cite{lam2023learning}.
All the forecasts are publicly available in WeatherBench2~\cite{rasp2023weatherbench}.
We utilize forecasts initialized at 12\,am each day, with a temporal resolution of 6\,h, meaning the first forecast time is 6\,am. 
The dataset covers the period from January 1, 2020, to December 31, 2020. 
All forecasts are provided at the same spatial resolution as ERA5 data.

\subsubsection{Preprocess}
\noindent \textbf{Data split.}
Each canonical dataset was divided into training, validation, and test sets. Specifically, the first 70\% of the data was used for training, the subsequent 15\% for validation, and the final 15\% for testing. \\

\noindent \textbf{Normalization.}
Z-score normalization was applied to all three canonical datasets, as defined in Eq.~\eqref{eqn:zscore}, using the mean and standard deviation computed from the training set only. Specifically, $\mu_{u_{train}}$ denotes the mean value of observable $u$ in training set, and $\sigma_{u_{train}}$ represents the standard deviation. 
 
\begin{equation}\label{eqn:zscore}
    u_{norm} = \frac{u-\mu_{u_{train}}}{\sigma_{u_{train}}}. 
\end{equation} \\

\subsubsection{Machine learning setup}
\noindent \textbf{Task configuration.} 
\label{sec:ml_task}
A typical machine learning forecasting task involves learning a mapping between input and target states, as illustrated in Supplementary Inforamtion Fig.~\ref{si-fig:task}, located in Supplementary Information section~\ref{si:method_illustration}.
Here, $m$ represents the input length, $n$ refers to the lead time steps -- where $n=1$ corresponds to forecasting the next time step -- and $l$ denotes the output length.

In this work, we adopt both single-step ($l=1$) and recursive ($l>1$) prediction approaches. 
For the single-step forecast, we evaluate the model performance across various input lengths ($m = 1, 3, 10, 20, 40$) and lead time steps ($n = 1, 40, 80, 160, 240$).
For the recursive approach, the model output is iteratively concatenated with the input, and the most recent sequence is fed back into the model to generate predictions over a longer time horizon. We conducted 1100, 279, and 60 steps of recursive forecast fr Lorenz, KS, and KF, respectively, corresponding to 10 LT, 3 LT, and 10 TU. \\

%
% \begin{equation}
%     f_{p}: \mathbb{R}^{s \times m} \rightarrow \mathbb{R}^{s \times l}
% \end{equation}
%

\noindent \textbf{ML models architectures.} 
In this study, we employ commonly used model architectures for forecasting tasks, including Convolutional Neural Networks (CNN), Long Short-Term Memory Networks (LSTM)\cite{sak2014long}, Transformers\cite{vaswani2017attention,dosovitskiy2020image}, and Graph Neural Networks (GNN)~\cite{kipf2016semi}. 
Specifically for weather, the Pangu-Weather~\cite{bi2023accurate} model is based on Transformer, and GraphCast~\cite{lam2023learning} leverages a GNN architecture.\\

\noindent \textbf{Choice of hyperparameters.} 
To ensure the reliability of the results and prevent the conclusion from being biased due to model deficiency, we optimized the hyperparameters for each model architecture and different input lengths. 
The hyperparameter optimization was performed using TPE (Tree-structured Parzen Estimator) algorithm within the open-source package Optuna~\cite{akiba2019optuna}. \\

\noindent \textbf{Training strategy.} 
All our models were trained to minimize the MSE of the first step forecast, as defined in Eq.~\eqref{eq:mse_metrics}a. The early stopping strategy~\cite{yao2007early} is adopted to prevent severe overfitting. \\

\noindent \textbf{Standard ML metrics.}
The standard ML metrics used in this work are shown in subequation (a) in Eqs.~\eqref{eq:mse_metrics}--\eqref{eq:nmae_metrics}. 
Specifically, latitude-weighted RMSE ($\mathrm{RMSE}_{Lat}$) is used in evaluating weather forecasts, defined as below:
\begin{subequations}\label{eq:rmse_lat}
    \begin{align}
        \mathrm{RMSE}_{Lat} &= \frac{1}{N}\sum_{i=1}^N \sqrt{\sum_{j=1}^{N_{Lat}} \sum_{k=1}^{N_{Lon}} W_j \left(\hat{y}_{ijk}-y_{ijk}\right)^2}, \\[1.0em]
        W_j &= \frac{\cos\left(Lat(j)\right)}{\frac{1}{N_{Lat}}\sum_{j=1}^{N_{Lat}} \cos\left(Lat(j)\right)},
    \end{align}
\end{subequations}
where $n$ is the number of samples, $\hat{y}_i$ is the predicted value for $i_{th}$ sample, $y_i$ be the true value, $N_{Lat}$,$N_{Lon}$ is the spatial dimension along latitude and longitude respectively. $\bar{y}=\frac{1}{n}\sum_i^n y_i$ is the mean value of truth.
$\mathrm{RMSE}_{Lat}$ is always positive, with a lower value indicating better performance. 

%TC:endignore

\subsection{Code availability}
All the relevant codes for this study are provided in a public repository \url{https://github.com/MathEXLab/DyEM}.

\backmatter

%%%%%%%%%%%%%%%%%%%
% EXTENDED DATA
%%%%%%%%%%%%%%%%%%%
\newpage
\section*{Extended Data}\label{sec:extended_data}
\renewcommand{\figurename}{Extended Data Fig.}
\renewcommand{\thefigure}{\arabic{figure}}
\setcounter{figure}{0}

\renewcommand{\tablename}{Extended Data Tab.}
\renewcommand{\thetable}{\arabic{table}}
\setcounter{table}{0}

%% TABLE MSE
%
\begin{table}[htpb]
\caption{\textbf{MSE for ML models across the datasets} (input length $m$ = 3, 1st step forecast). The best values are denoted with an underline; a smaller value indicates better performance for all metrics.}
\label{tab:ml_metrics_mse}
\begin{tabular}{cccccc}
\toprule
\textbf{Dataset}                 & \multicolumn{1}{l|}{\textbf{Metrics}}                & \textbf{Transformer}          & \textbf{LSTM}                 & \textbf{CNN}                  & \textbf{GNN} \\ 
\midrule
\multirow{10}{*}{\textbf{Lorenz}} & \multicolumn{1}{l|}{MSE (best)}            &8.69e-7            & \textbf{\underline{4.1e-7}}               & 7.84e-7            & 6.32e-7       \\
                        & \multicolumn{1}{l|}{MSE (mean)}     & 7.07e-6 & 1.02e-6  &8.67e-7  & \textbf{\underline{7.39e-7}}        \\
                        & \multicolumn{1}{l|}{MSE (std)}     & 6.57e-6 & 4.82e-7  &\textbf{\underline{1.10e-7}} &  9.40e-8 \\
                        \cline{2-6}
                        & \multicolumn{1}{l|}{MSE$_d$ (best)}           & 8.38e-4           & 7.10e-4         & 7.26e-4         &   \textbf{\underline{7.03e-4}}       \\
                        & \multicolumn{1}{l|}{MSE$_d$ (mean)}     & 1.11e-3 &8.79e-4& \textbf{\underline{5.88e-4}}  & 7.47e-4 \\
                        & \multicolumn{1}{l|}{MSE$_d$ (std)}     & 2.35e-4  &1.52e-4 & 2.43e-05  & \textbf{\underline{9.74e-6}} \\
                        \cline{2-6}
                        & \multicolumn{1}{l|}{MSE$_\theta$ (best)}     & \textbf{\underline{9.05e-5}}          & 9.14e-5          &9.52e-5        & 9.08e-5   \\
                        & \multicolumn{1}{l|}{MSE$_\theta$ (mean)} &  1.24e-4  & 9.55e-05  & 9.90e-5   &  \textbf{\underline{9.21e-5}}\\ 
                        & \multicolumn{1}{l|}{MSE$_\theta$ (std)} & 3.44e-5 & 3.59e-06  &  5.10e-6     &  \textbf{\underline{1.90e-6}} \\ 
                        \cline{1-6}
\multirow{10}{*}{\textbf{KS}} & \multicolumn{1}{l|}{MSE (best)}  & \textbf{\underline{1.35e-6}}     &1.92e-5   &6.61e-5     &9.45e-6   \\
                        & \multicolumn{1}{l|}{MSE (mean)}       & \textbf{\underline{2.95e-6 }}     &2.15e-5 & 8.13e-5   &  7.80e-5   \\
                        & \multicolumn{1}{l|}{MSE (std)}       & 1.90e-6     &3.29e-6 &  1.03e-5    &   \textbf{\underline{1.84e-6}}   \\
                        \cline{2-6}
                        & \multicolumn{1}{l|}{MSE$_d$ (best)}   & \textbf{\underline{4.47e-5}}     & 2.57e-4  & 6.84e-4   &8.51e-5   \\
                        & \multicolumn{1}{l|}{MSE$_d$ (mean)}     & \textbf{\underline{6.59e-5}}     &   3.06e-4  &  8.18e-4  & 9.09e-5         \\
                        & \multicolumn{1}{l|}{MSE$_d$ (std)}     & \textbf{\underline{ 2.10e-5}}     &   4.53e-5    &  1.23e-4  &  5.10e-6        \\
                        \cline{2-6}
                        & \multicolumn{1}{l|}{MSE$_\theta$ (best)}     & \textbf{\underline{3.93e-7}}     & 5.25e-7   & 6.97e-7   & 4.27e-7   \\
                        & \multicolumn{1}{l|}{MSE$_\theta$ (mean)} & \textbf{\underline{4.17e-7}}    &5.61e-7  &  7.32e-7    & 4.37e-7 \\ 
                        & \multicolumn{1}{l|}{MSE$_\theta$ (std)} & 2.30e-8   & 3.09e-8   & 3.12e-8  & \textbf{\underline{1.24e-8 }}\\ 
                        \cline{1-6}
\multirow{10}{*}{\textbf{KF}} & \multicolumn{1}{l|}{MSE (best)}   &  6.74e-3     &  \textbf{\underline{3.67e-4}}   &  8.46e-4      &  6.82e-3    \\
                        & \multicolumn{1}{l|}{MSE (mean)}  & 6.97e-3     & 8.00e-4    & \textbf{\underline{1.19e-4}}      & 7.46e-3 \\
                        & \multicolumn{1}{l|}{MSE (std)}    & \textbf{\underline{2.07e-4}}     &  7.51e-4     & 5.88e-4     & 1.09e-3 \\
                        \cline{2-6}
                        & \multicolumn{1}{l|}{MSE$_d$ (best)} & 3.66e-1 &\textbf{\underline{1.12e-2}}      &  2.26e-2       & 2.71e-1      \\
                        & \multicolumn{1}{l|}{MSE$_d$ (mean)}     & 4.13e-1& \textbf{\underline{1.17e-2}}   & 3.40e-2      & 3.17e-1 \\
                        & \multicolumn{1}{l|}{MSE$_d$ (std)}     & 4.06e-2 &  \textbf{\underline{9.00e-3}}   & 1.36e-2      & 7.15e-2  \\
                        \cline{2-6}
                        & \multicolumn{1}{l|}{MSE$_\theta$ (best)}     & 3.49e-5           &8.94e-6             &\textbf{\underline{7.64e-6}}       &  2.67e-9          \\
                        & \multicolumn{1}{l|}{MSE$_\theta$ (mean)} & 3.62e-5  & \textbf{\underline{1.03e-5}}     &  1.10e-5    &  2.84e-5     \\ 
                        & \multicolumn{1}{l|}{MSE$_\theta$ (std)} & \textbf{\underline{1.82e-6}}   & 2.30e-6   &  3.08e-6    &  2.69e-6 \\
\bottomrule
\end{tabular}
\end{table}
%

%% DIRECT FORECASTS LONGER TIME LEADS
%
\begin{figure}[H]
    \centering
    \includegraphics[width=0.9\linewidth]{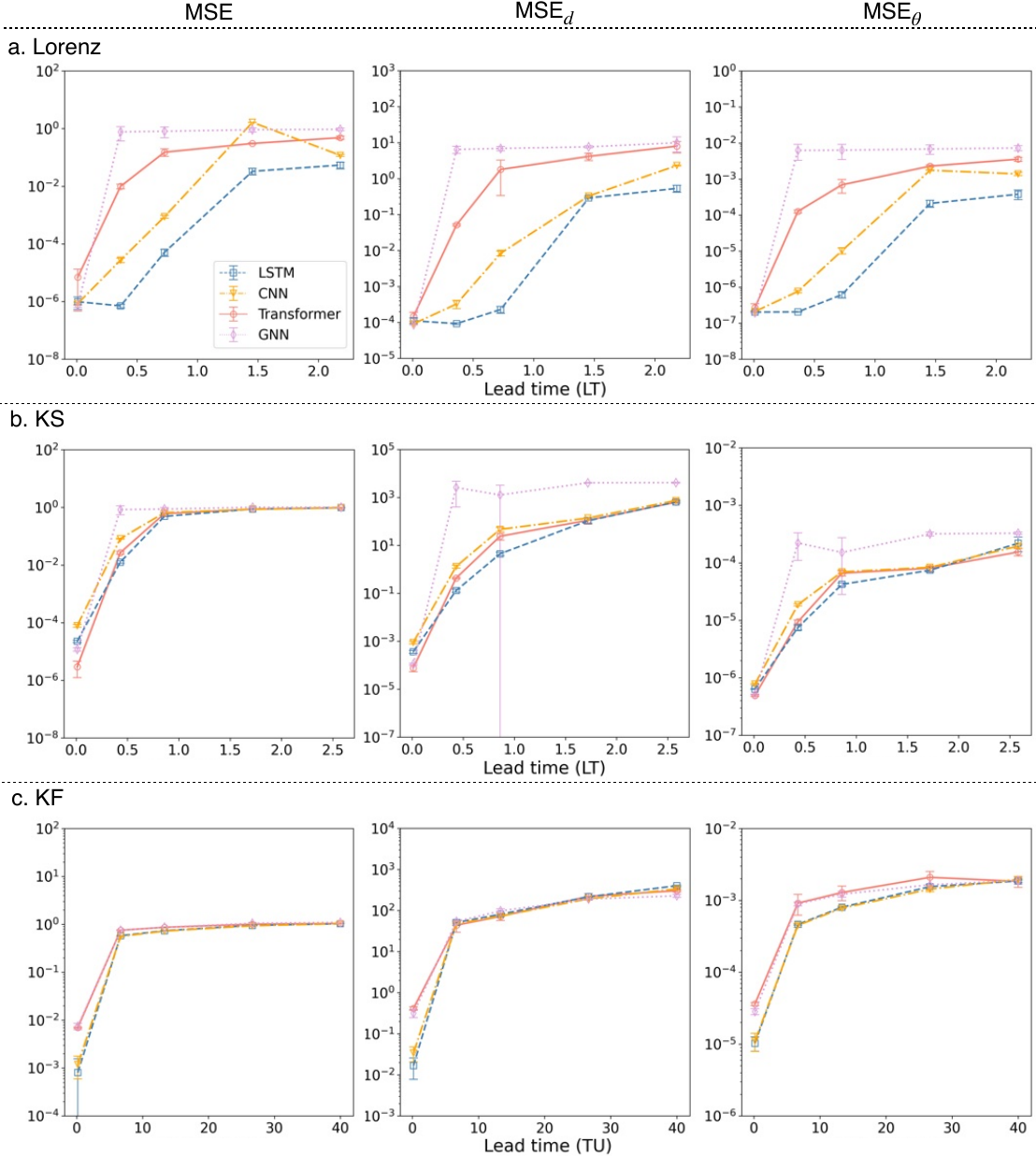}
    \caption{\textbf{Forecast error as a function of lead time.} Mean squared error (MSE) is shown for varying forecast lead times. Error bars denote the standard deviation computed from three independent runs, each initialized with different random model parameters.}
    \label{fig:FigS2_MSE}
\end{figure}
%

%% RECURSIVE ERRORS
%
\begin{figure}[H]
    \centering
    \includegraphics[width=0.75\linewidth]{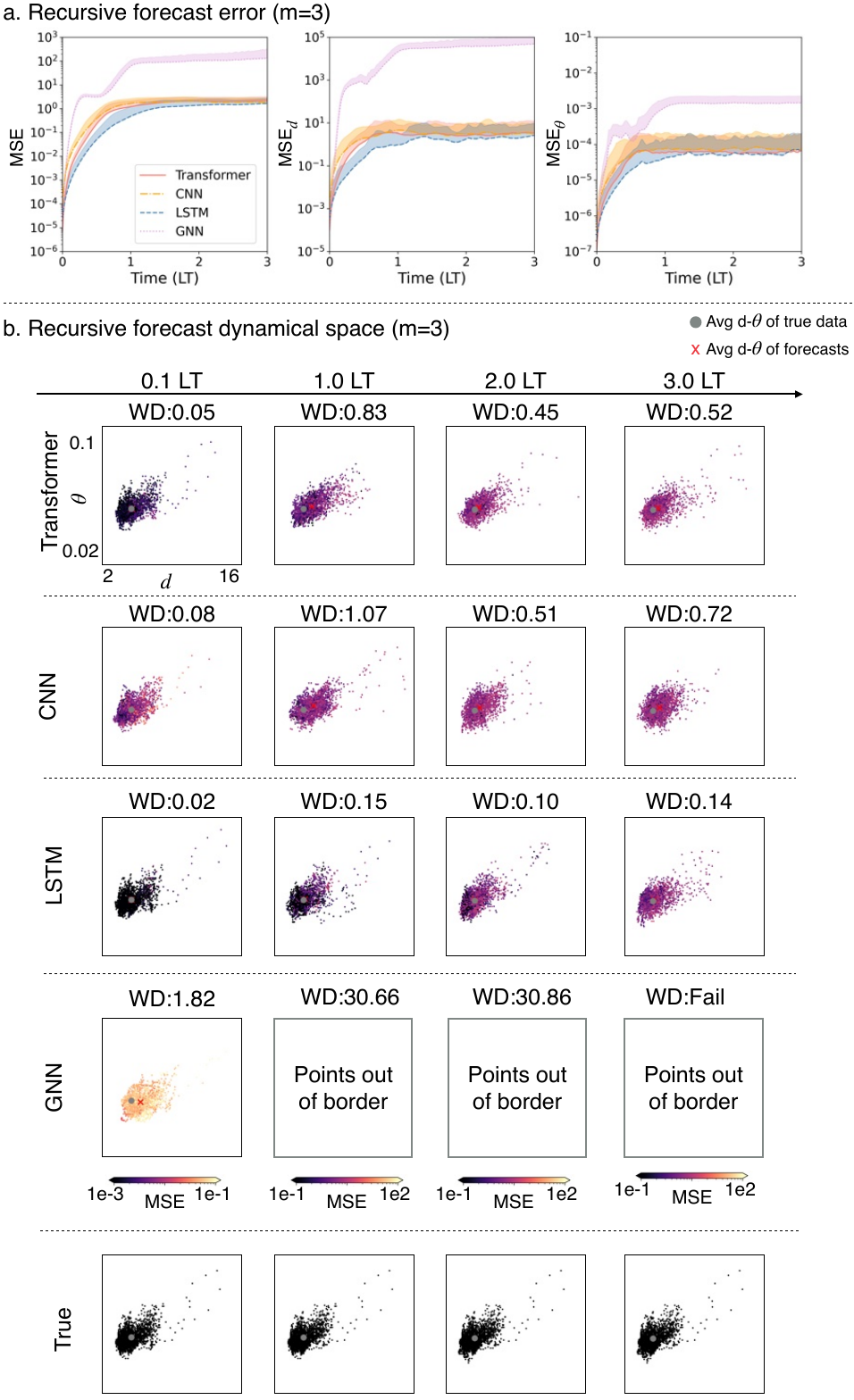}
    \caption{\textbf{Forecast errors and dynamical space for recursive runs on the KS dataset.} Panel (a) Mean squared error (MSE) as a function of Lyapunov Time (LT). Shaded regions represent the standard deviation computed across forecasts initialized from 2000 distinct initial states. Panel (b) Distribution of trajectories in the $d$–$\theta$ dynamical space at forecast times 0.1 LT, 1.0 LT, 2.0 LT, and 3.0 LT. Axes represent dynamical indices $d$ (horizontal) and $\theta$ (vertical), with consistent ranges across all subplots, as shown in the top-left panel. 
    Mean values of $d$ and $\theta$ indices and Wasserstein Distance (WD) between predicted and true distributions are annotated within each subplot.}
    \label{fig:Fig3_ks}
\end{figure}
\begin{figure}[H]
    \centering
    \includegraphics[width=0.75\linewidth]{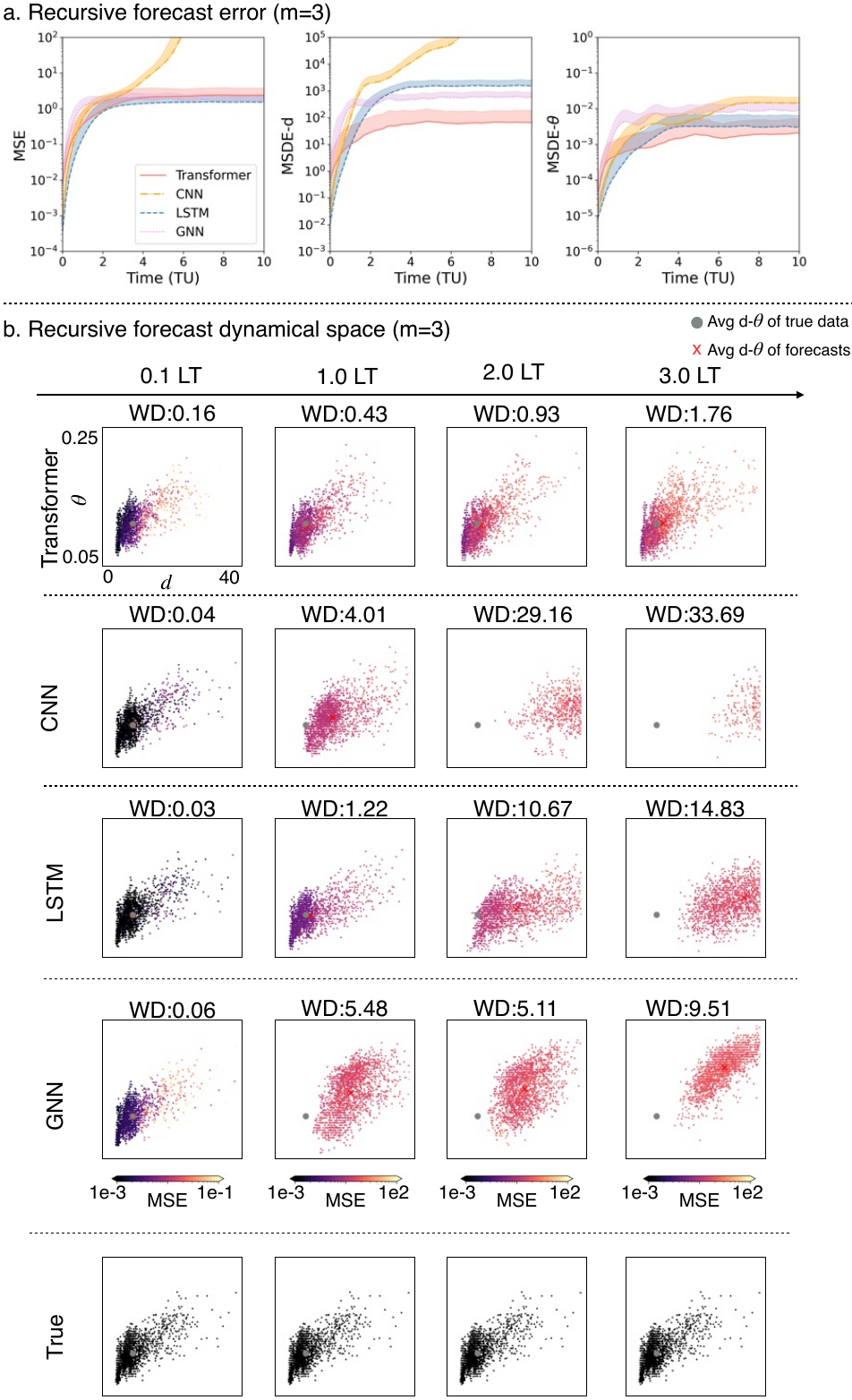}
    \caption{\textbf{Forecast errors and dynamical space for recursive runs on the KF dataset.} Panel (a) Mean squared error (MSE) as a function of characteristic Time Units (TU). Shaded regions represent the standard deviation computed across forecasts initialized from 2000 distinct initial states. Panel (b) Distribution of trajectories in the $d$–$\theta$ dynamical space at forecast times 0.1 TU, 1.0 TU, 2.0 TU, and 3.0 TU. Axes represent dynamical indices $d$ (horizontal) and $\theta$ (vertical), with consistent ranges across all subplots, as shown in the top-left panel. 
    Mean values of $d$ and $\theta$ indices and Wasserstein Distance (WD) between predicted and true distributions are annotated within each subplot.}
    \label{fig:Fig3_kf}
\end{figure}
\begin{figure}[H]
    \centering
    \includegraphics[width=0.9\linewidth]{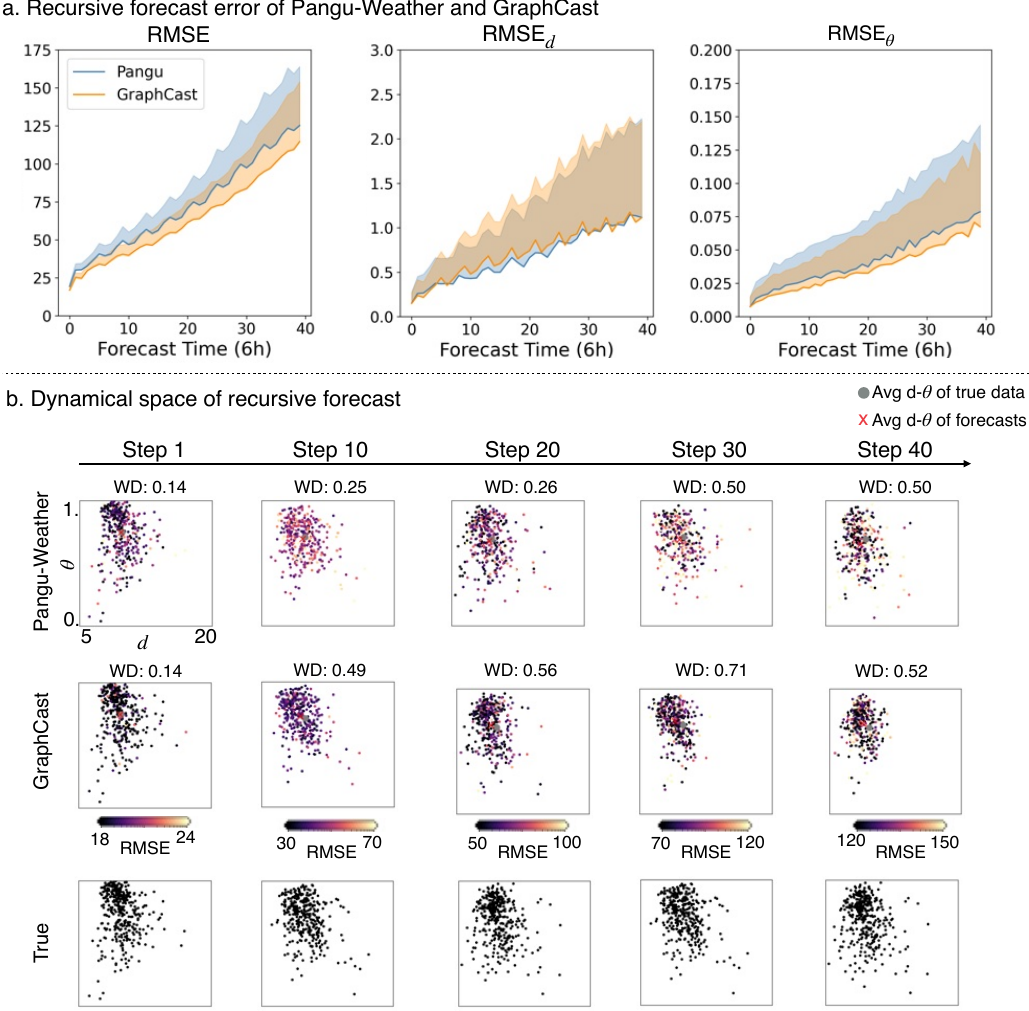}
    \caption{\textbf{Forecast errors and dynamical space for recursive runs on the weather dataset.} Panel (a) Mean squared error (MSE) as a function of recursive forecast lead time. Shaded regions indicate the standard deviation calculated from forecasts for the year 2020.  
    Panel (b) Distribution of forecasted states in the $d$–$\theta$ dynamical space at lead times of 1, 10, 20, 30, and 40 steps (each step corresponds to 6 hours). Axes represent dynamical indices $d$ (horizontal) and $\theta$ (vertical), with consistent ranges across all subplots, as shown in the top-left panel. 
    Mean values of $d$, $\theta$, and the Wasserstein Distance (WD) between forecasted and true distributions are annotated within each subplot.}
    \label{fig:fig3_weather}
\end{figure}

%% RECURSIVE DID
\begin{figure}[H]
    \centering
    \includegraphics[width=0.9\linewidth]{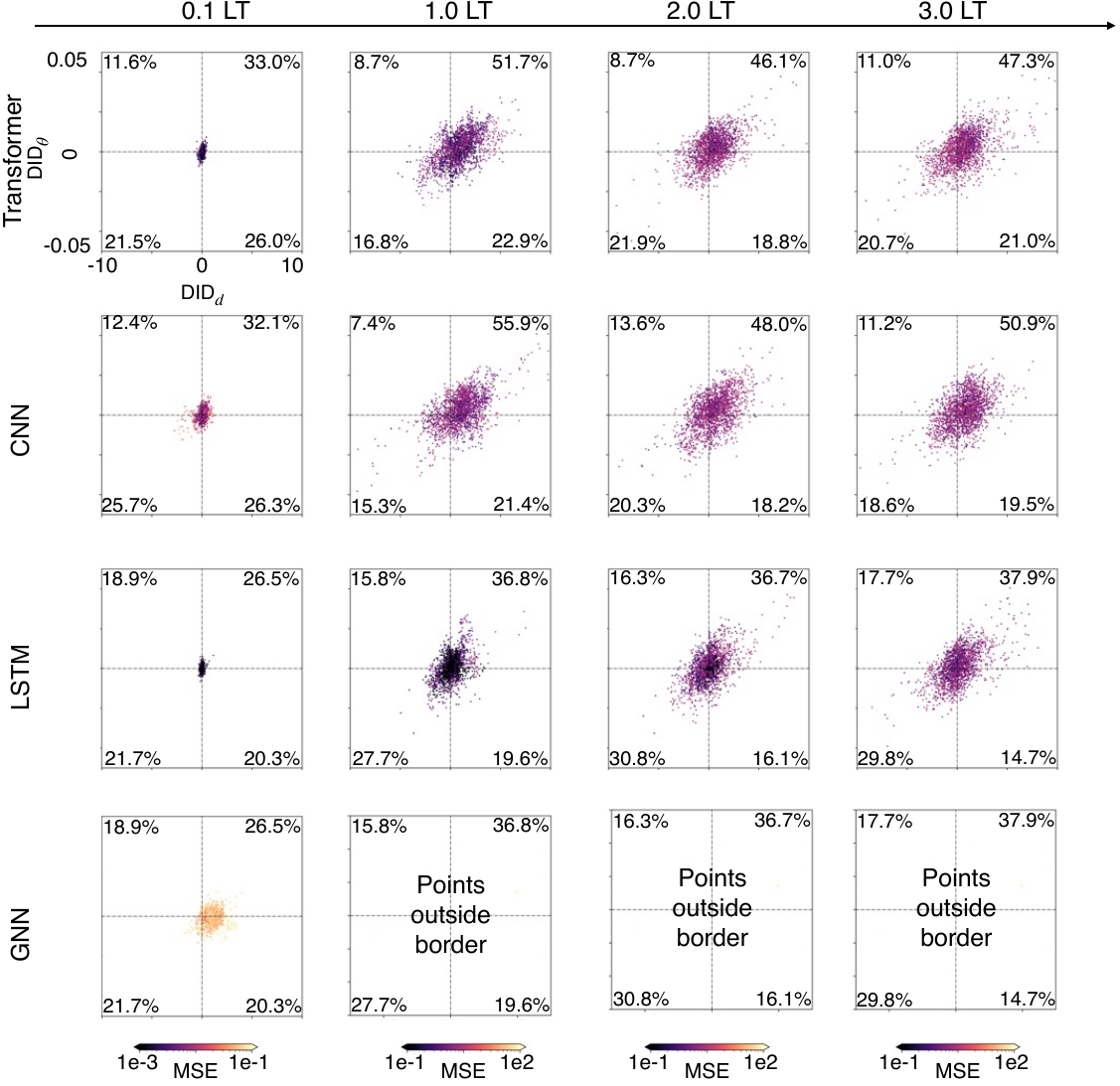}
    \caption{\textbf{DID of recursive forecast for KS.} The timeline at the top indicates the recursive forecast horizon. Each row of subplots corresponds to a distinct machine learning architecture. In each subplot, the $x$-axis represents $\mathrm{DID}_d$ and the $y$-axis represents $\mathrm{DID}_\theta$, with consistent ranges across all subplots, as shown in the top-left panel. The percentage of points falling into each quadrant is displayed at the corresponding corner. Points are colored according to their MSE.}
    \label{fig:recursive_did_ks}
\end{figure}

\begin{figure}[H]
    \centering
    \includegraphics[width=0.9\linewidth]{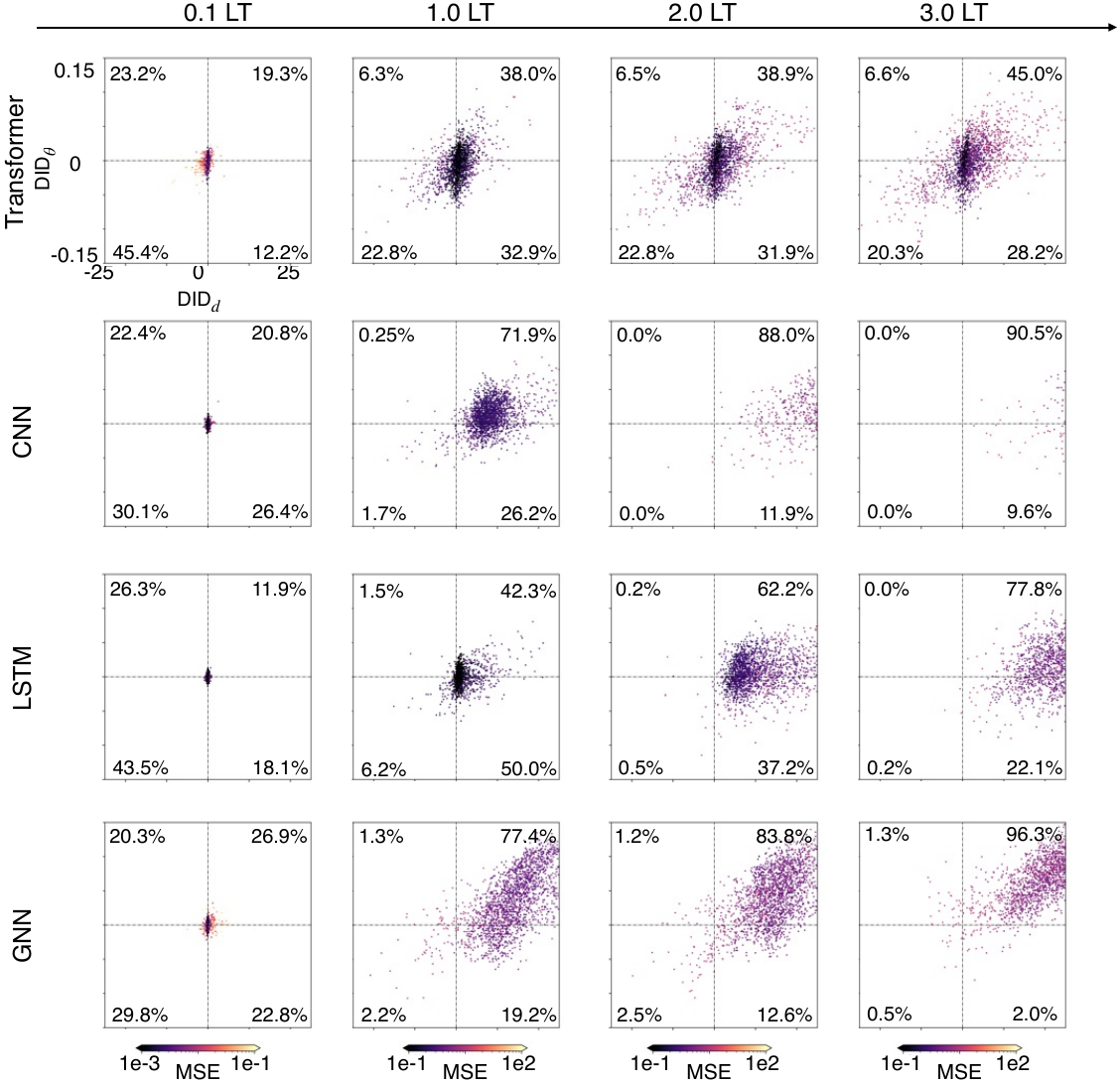}
    \caption{\textbf{DID of recursive forecast for KF.} The timeline at the top indicates the recursive forecast horizon. Each row of subplots corresponds to a distinct machine learning architecture. In each subplot, the $x$-axis represents $\mathrm{DID}_d$ and the $y$-axis represents $\mathrm{DID}_\theta$, with consistent ranges across all subplots, as shown in the top-left panel. The percentage of points falling into each quadrant is displayed at the corresponding corner. Points are colored according to their MSE.}
    \label{fig:recursive_did_kf}
\end{figure}

\begin{figure}[H]
    \centering
    \includegraphics[width=0.9\linewidth]{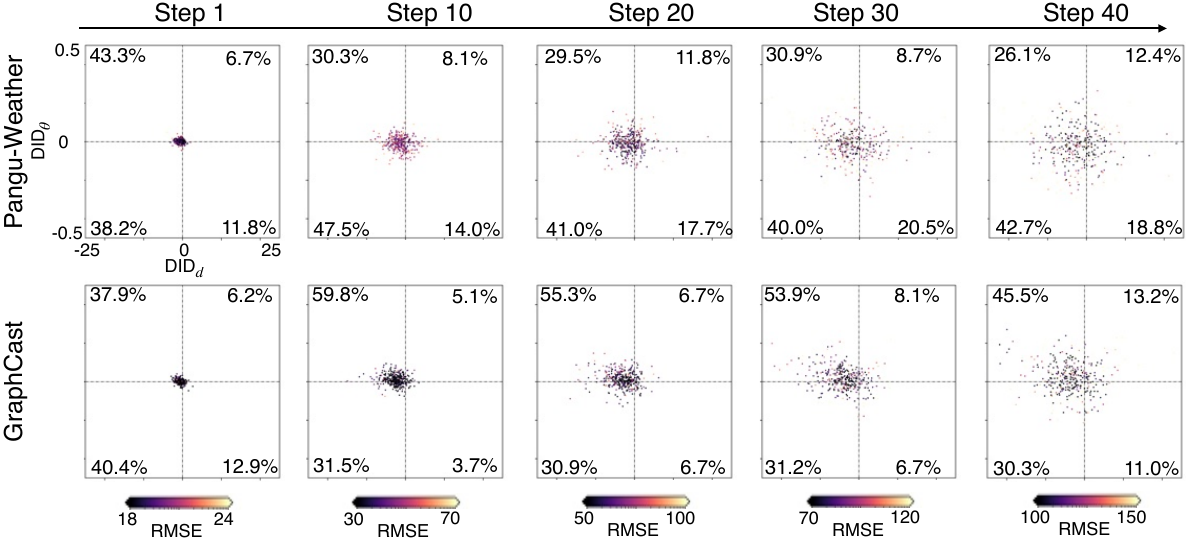}
    \caption{\textbf{DID of recursive forecast for weather dataset} The timeline at the top indicates the recursive forecast horizon. The two rows corresponds to Pangu-Weather and GraphCast, respectively. In each subplot, the $x$-axis represents $\mathrm{DID}_d$ and the $y$-axis represents $\mathrm{DID}_\theta$, with consistent ranges across all subplots, as shown in the top-left panel. The percentage of points falling into each quadrant is displayed at the corresponding corner. Points are colored according to their MSE.}
    \label{fig:recursive_did_weather}
\end{figure}

\bmhead{Acknowledgments}
This work is supported by Singapore’s MOE Tier 2 Grant MOE-T2EP50221-0006: ‘Prediction-to-Mitigation with Digital Twins of the Earth’s Weather’. We also thank Professor Nils Thuerey for fruitful discussions.

%% BIBLIOGRAPHY
\bibliography{sn-bibliography}

\renewcommand{\figurename}{Supplementary Fig.}
\captionsetup[table]{name=Supplementary Fig.}
\renewcommand{\thefigure}{S\arabic{figure}}
\setcounter{figure}{0}

\renewcommand{\tablename}{Supplementary Tab.}
\captionsetup[table]{name=Supplementary Tab.}
\renewcommand{\thetable}{S\arabic{table}}
\setcounter{table}{0}

\newpage 

\appendix
% Redefine subsection numbering for Supplementary Information
\renewcommand{\thesubsection}{S.\arabic{subsection}}

\section*{Supplementary Information}
\addcontentsline{toc}{section}{Supplementary Information}

\subsection{Normalized ML error results}\label{si:ml_stats}
Supplementary Tab.~\ref{si-tab:ml_metrics_nmse} provides a summary of the normalised MSE (NMSE) metrics across three benchmark datasets and four model architectures, using an input length of $m = 3$, including NMSE for $d$ and $\theta$, namely NMSE$_{d}$ and NMSE$_{\theta}$. 
Each experiment was repeated three times with different random initializations. 
For each model and dataset, the best, mean, and standard deviation (std) of the error across the three runs are reported. 
For all metrics, lower values indicate better performance.
\begin{table}[htpb]
\caption{\textbf{Overall NMSE for ML models across the datasets.} (input length $m$ = 3, 1st step forecast). 
The best values are denoted with an underline; a smaller value indicates better performance for all metrics.}
\label{si-tab:ml_metrics_nmse}
\begin{tabular}{cccccc}
\toprule
\textbf{Dataset}                 & \multicolumn{1}{l|}{\textbf{Metrics}}                & \textbf{Transformer}          & \textbf{LSTM}                 & \textbf{CNN}                  & \textbf{GNN} \\ 
\midrule
\multirow{10}{*}{\textbf{Lorenz}} & \multicolumn{1}{l|}{NMSE (best)}            &8.59e-07            & \textbf{\underline{4.77e-7}}               & 7.74e-7            & 6.14e-7       \\
                        & \multicolumn{1}{l|}{NMSE (mean)}     & 6.99e-6 & 9.94e-7  &8.60e-7  & \textbf{\underline{7.18e-7}}        \\
                        & \multicolumn{1}{l|}{NMSE (std)}     & 6.51e-6 & 4.67e-7  &\textbf{\underline{1.10e-7}} &  9.14e-8 \\
                        \cline{2-6}
                        & \multicolumn{1}{l|}{NMSE$_d$ (best)}           & 6.55e-4           & 5.50e-4         & 5.67e-4         &   \textbf{\underline{5.45e-4}}       \\
                        & \multicolumn{1}{l|}{NMSE$_d$ (mean)}     & 9.64e-4 &7.09e-4& 5.88e-4  & \textbf{\underline{5.53e-4}} \\
                        & \multicolumn{1}{l|}{NMSE$_d$ (std)}     & 2.72e-4  &1.46e-4 & 2.08e-05  & \textbf{\underline{7.39e-6}} \\
                        \cline{2-6}
                        & \multicolumn{1}{l|}{NMSE$_\theta$ (best)}     & \textbf{\underline{9.05e-5}}          & 9.14e-5          &9.52e-5        & 9.08e-5   \\
                        & \multicolumn{1}{l|}{NMSE$_\theta$ (mean)} &  1.24e-4  & 9.55e-05  & 9.90e-5   &  \textbf{\underline{9.21e-5}}\\ 
                        & \multicolumn{1}{l|}{NMSE$_\theta$ (std)} & 3.44e-5 & 3.59e-06  &  5.10e-6     &  \textbf{\underline{1.90e-6}} \\ 
                        \cline{1-6}
\multirow{10}{*}{\textbf{KS}} & \multicolumn{1}{l|}{NMSE (best)}  & \textbf{\underline{1.41e-6}}     &2.01e-5   &6.89e-5     &9.85e-6   \\
                        & \multicolumn{1}{l|}{NMSE (mean)}       & \textbf{\underline{3.07e-6 }}     &2.24e-5 & 8.13e-5   &  1.14e-5   \\
                        & \multicolumn{1}{l|}{NMSE (std)}       & 1.98e-6     &3.40e-6 &  1.07e-5    &   \textbf{\underline{1.90e-6}}   \\
                        \cline{2-6}
                        & \multicolumn{1}{l|}{NMSE$_d$ (best)}   & \textbf{\underline{2.98e-5}}     & 1.71e-4  & 4.56e-4   &5.68e-5   \\
                        & \multicolumn{1}{l|}{NMSE$_d$ (mean)}     & \textbf{\underline{4.39e-5}}     &   2.04e-4  &  5.45e-4  & 6.06e-5         \\
                        & \multicolumn{1}{l|}{NMSE$_d$ (std)}     & \textbf{\underline{ 1.40e-5}}     &   3.02e-5    &  8.21e-5  &  3.40e-6        \\
                        \cline{2-6}
                        & \multicolumn{1}{l|}{NMSE$_\theta$ (best)}     & \textbf{\underline{1.25e-2}}     & 1.66e-2   & 2.21e-2   & 1.35e-2   \\
                        & \multicolumn{1}{l|}{NMSE$_\theta$ (mean)} & \textbf{\underline{1.32e-2}}    &1.78e-2  &  2.32e-2    & 1.39e-2 \\ 
                        & \multicolumn{1}{l|}{NMSE$_\theta$ (std)} & 7.30e-4   & 9.82e-4   & 9.89e-4  & \textbf{\underline{3.93e-4 }}\\ 
                        \cline{1-6}
\multirow{10}{*}{\textbf{KF}} & \multicolumn{1}{l|}{NMSE (best)}   &  6.86e-3     &  \textbf{\underline{3.67e-4}}   &  8.50e-4      &  6.96e-3    \\
                        & \multicolumn{1}{l|}{NMSE (mean)}  & 7.07e-4     &  4.79e-4    & \textbf{\underline{1.19e-4}}      & 7.78e-3 \\
                        & \multicolumn{1}{l|}{NMSE (std)}    & 1.85e-4     &  \textbf{\underline{1.23e-4}}     & 5.9e-4     & 1.25e-3 \\
                        \cline{2-6}
                        & \multicolumn{1}{l|}{NMSE$_d$ (best)} & 1.49e-2 &\textbf{\underline{4.72e-4}}      &  9.21e-4       & 1.10e-2      \\
                        & \multicolumn{1}{l|}{NMSE$_d$ (mean)}     & 1.68e-2& \textbf{\underline{6.62e-4}}   & 1.39e-3      & 1.29e-2 \\
                        & \multicolumn{1}{l|}{NMSE$_d$ (std)}     & 1.66e-3 &  \textbf{\underline{8.90e-5}}   & 5.52e-4      & 2.92e-3  \\
                        \cline{2-6}
                        & \multicolumn{1}{l|}{NMSE$_\theta$ (best)}     & 6.42e-2           &1.65e-2             &\textbf{\underline{1.41e-2}}       &  4.93e-2          \\
                        & \multicolumn{1}{l|}{NMSE$_\theta$ (mean)} & 6.67e-2  & \textbf{\underline{1.65e-2}}     &  2.02e-2    &  5.23e-2     \\ 
                        & \multicolumn{1}{l|}{NMSE$_\theta$ (std)} & \textbf{\underline{3.35e-3}}   & 4.23e-3   &  5.67e-3    &  4.95e-3 \\
\bottomrule
\end{tabular}
\end{table}

\clearpage
\subsection{Other standard error metrics for direct forecasts}
\label{si:errors-vs-quantile}
\begin{figure}[H]
    \centering
    \includegraphics[width=0.95\linewidth]{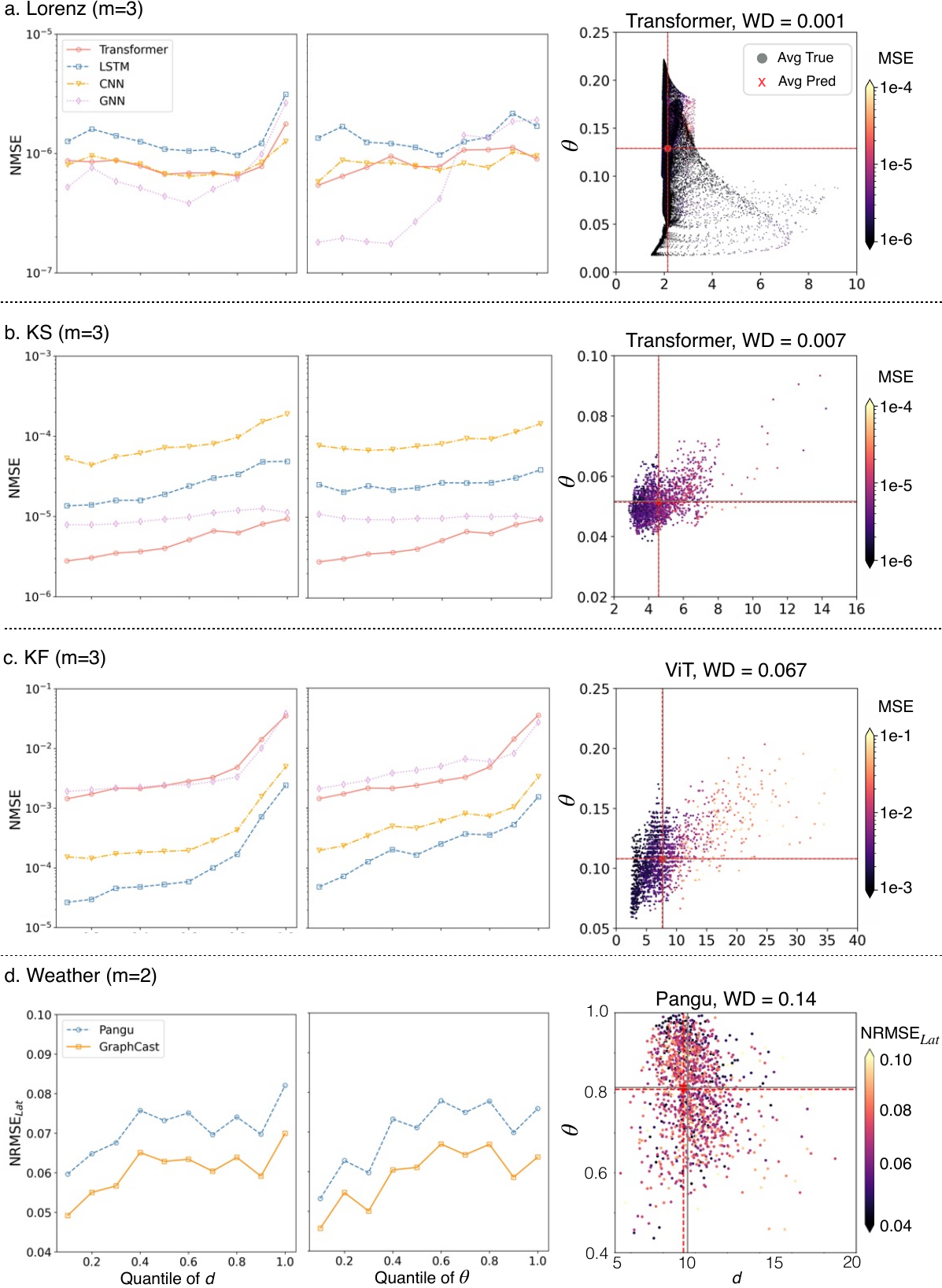}
    \caption{\textbf{Relationship between NMSE and dynamical indices for one step lead time.} Left and middle columns: the averaged forecast error for direct single-step prediction with a lead time of one step, over the quantile of $d$ (left) and $\theta$ (middle). The average dynamical indices of both true and predicted data are marked in the plots, along with the WD.}
    \label{si-fig:Fig2_NMSE}
\end{figure}

\begin{figure}[H]
    \centering
    \includegraphics[width=0.95\linewidth]{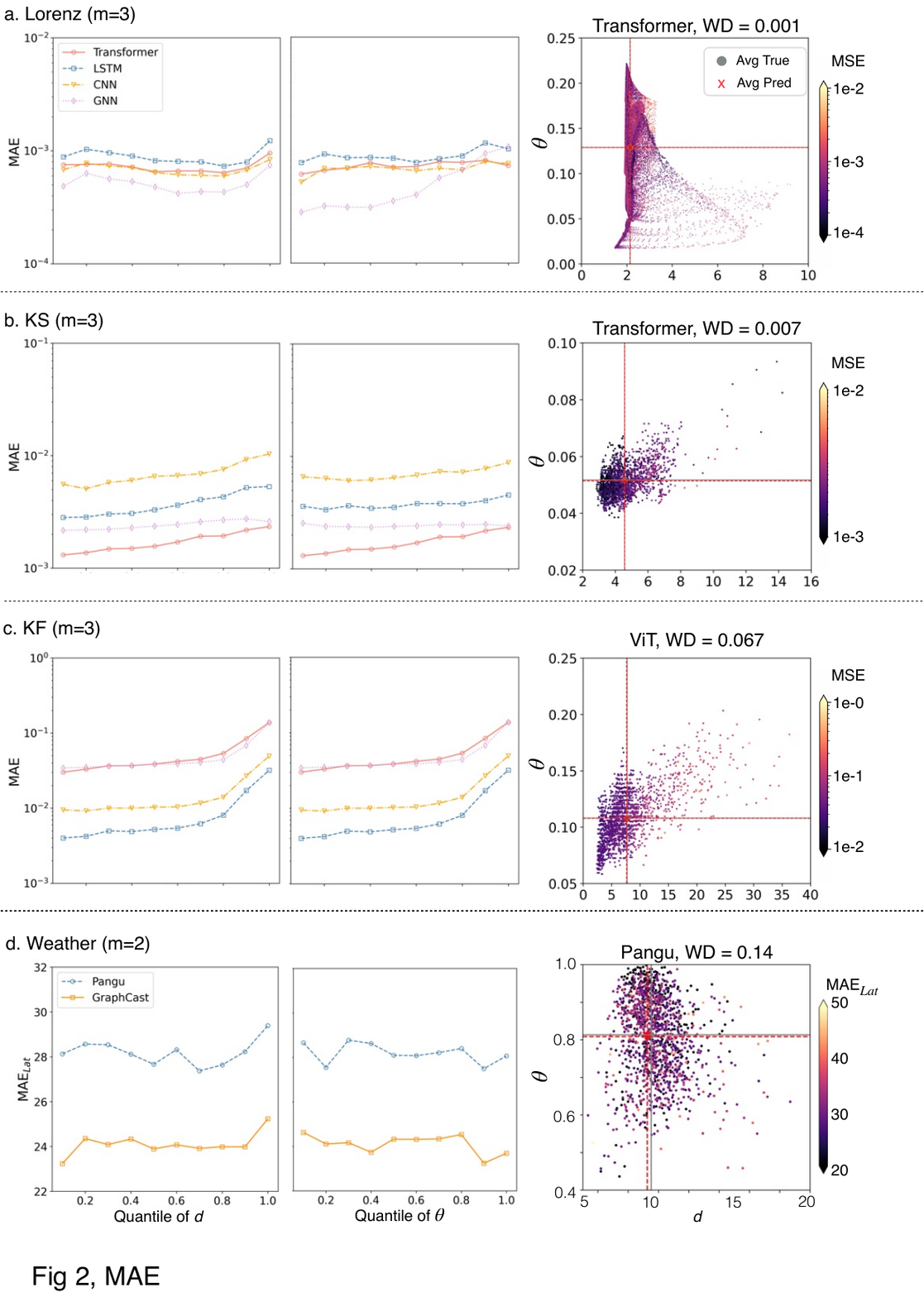}
    \caption{\textbf{Relationship between MAE and dynamical indices for one step lead time.} Left and middle columns: the averaged forecast error for direct single-step prediction with a lead time of one step, over the quantile of $d$ (left) and $\theta$ (middle). The average dynamical indices of both true and predicted data are marked in the plots, along with the WD.}
    \label{si-fig:Fig2_MAE}
\end{figure}

\begin{figure}[H]
    \centering
    \includegraphics[width=0.95\linewidth]{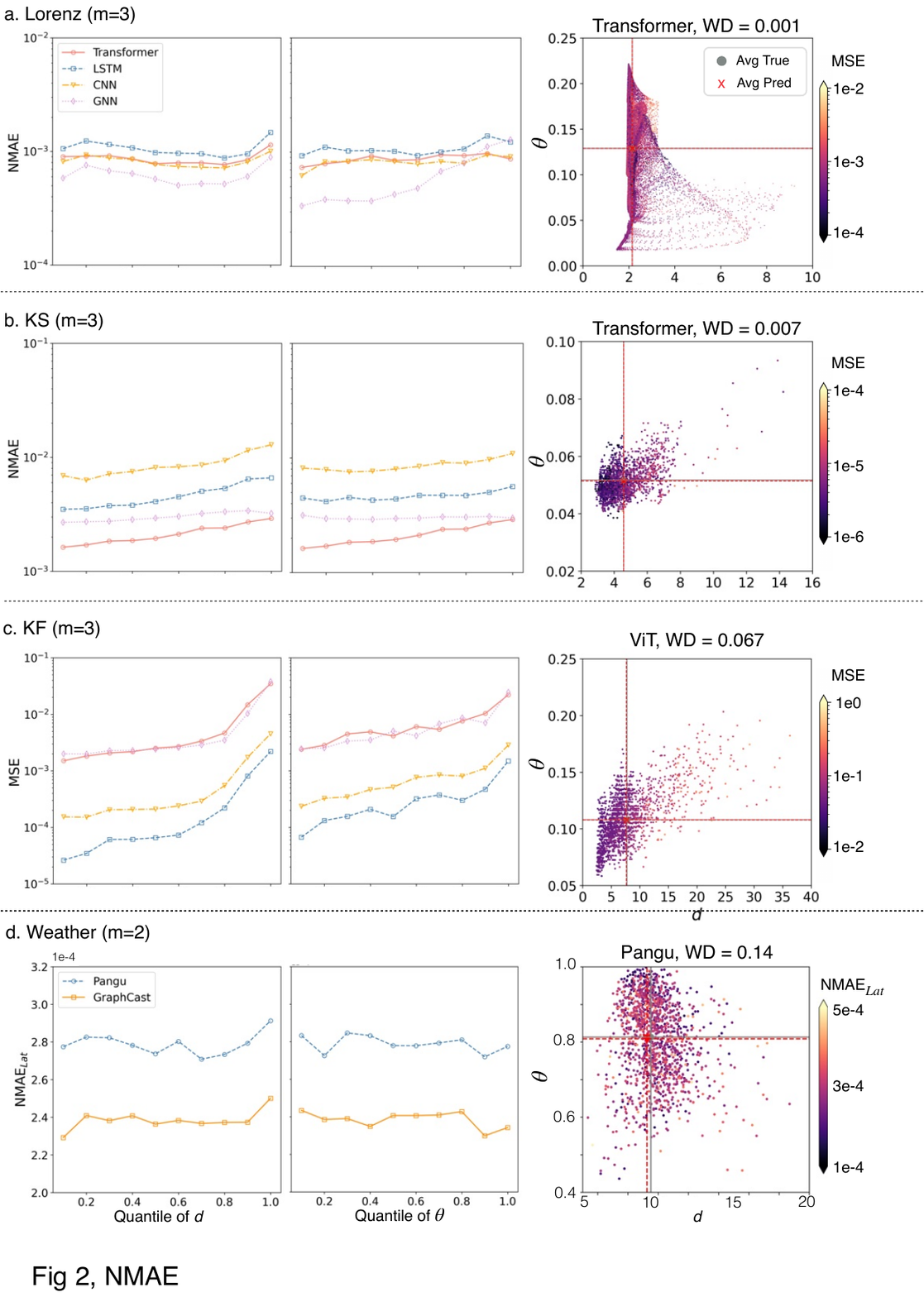}
    \caption{\textbf{Relationship between NMAE and dynamical indices for one step lead time.} Left and middle columns: the averaged forecast error for direct single-step prediction with a lead time of one step, over the quantile of $d$ (left) and $\theta$ (middle). The average dynamical indices of both true and predicted data are marked in the plots, along with the WD.}
    \label{si-fig:Fig2_NMAE}
\end{figure}

Fig.~\ref{si-fig:Fig2_NMSE}, \ref{si-fig:Fig2_MAE} and \ref{si-fig:Fig2_NMAE} show the same information as Fig.~\ref{fig:Fig2}, which is the forecast error as a function of DI quantile and $d-\theta$ space, but for NMSE, MAE, and NMAE, respectively.

\clearpage
\subsection{Direct forecast errors for longer lead times}\label{si:lead_forecast}
\begin{figure}[H]
    \centering
    \includegraphics[width=0.9\linewidth]{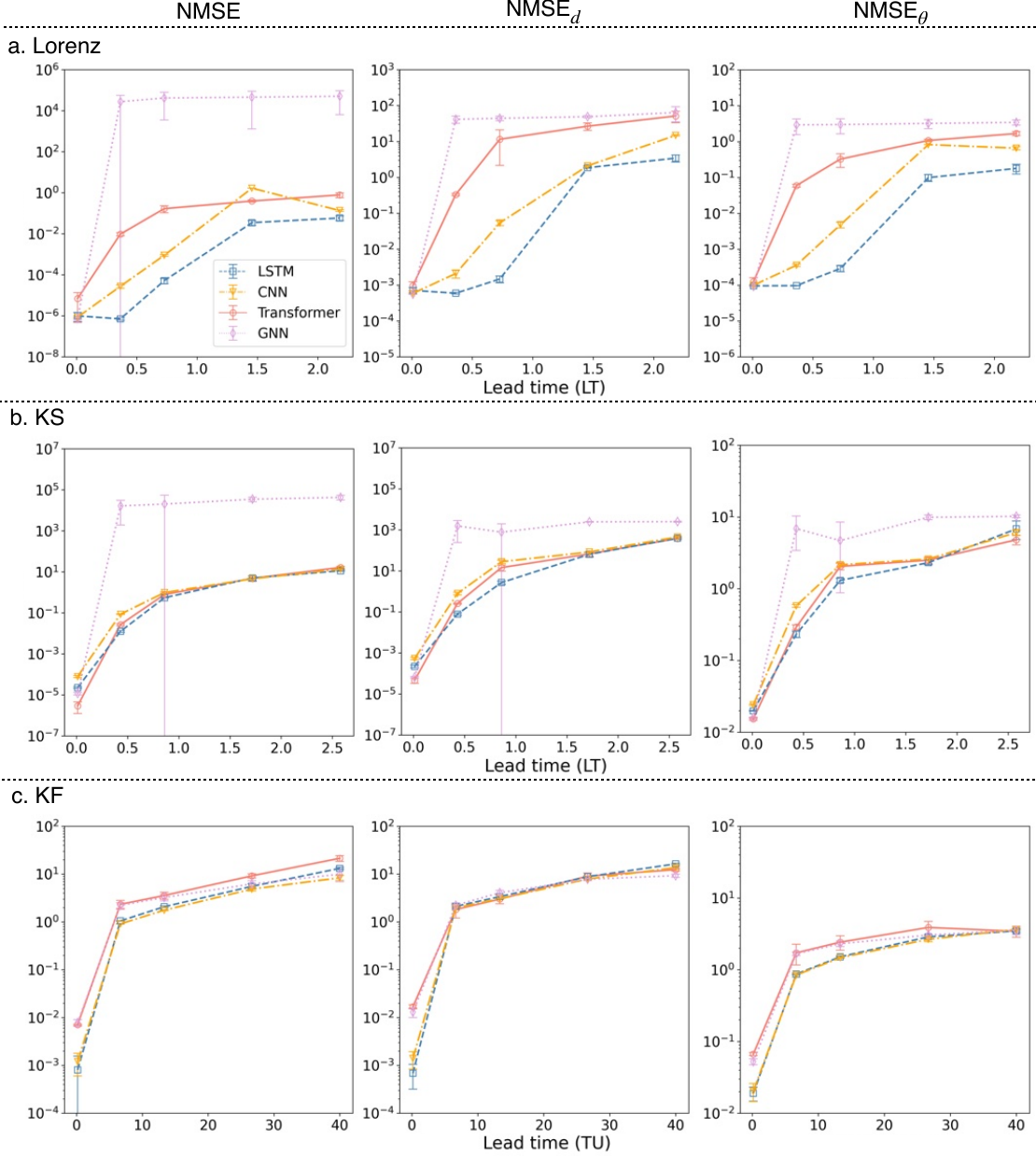}
    \caption{\textbf{NMSE vs forecast lead time} This figure shows the forecast error for lead prediction. The error bar is created using the standard deviation of 3 runs with different random initialization of the model. }
    \label{si-fig:FigS2_NMSE}
\end{figure}

\begin{figure}[H]
    \centering
    \includegraphics[width=0.9\linewidth]{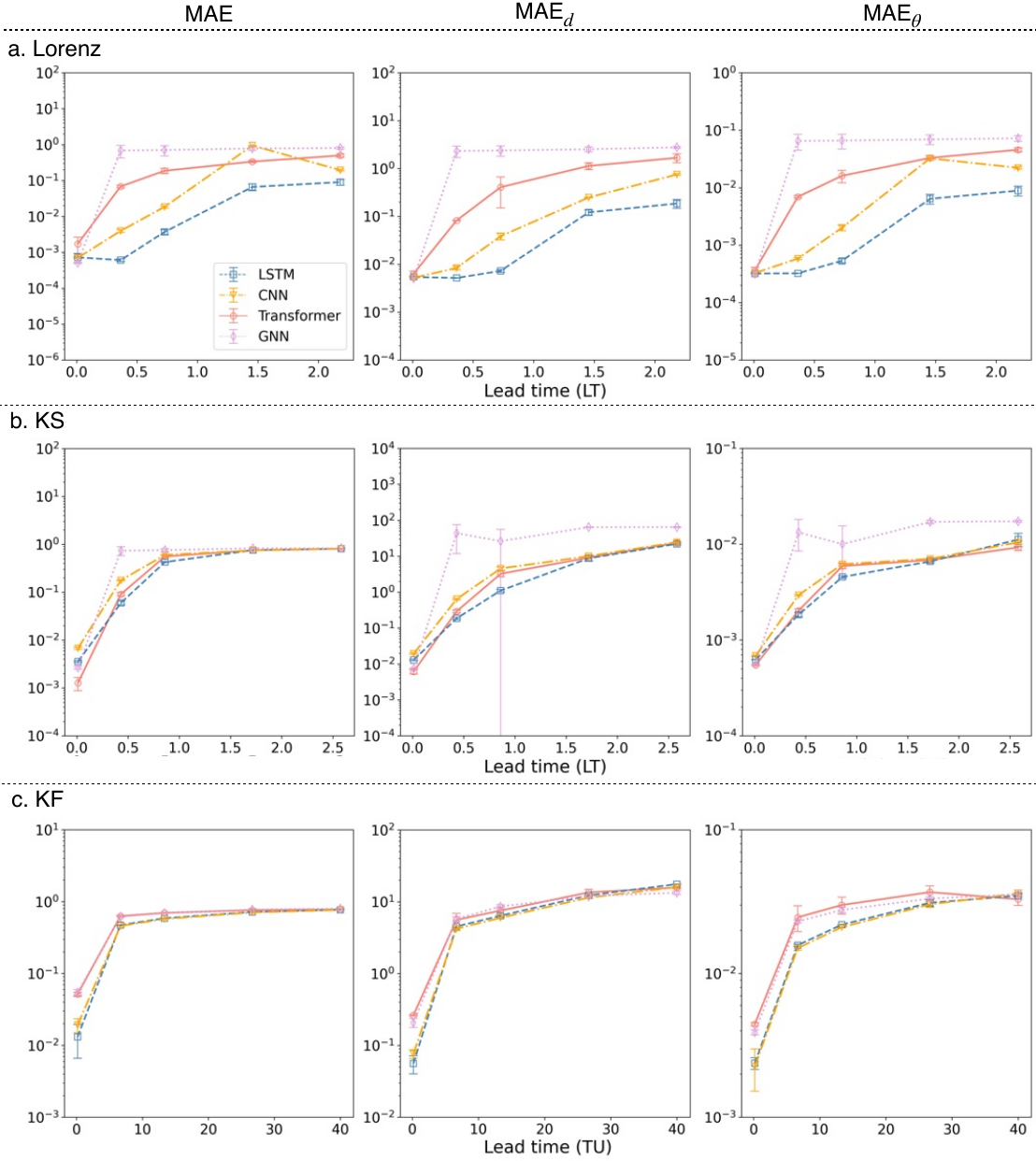}
    \caption{\textbf{MAE vs forecast lead time} This figure shows the forecast error for lead prediction. The error bar is created using the standard deviation of 3 runs with different random initialization of the model. }
    \label{si-fig:FigS2_MAE}
\end{figure}

\begin{figure}[H]
    \centering
    \includegraphics[width=0.9\linewidth]{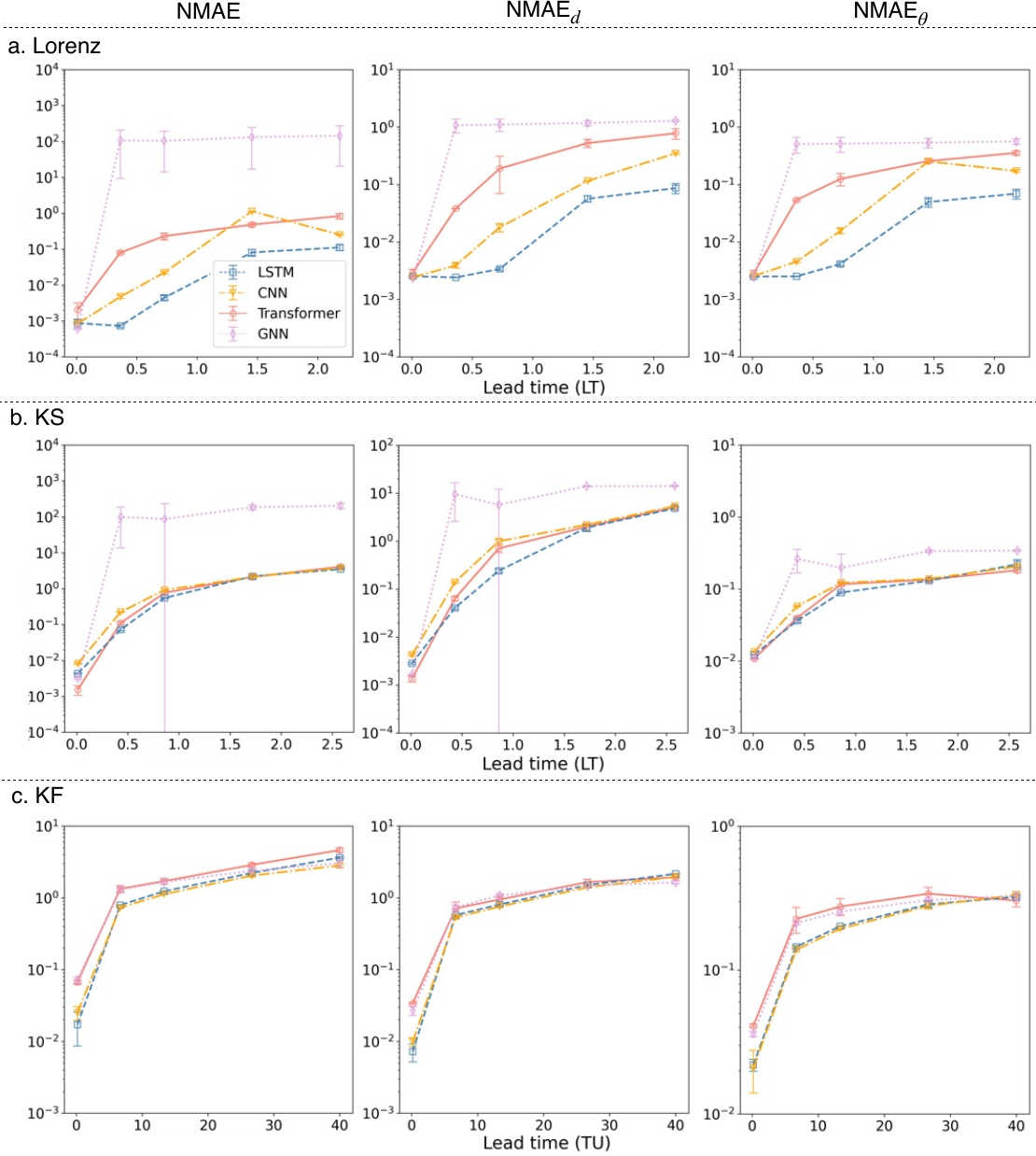}
    \caption{\textbf{NMAE vs forecast lead time} This figure shows the forecast error for lead prediction The error bar is created using the standard deviation of 3 runs with different random initialization of the model. }
    \label{si-fig:FigS2_NMAE}
\end{figure}

Fig.~\ref{si-fig:FigS2_NMSE}, \ref{si-fig:FigS2_MAE}, and ~\ref{si-fig:FigS2_NMAE} present the NMSE, MAE, and NMAE values across different forecast lead times for three canonical datasets. The lead times considered are 1, 40, 80, 160, and 240 time steps, and are expressed in their corresponding time units in the plots.

\begin{figure}[H]
    \centering
    \includegraphics[width=0.9\linewidth]{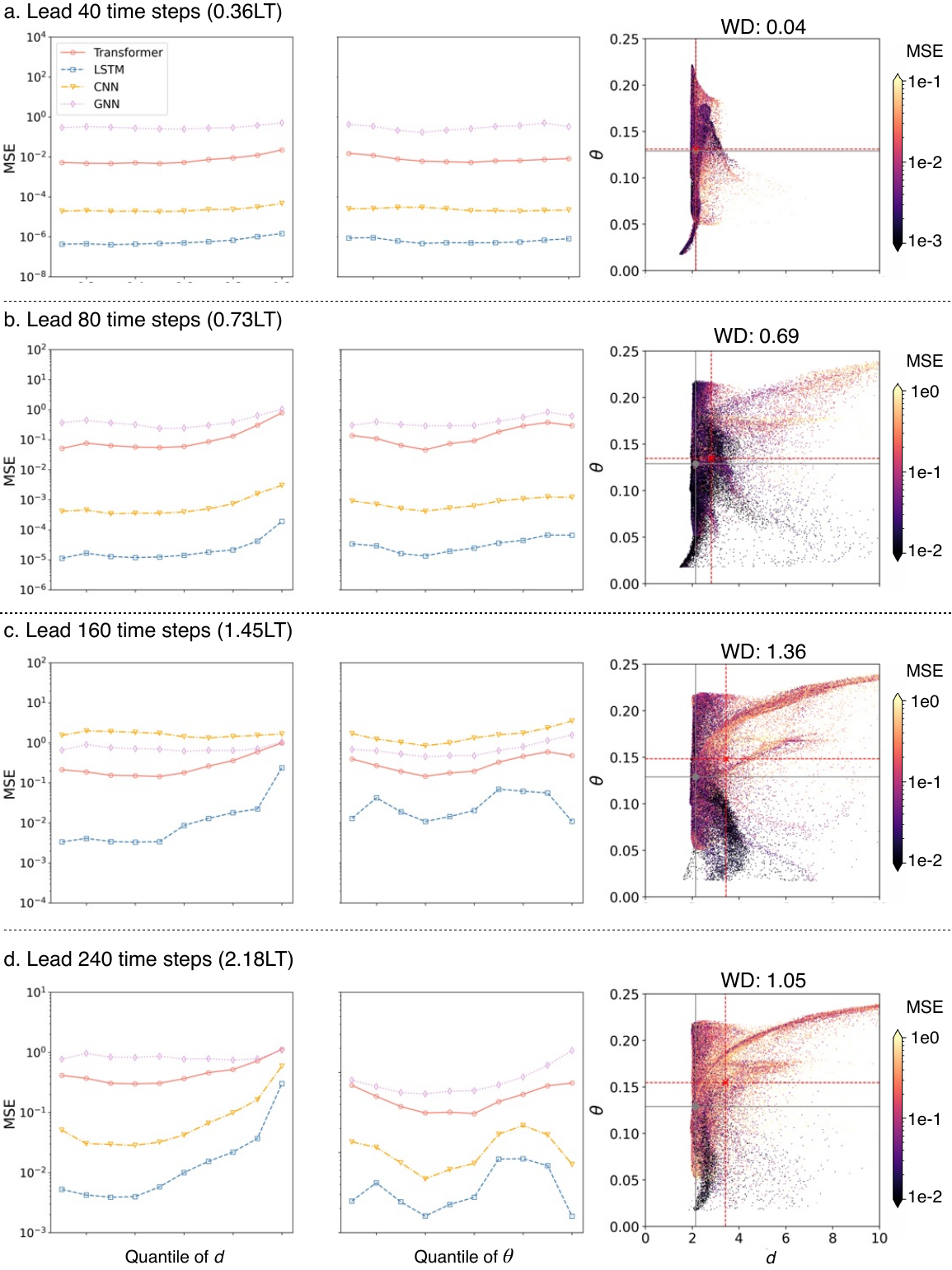}
    \caption{\textbf{Relationship between MSE and dynamical indices of Lorenz for longer lead time.} Left and middle columns: the averaged forecast error for direct single-step prediction with a lead time of one step, measured by MSE, over the quantile of $d$ (left) and $\theta$ (middle). Right column: The dynamical space of forecasts}
    \label{si-fig:Fig2_lead_lorenz}
\end{figure}

\begin{figure}[H]
    \centering
    \includegraphics[width=0.9\linewidth]{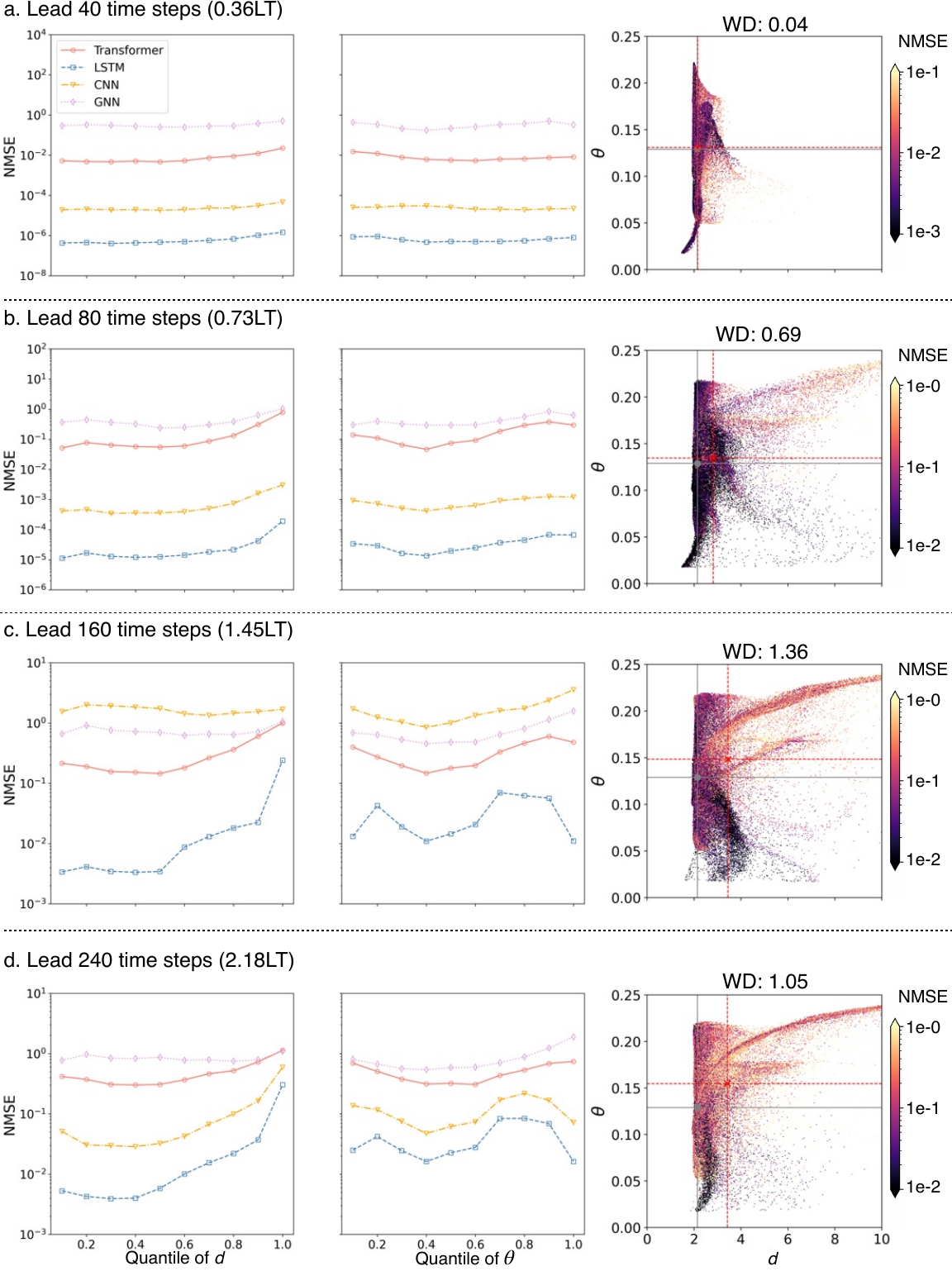}
    \caption{\textbf{Relationship between NMSE and dynamical indices of Lorenz for longer lead time.} Left and middle columns: the averaged forecast error for direct single-step prediction with a lead time of one step, measured by NMSE, over the quantile of $d$ (left) and $\theta$ (middle). Right column: The dynamical space of forecasts}
    \label{si-fig:Fig2_lead_lorenz_nmse}
\end{figure}

\begin{figure}[H]
    \centering
    \includegraphics[width=0.9\linewidth]{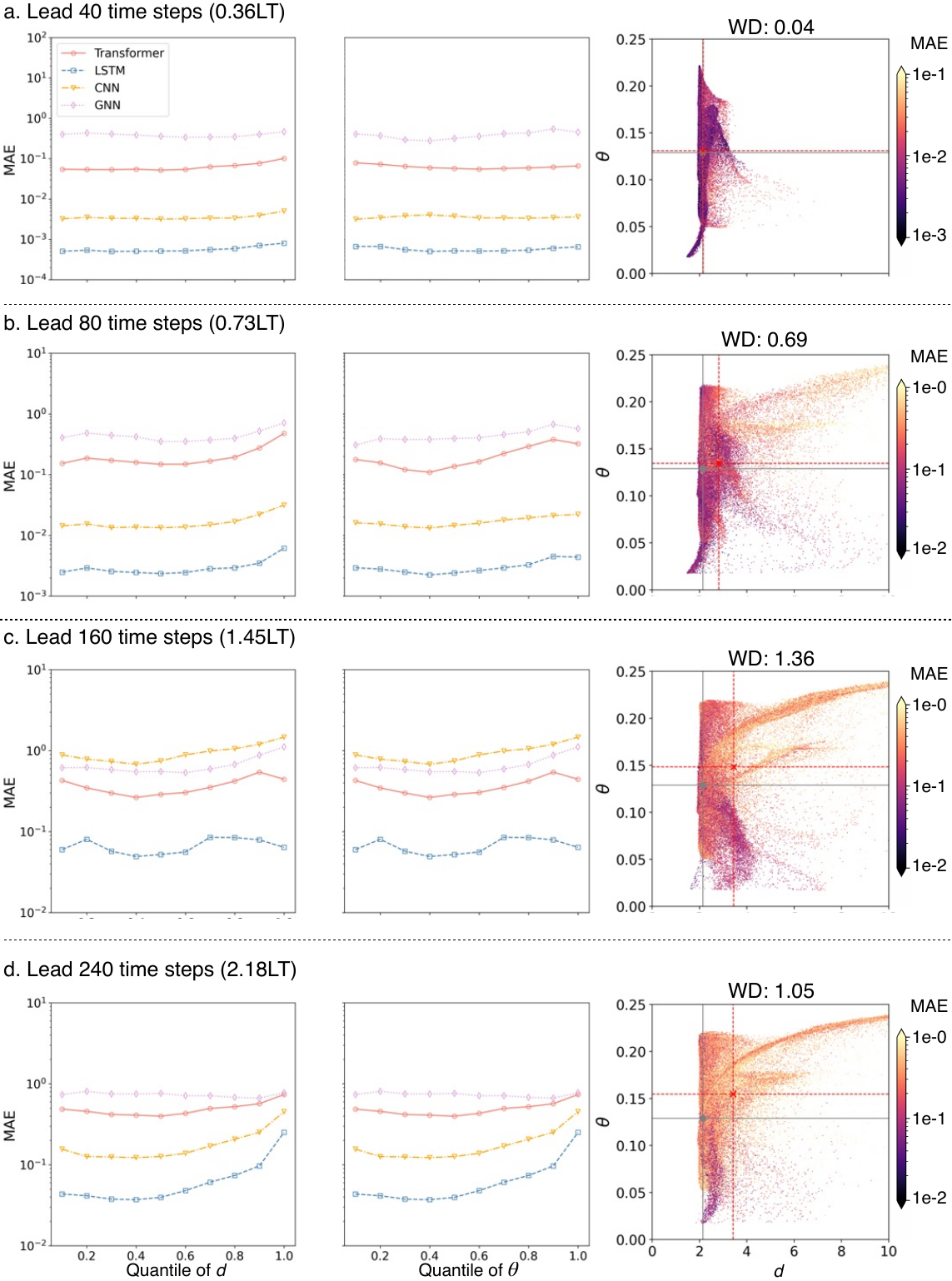}
    \caption{\textbf{Relationship between MAE and dynamical indices of Lorenz for longer lead time.} Left and middle columns: the averaged forecast error for direct single-step prediction with a lead time of one step, measured by MAE, over the quantile of $d$ (left) and $\theta$ (middle). Right column: The dynamical space of forecasts}
    \label{si-fig:Fig2_lead_lorenz_mae}
\end{figure}

\begin{figure}[H]
    \centering
    \includegraphics[width=0.9\linewidth]{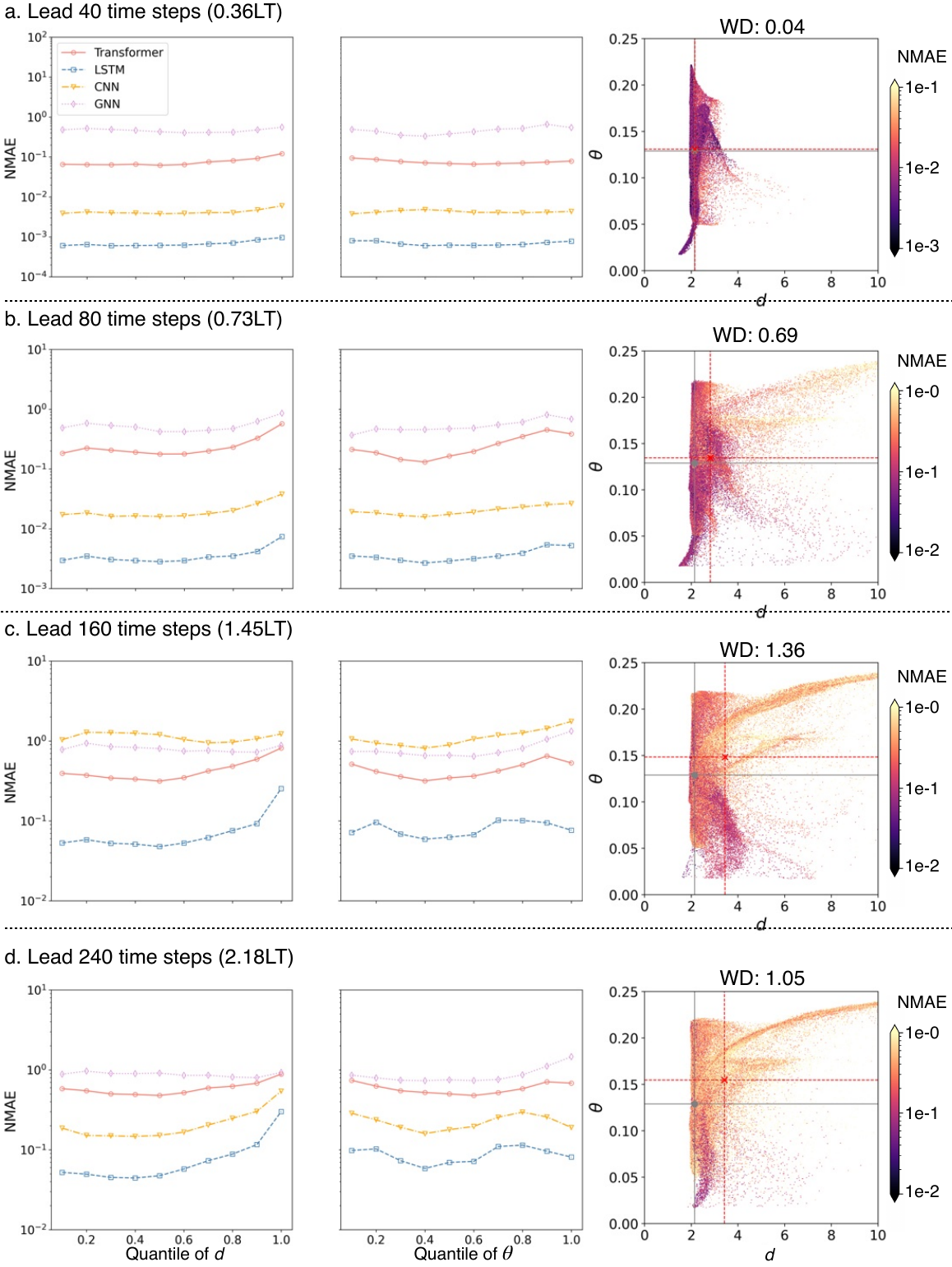}
    \caption{\textbf{Relationship between NMAE and dynamical indices of Lorenz for longer lead time.} Left and middle columns: the averaged forecast error for direct single-step prediction with a lead time of one step, measured by NMAE, over the quantile of $d$ (left) and $\theta$ (middle). Right column: The dynamical space of forecasts}
    \label{si-fig:Fig2_lead_lorenz_nmae}
\end{figure}

\begin{figure}[H]
    \centering
    \includegraphics[width=0.9\linewidth]{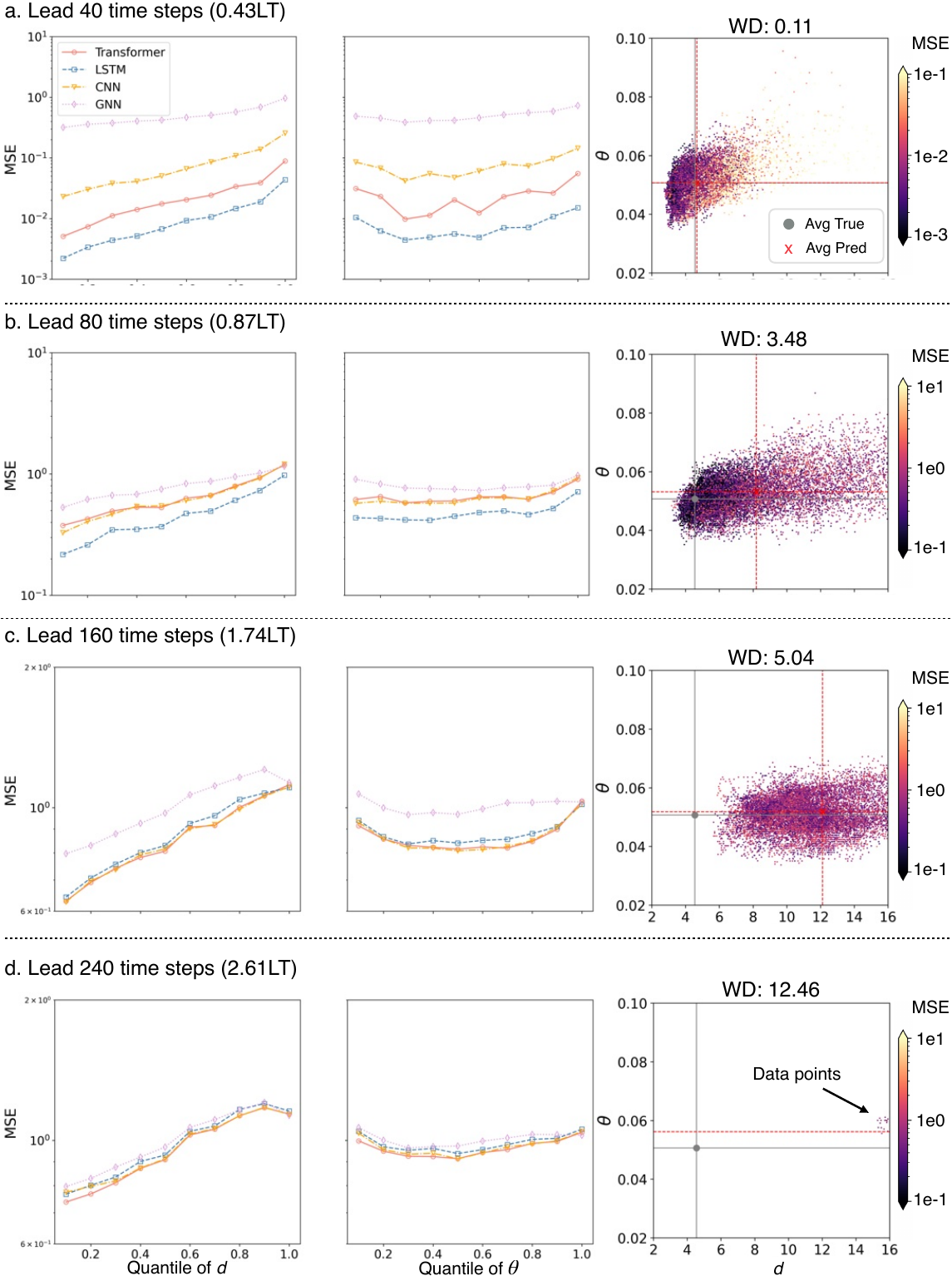}
    \caption{\textbf{Relationship between MSE and dynamical indices of KS for longer lead time.} Left and middle columns: the averaged forecast error for direct single-step prediction with a lead time of one step, measured by MSE, over the quantile of $d$ (left) and $\theta$ (middle). Right column: The dynamical space of forecasts}
    \label{si-fig:Fig2_lead_ks}
\end{figure}

\begin{figure}[H]
    \centering
    \includegraphics[width=0.9\linewidth]{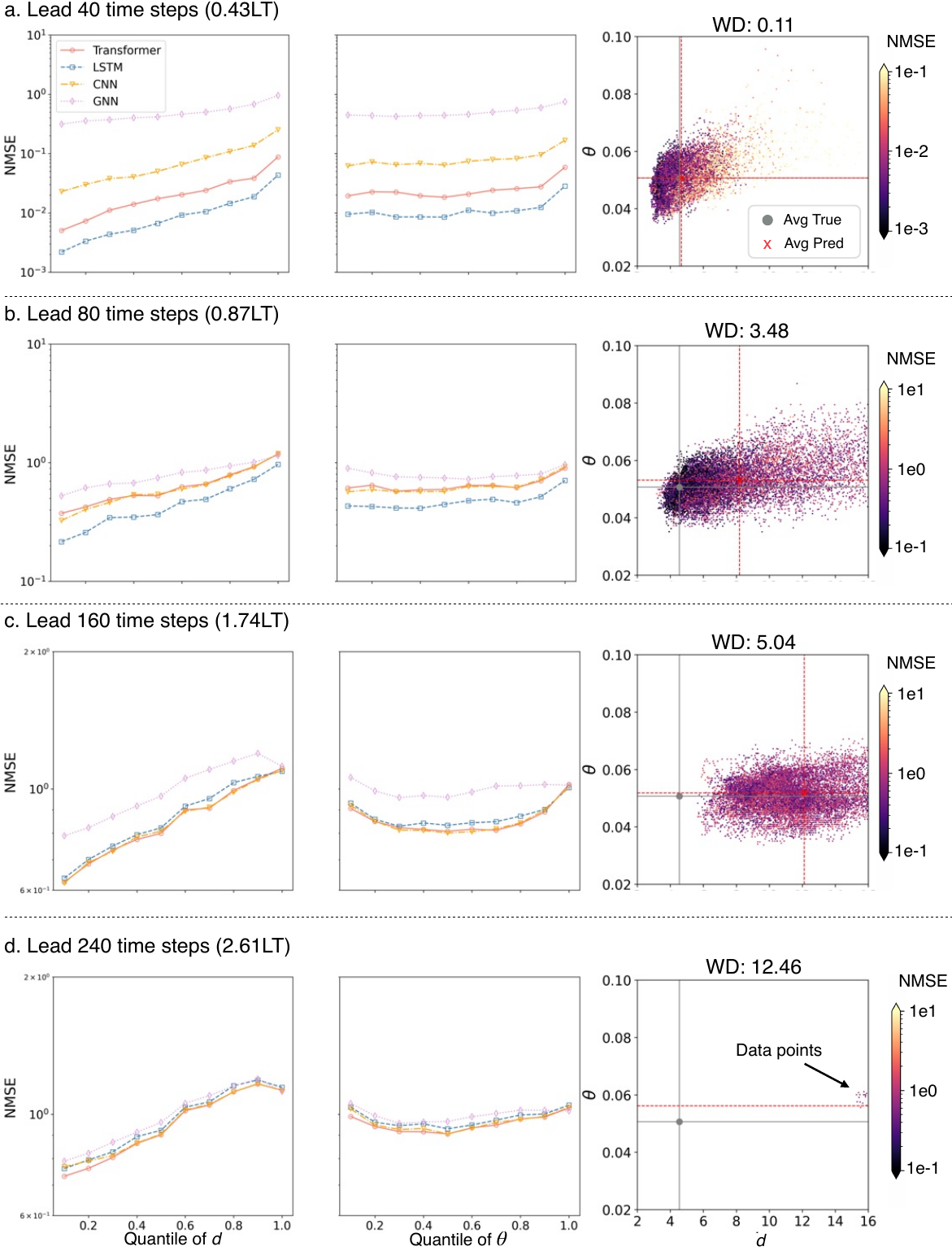}
    \caption{\textbf{Relationship between NMSE and dynamical indices of KS for longer lead time.} Left and middle columns: the averaged forecast error for direct single-step prediction with a lead time of one step, measured by NMSE, over the quantile of $d$ (left) and $\theta$ (middle). Right column: The dynamical space of forecasts}
    \label{si-fig:Fig2_lead_ks_NMSE}
\end{figure}

\begin{figure}[H]
    \centering
    \includegraphics[width=0.9\linewidth]{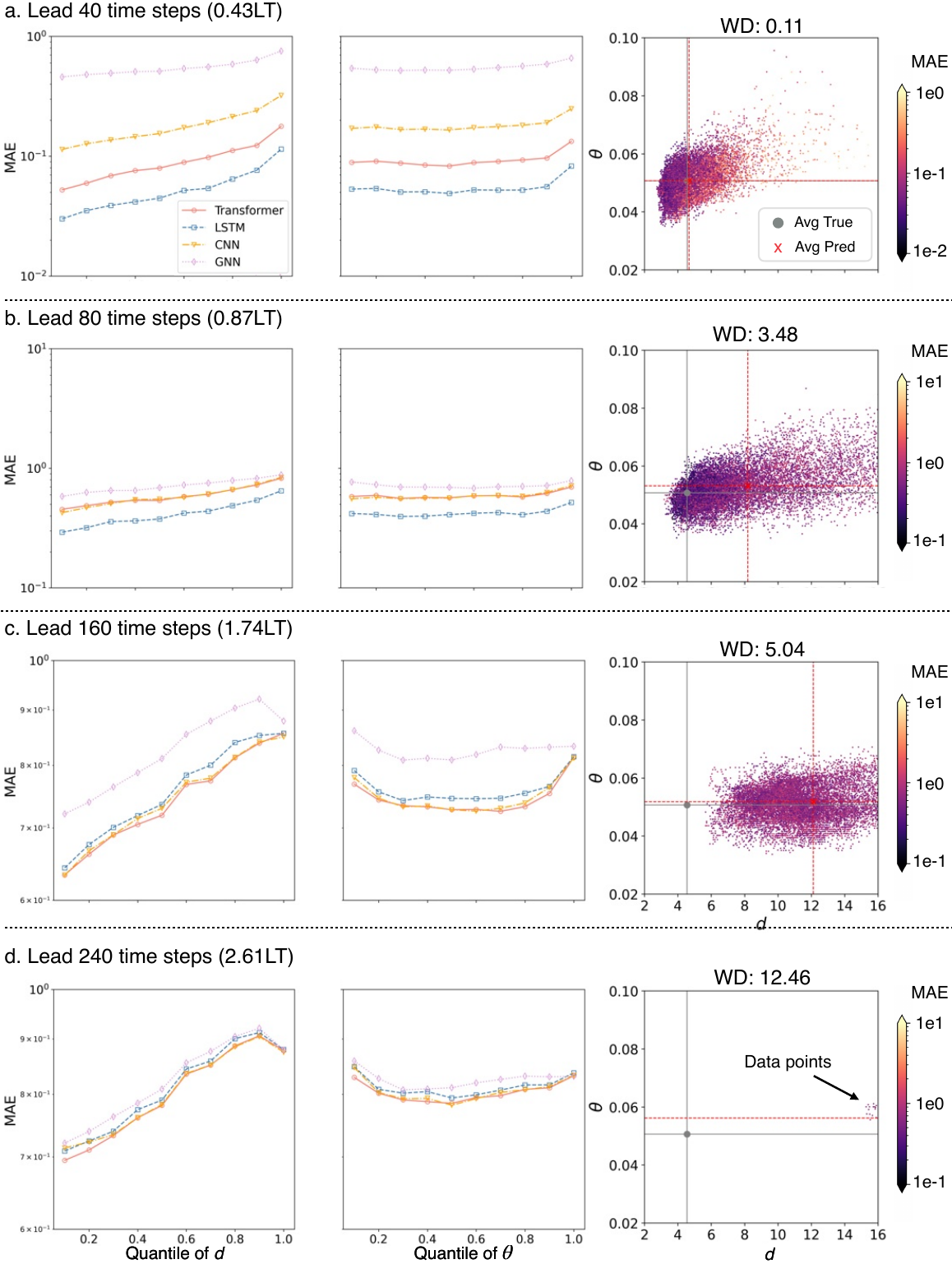}
    \caption{\textbf{Relationship between MAE and dynamical indices of KS for longer lead time.} Left and middle columns: the averaged forecast error for direct single-step prediction with a lead time of one step, measured by MSE, over the quantile of $d$ (left) and $\theta$ (middle). Right column: The dynamical space of forecasts}
    \label{si-fig:Fig2_lead_ks_MAE}
\end{figure}

\begin{figure}[H]
    \centering
    \includegraphics[width=0.9\linewidth]{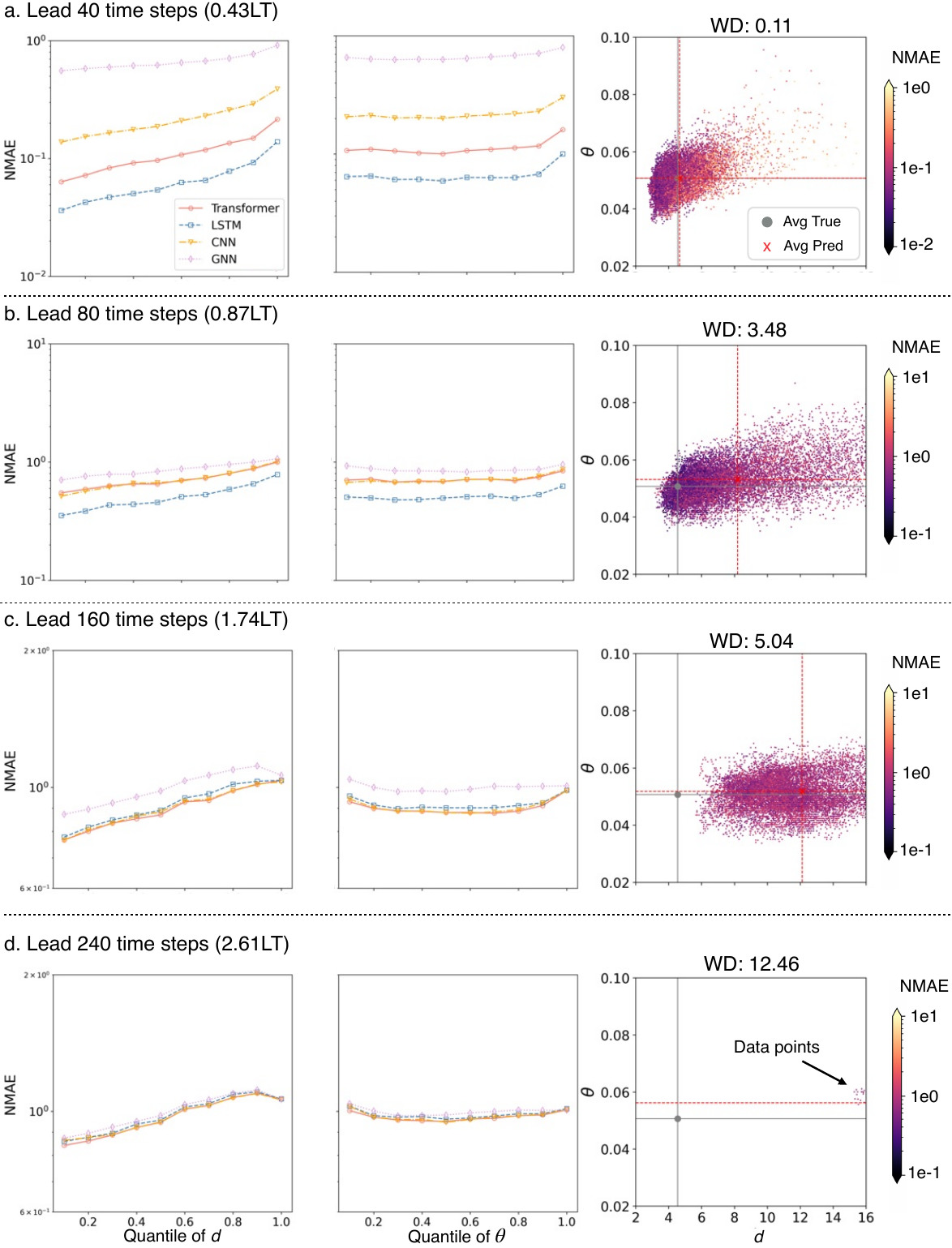}
    \caption{\textbf{Relationship between NMAE and dynamical indices of KS for longer lead time.} Left and middle columns: the averaged forecast error for direct single-step prediction with a lead time of one step, measured by NMAE, over the quantile of $d$ (left) and $\theta$ (middle). Right column: The dynamical space of forecasts}
    \label{si-fig:Fig2_lead_ks_nmae}
\end{figure}

\begin{figure}[H]
    \centering
    \includegraphics[width=0.9\linewidth]{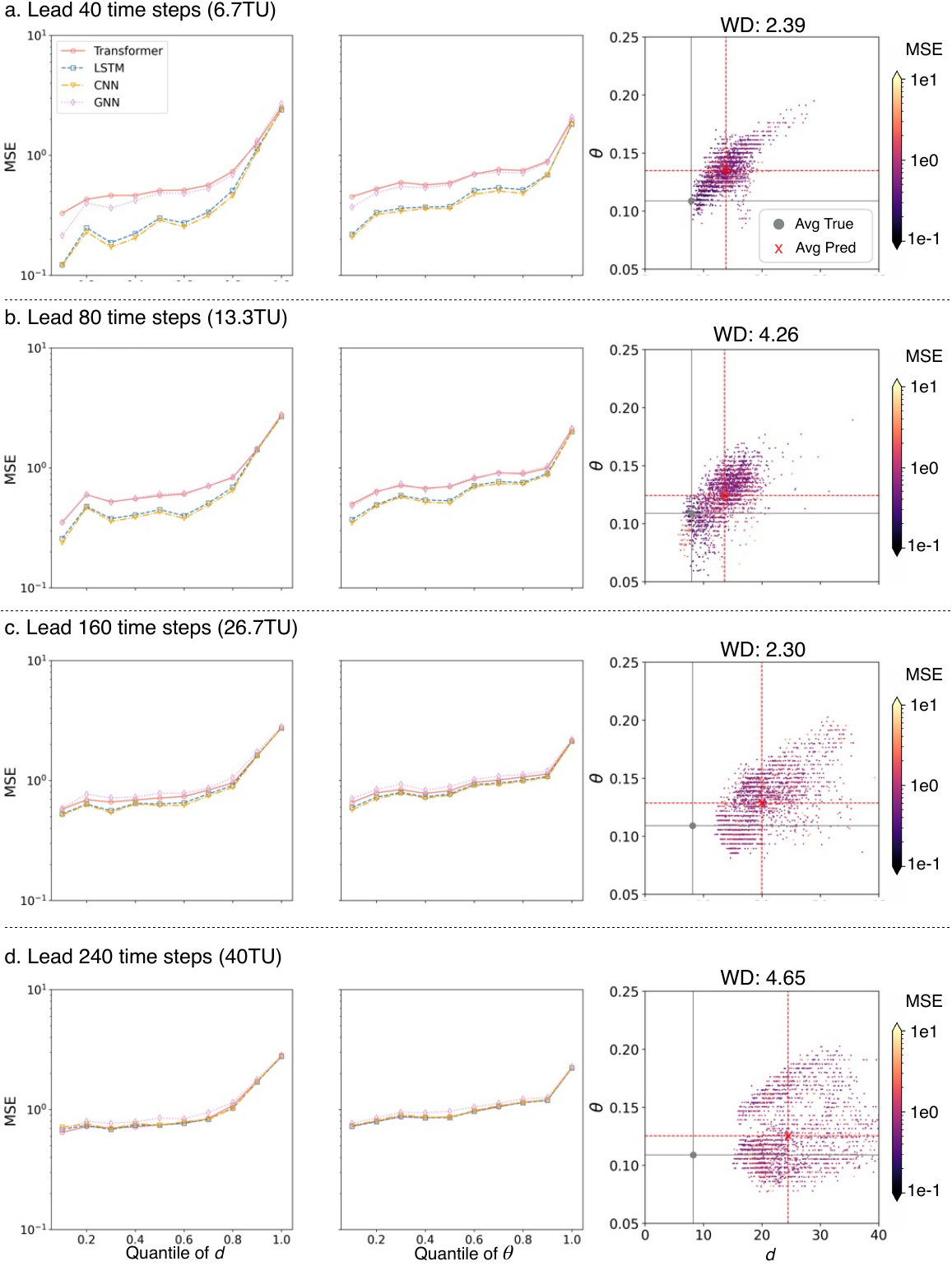}
    \caption{\textbf{Relationship between MSE and dynamical indices of KF for longer lead time.} Left and middle columns: the averaged forecast error for direct single-step prediction with a lead time of one step, measured by MSE, over the quantile of $d$ (left) and $\theta$ (middle). Right column: The dynamical space of forecasts}
    \label{si-fig:Fig2_lead_kf_MSE}
\end{figure}

\begin{figure}[H]
    \centering
    \includegraphics[width=0.9\linewidth]{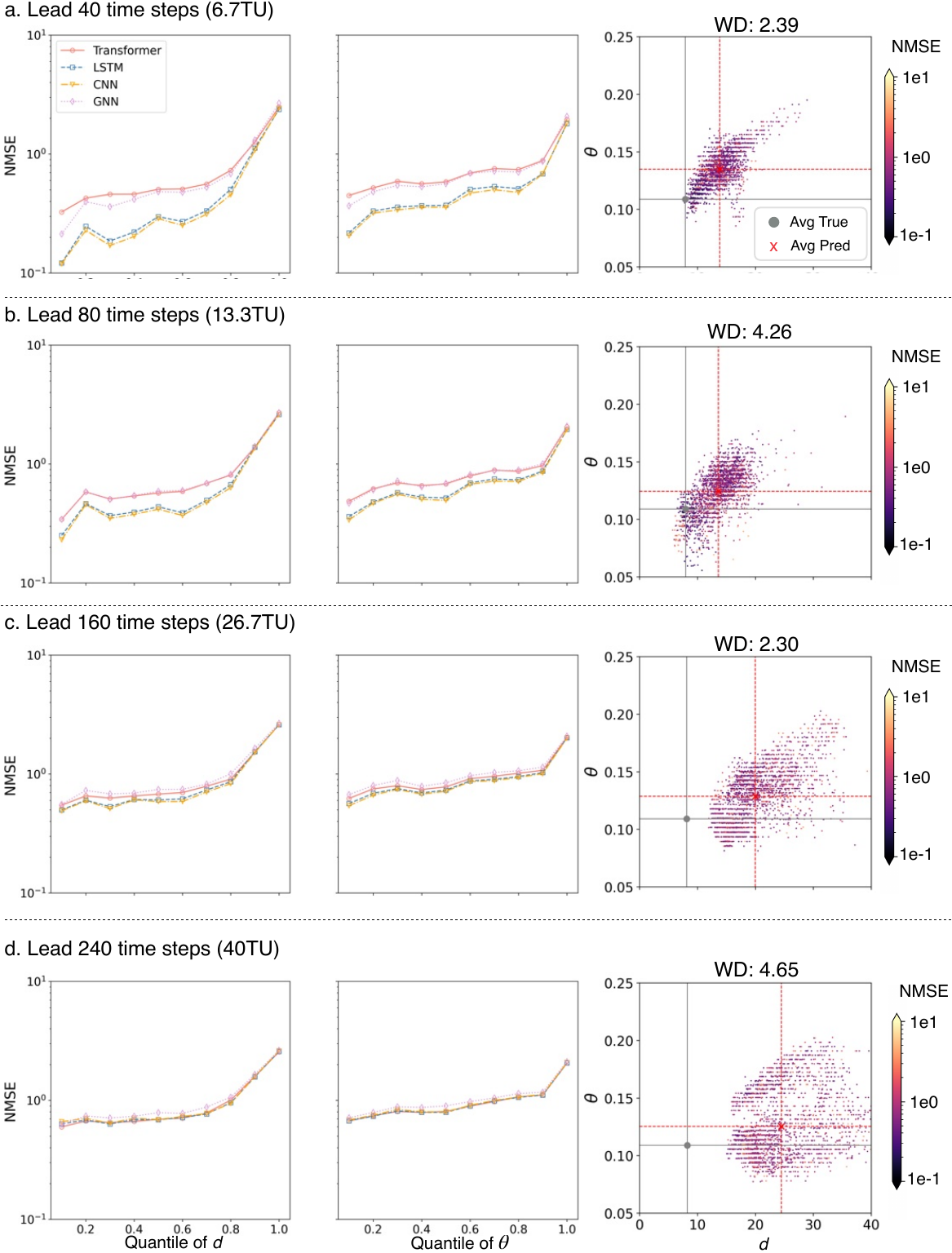}
    \caption{\textbf{Relationship between NMSE and dynamical indices of KF for longer lead time.} Left and middle columns: the averaged forecast error for direct single-step prediction with a lead time of one step, measured by NMSE, over the quantile of $d$ (left) and $\theta$ (middle). Right column: The dynamical space of forecasts}
    \label{si-fig:Fig2_lead_kf_NMSE}
\end{figure}

\begin{figure}[H]
    \centering
    \includegraphics[width=0.9\linewidth]{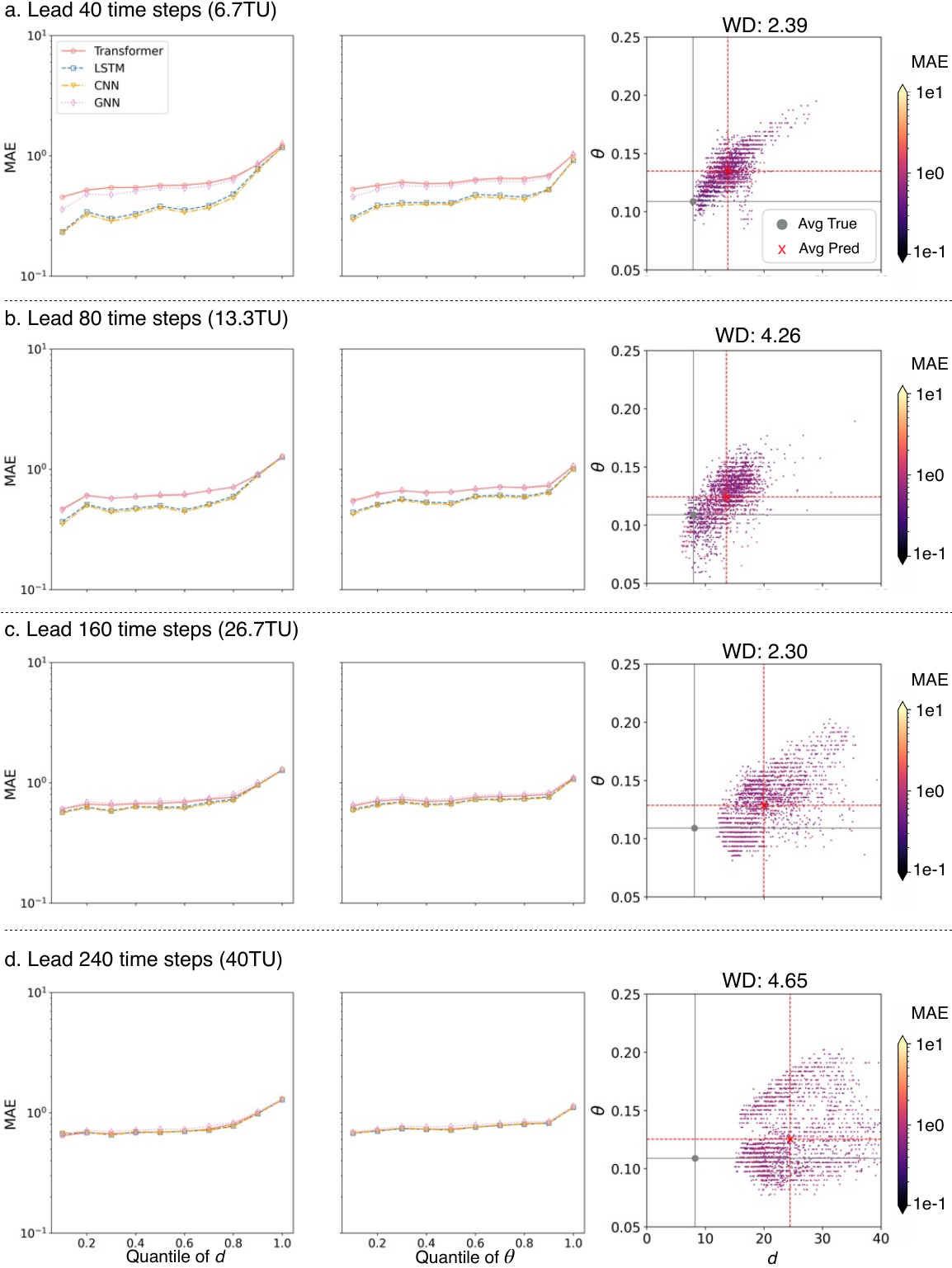}
    \caption{\textbf{Relationship between MAE and dynamical indices of KF for longer lead time.} Left and middle columns: the averaged forecast error for direct single-step prediction with a lead time of one step, measured by MAE, over the quantile of $d$ (left) and $\theta$ (middle). Right column: The dynamical space of forecasts}
    \label{si-fig:Fig2_lead_kf_MAE}
\end{figure}

\begin{figure}[H]
    \centering
    \includegraphics[width=0.9\linewidth]{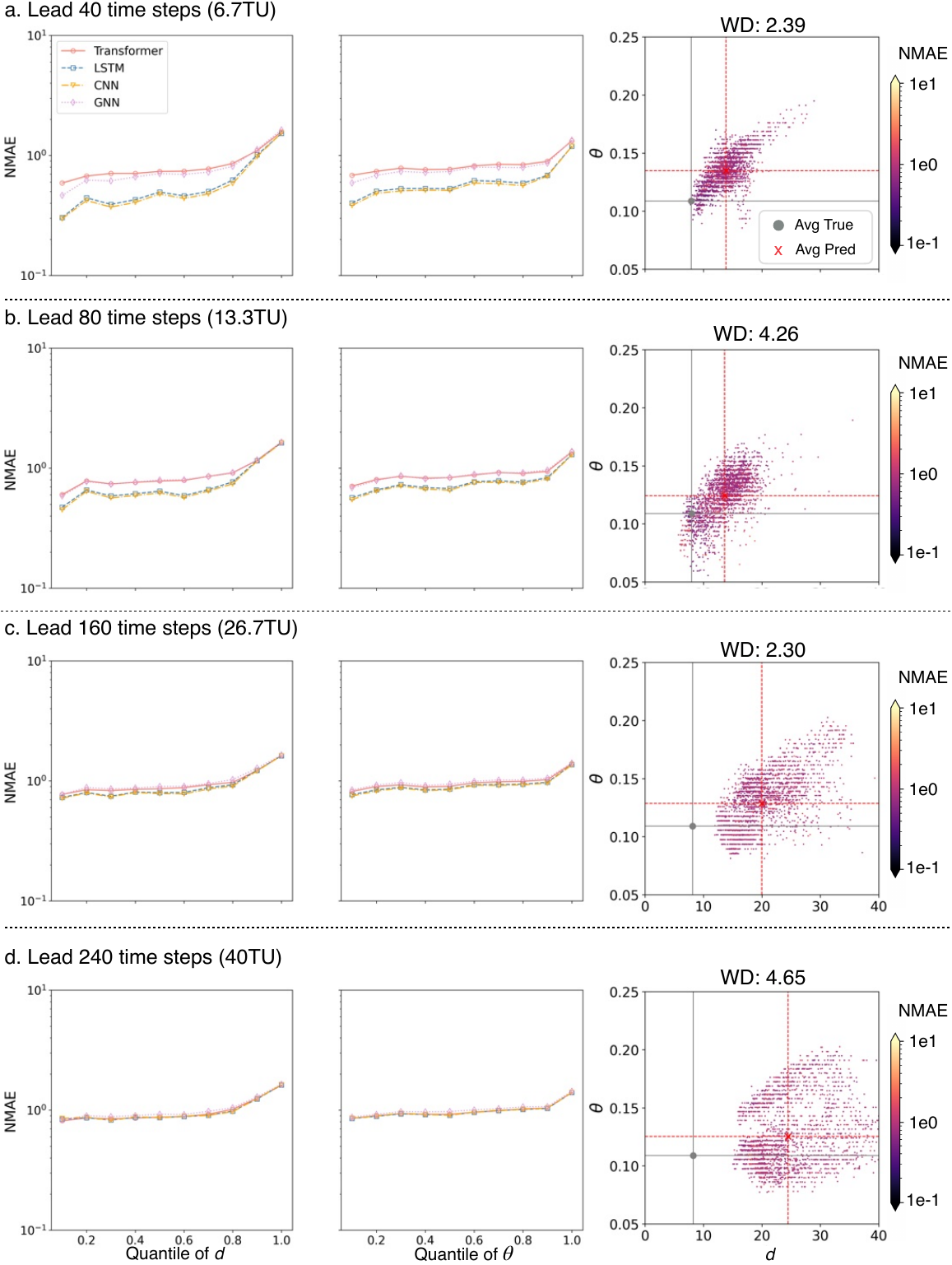}
    \caption{\textbf{Relationship between NMAE and dynamical indices of KF for longer lead time.} Left and middle columns: the averaged forecast error for direct single-step prediction with a lead time of one step, measured by NMAE, over the quantile of $d$ (left) and $\theta$ (middle). Right column: The dynamical space of forecasts}
    \label{si-fig:Fig2_lead_kf_NMAE}
\end{figure}

Fig.~\ref{si-fig:Fig2_lead_lorenz}--\ref{si-fig:Fig2_lead_kf_NMAE} illustrate the forecast errors as a function of DI quantile for 1, 40, 80, 160, and 240 lead time steps, along with the corresponding $d–\theta$ dynamical space of the forecasts.


\clearpage
\subsection{Direct forecast errors for different input lengths}
\label{si:inputs}
\begin{figure}[H]
    \centering
    \includegraphics[width=0.9\linewidth]{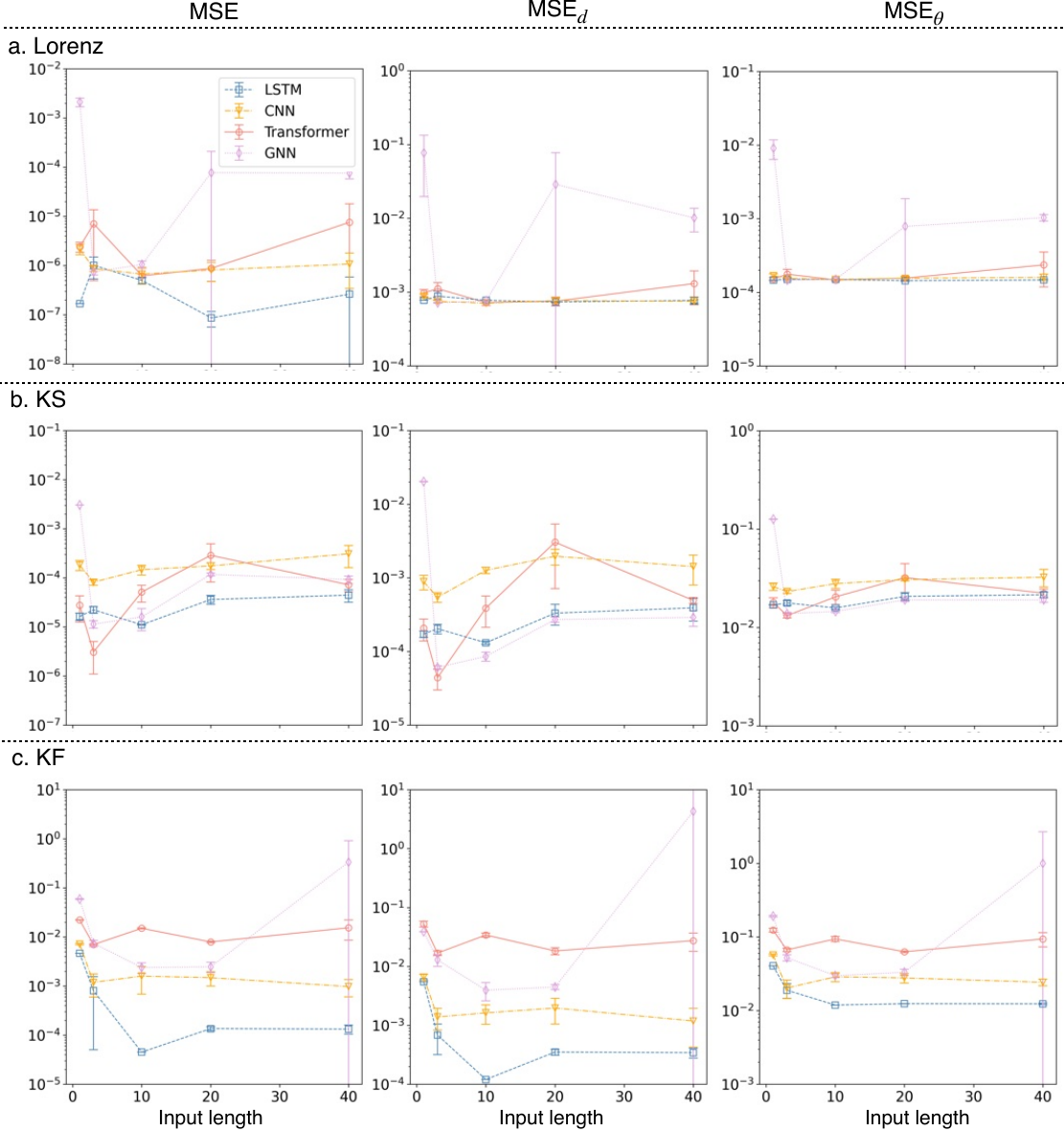}
    \caption{\textbf{MSE, MSE$_d$ and MSE$_\theta$ vs different input length.} The $x$-axis shows different initialization time steps, $y$-axis is the corresponding forecast error for the lead one step forecast: Left: MSE; Middle and right: MSE$_d$ and MSE$_\theta$}
    \label{si-fig:FigS3_MSE}
\end{figure}

\begin{figure}[H]
    \centering
    \includegraphics[width=0.9\linewidth]{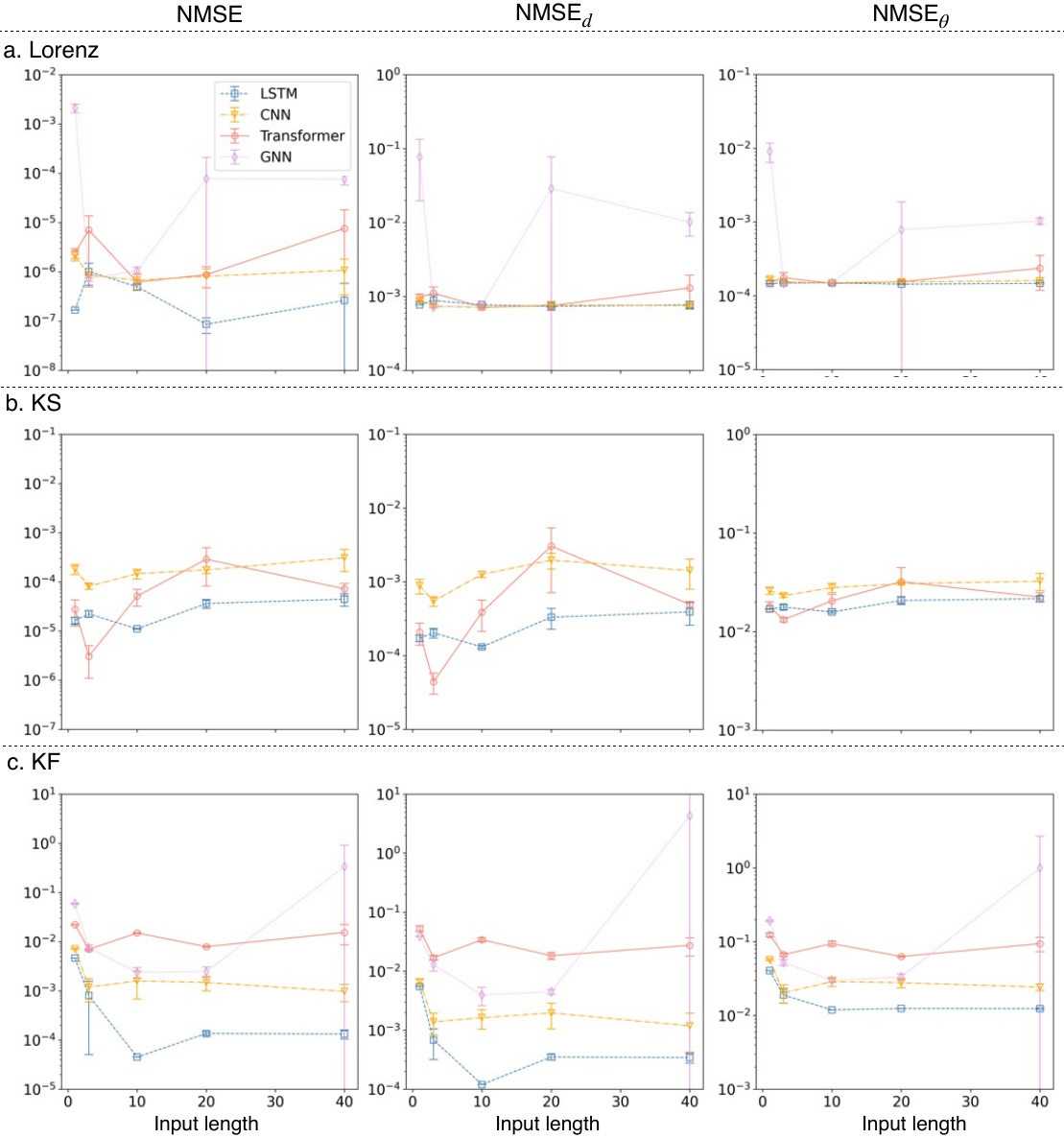}
    \caption{\textbf{NMSE, NMSE$_d$ and NMSE$_\theta$ vs different input length.} $x$-axis shows different initialization time steps, $y$-axis is the corresponding forecast error for the lead one step forecast: Left: NMSE; Middle and right: NMSE$_d$ and NMSE$_\theta$}
    \label{si-fig:FigS3_NMSE}
\end{figure}

\begin{figure}[H]
    \centering
    \includegraphics[width=0.9\linewidth]{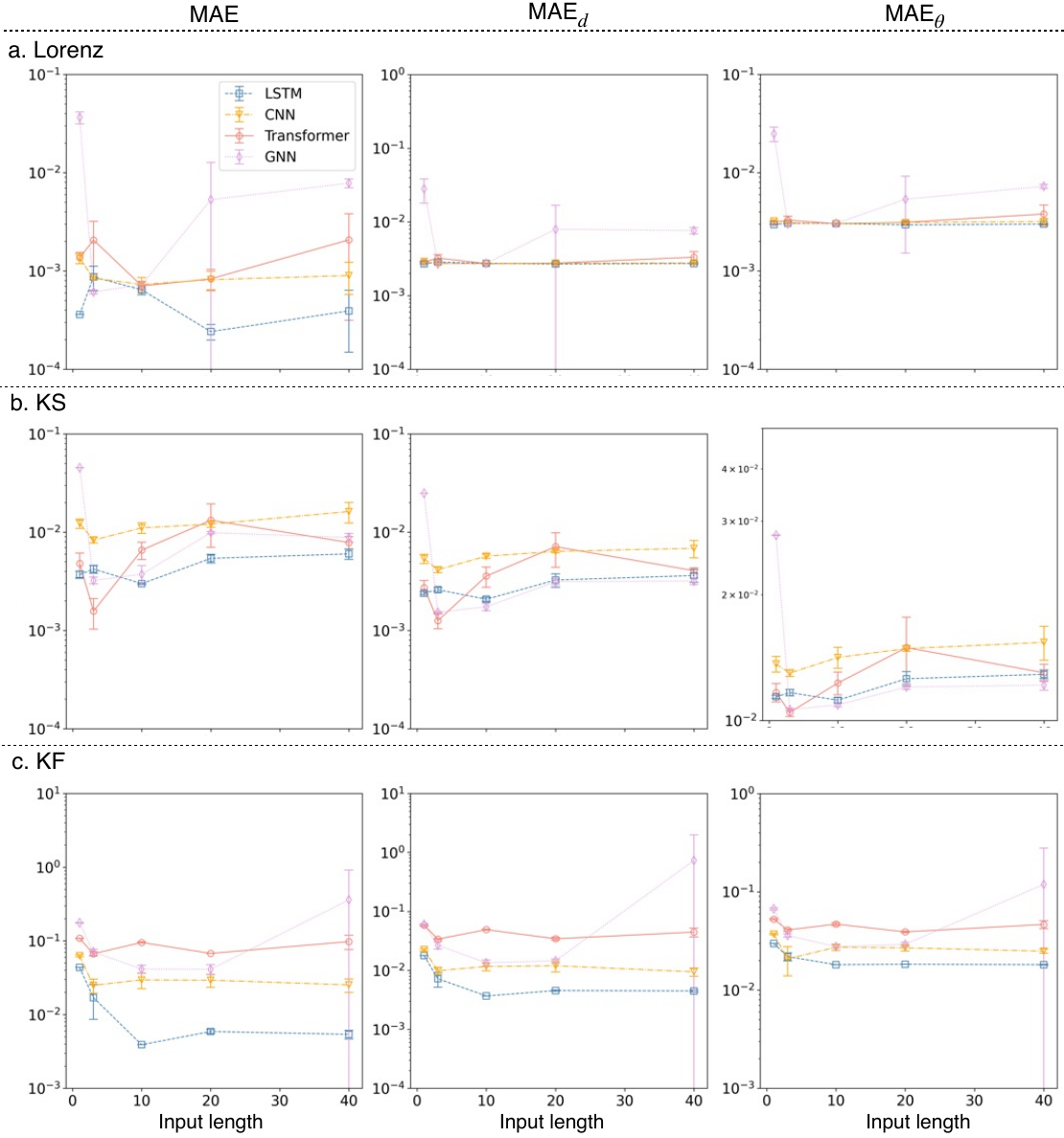}
    \caption{\textbf{MAE, MAE$_d$ and MAE$_\theta$ vs different input length.} $x$-axis shows different initialization time steps, $y$-axis is the corresponding forecast error for the lead one step forecast: Left: MAE; Middle and right: MAE$_d$ and MAE$_\theta$}
    \label{si-fig:FigS3_MAE}
\end{figure}

\begin{figure}[H]
    \centering
    \includegraphics[width=0.9\linewidth]{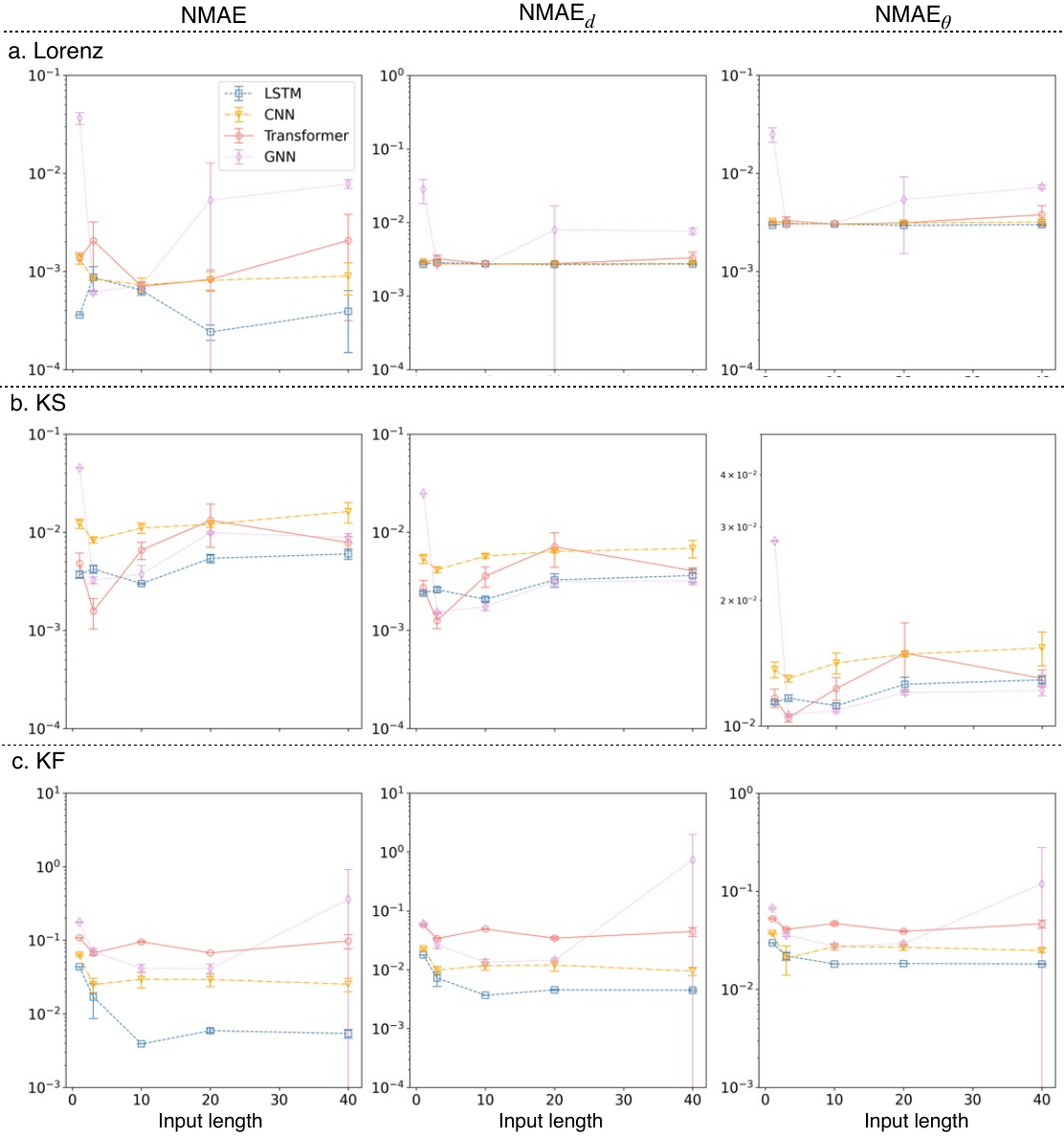}
    \caption{\textbf{NMAE, NMAE$_d$ and NMAE$_\theta$ vs different input length.} $x$-axis shows different initialization time steps, $y$-axis is the corresponding forecast error for the lead one step forecast: Left: NMAE; Middle and right: NMAE$_d$ and NMAE$_\theta$}
    \label{si-fig:FigS3_NMAE}
\end{figure}

Fig.~\ref{si-fig:FigS3_MSE}--~\ref{si-fig:FigS3_NMAE} show the forecast errors as a function of input length. Specifically, experiments are conducted using input sequences of 1, 20, 40, and 80 time steps.

\clearpage
\subsection{Other standard error metrics for recursive forecast}\label{si:recursive}
\begin{figure}[H]
    \centering
    \includegraphics[width=0.8\linewidth]{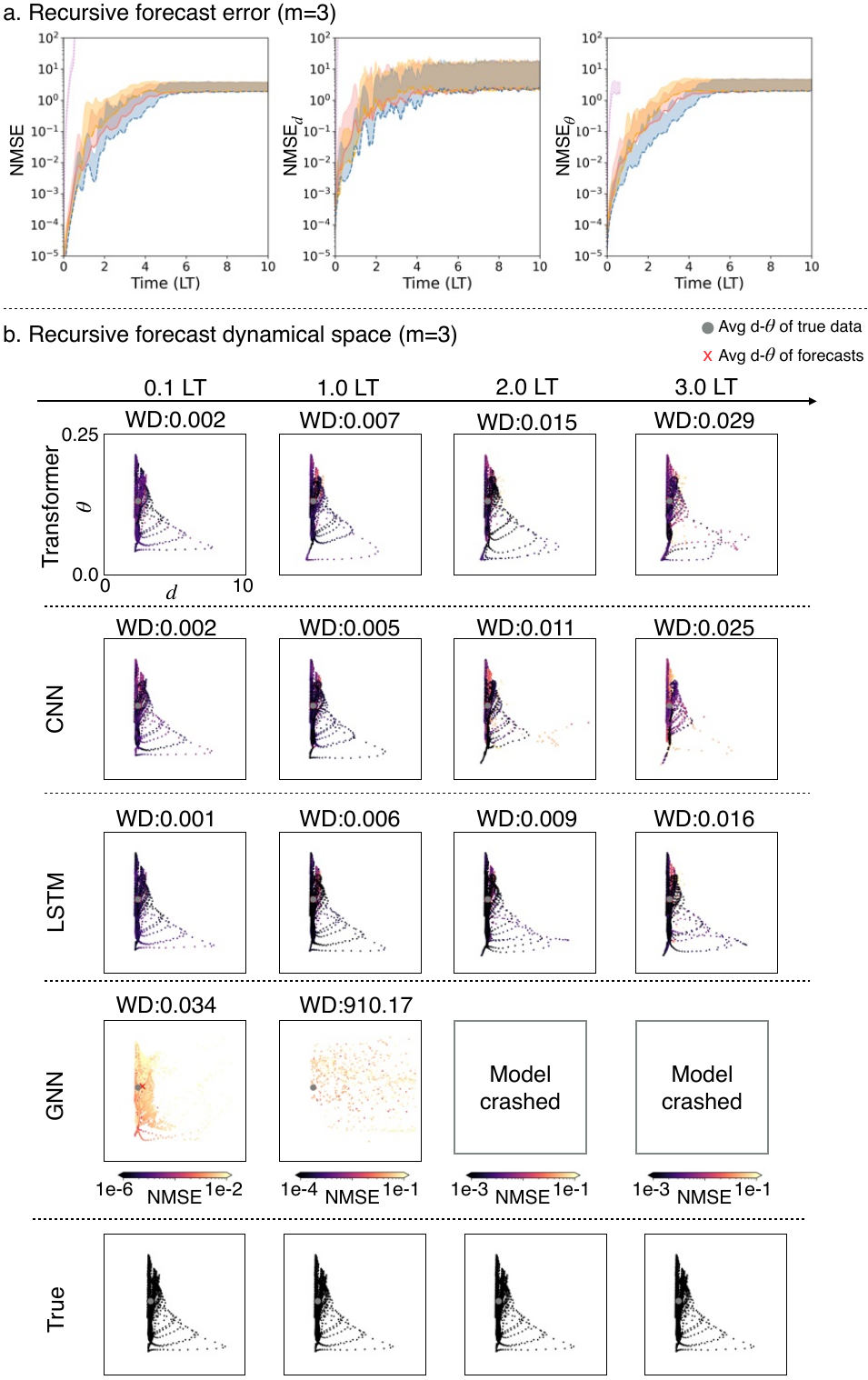}
    \caption{\textbf{NMSE and dynamical space of Lorenz recursive forecast.} Panel (a): Forecast error vs recursive forecast time in terms of Lyapunov time (LT). The shaded area represents the standard deviation of forecasts starting from 5000 initial states. Panel (b): $d-\theta$ space of the 5000 trajectories at forecast time 0.1 LT, 1.0 LT, 2.0 LT and 3.0 LT. The horizontal and vertical coordinates are $d$ and $\theta$, with the mean value of indices and WD annotated on the figure. The GNN $d-\theta$ spaces for 2.0LT and 3.0LT are not plotted since the model crashed and produced NAN results.}
    \label{si-fig:Fig3_NMSE}
\end{figure}

\begin{figure}[H]
    \centering
    \includegraphics[width=0.8\linewidth]{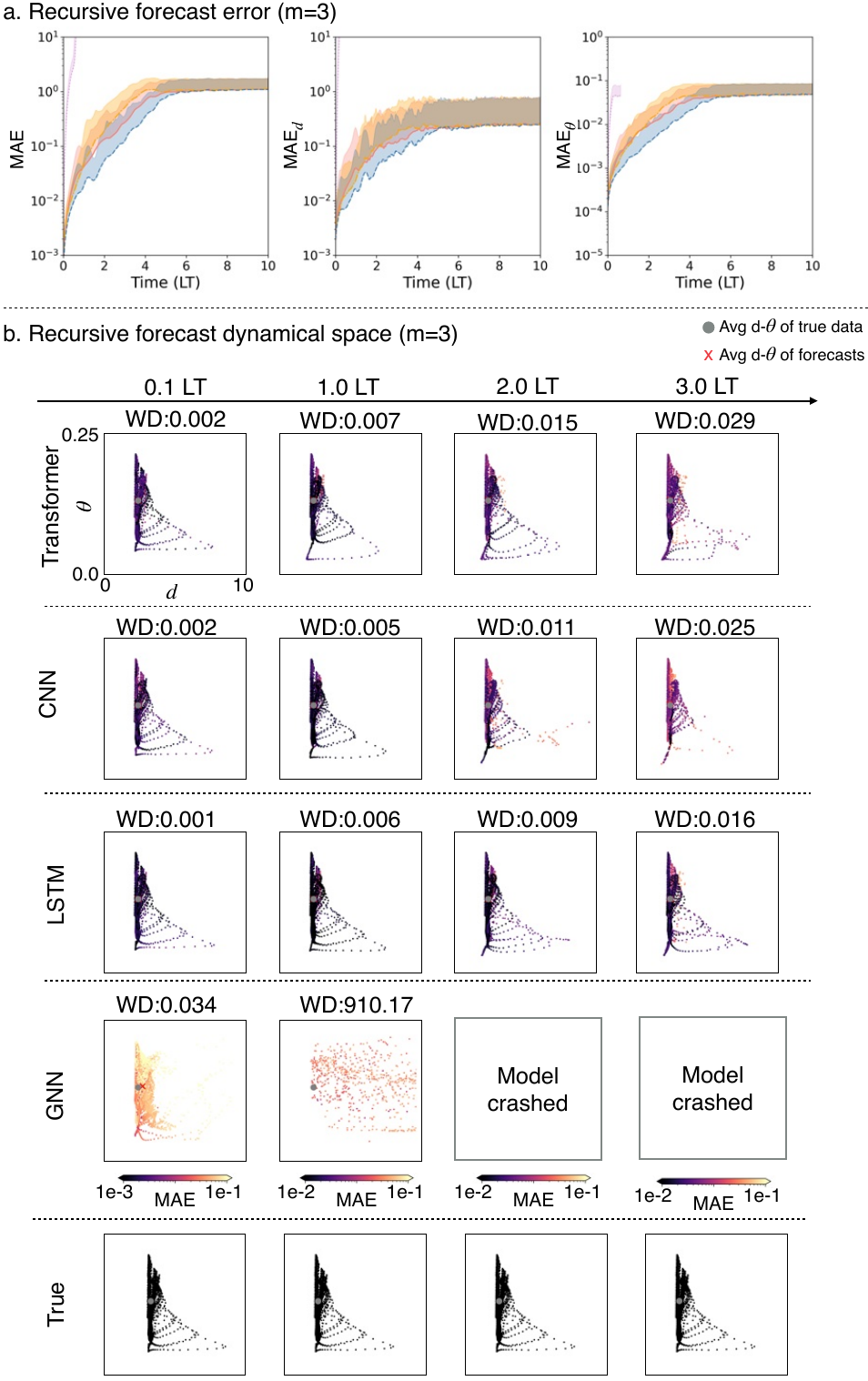}
    \caption{\textbf{MAE and dynamical space of Lorenz recursive forecast.} Panel (a): Forecast error vs recursive forecast time in terms of Lyapunov time (LT). The shaded area represents the standard deviation of forecasts starting from 5000 initial states. Panel (b): $d-\theta$ space of the 5000 trajectories at forecast time 0.1 LT, 1.0 LT, 2.0 LT and 3.0 LT. The horizontal and vertical coordinates are $d$ and $\theta$, with the mean value of indices and WD annotated on the figure. The GNN $d-\theta$ spaces for 2.0LT and 3.0LT are not plotted since the model crashed and produced NAN results.}
    \label{si-fig:Fig3_MAE}
\end{figure}

\begin{figure}[H]
    \centering
    \includegraphics[width=0.8\linewidth]{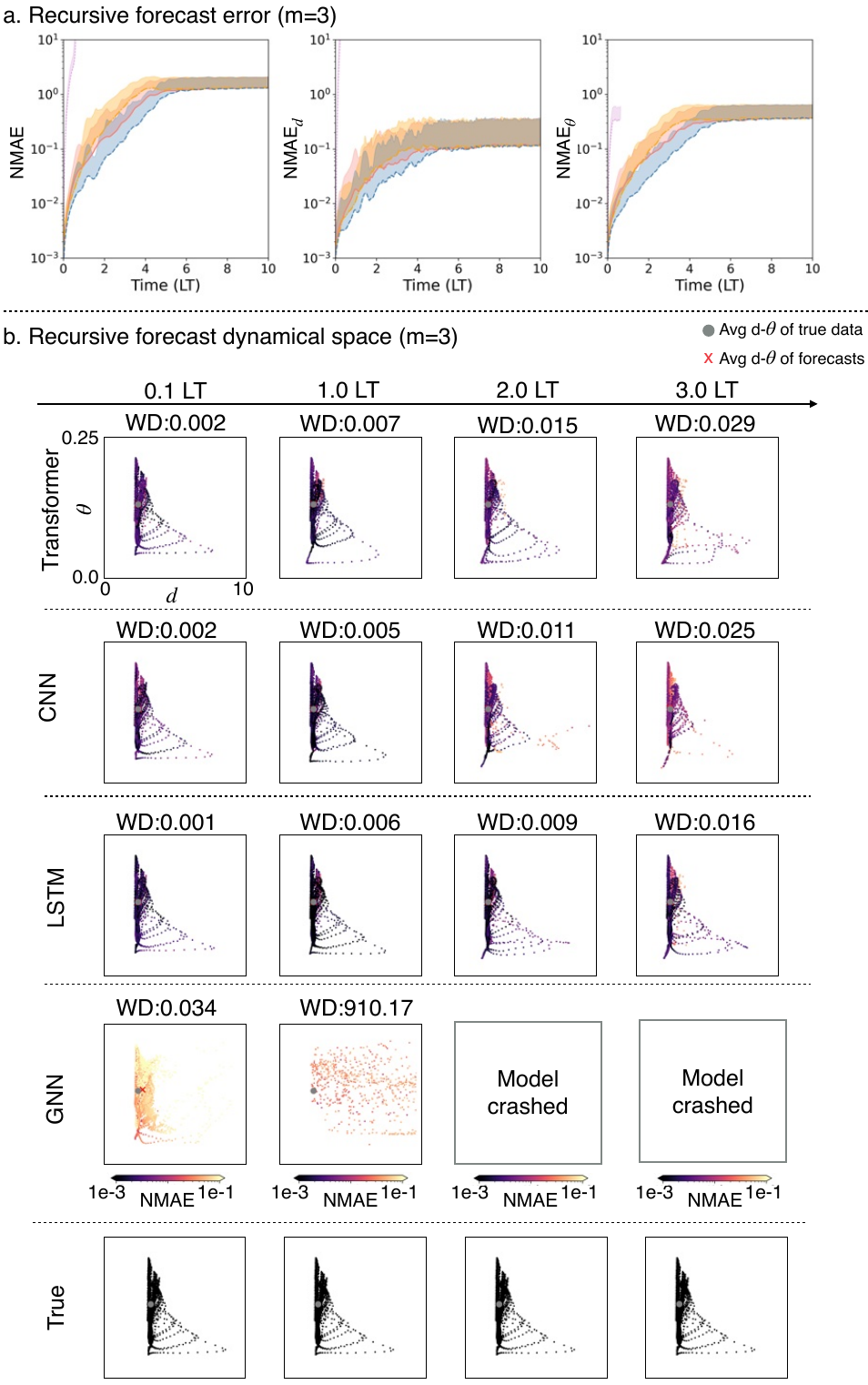}
    \caption{\textbf{NMAE and dynamical space of Lorenz recursive forecast.} Panel (a): Forecast error vs recursive forecast time in terms of Lyapunov time (LT). The shaded area represents the standard deviation of forecasts starting from 5000 initial states. Panel (b): $d-\theta$ space of the 5000 trajectories at forecast time 0.1 LT, 1.0 LT, 2.0 LT and 3.0 LT. The horizontal and vertical coordinates are $d$ and $\theta$, with the mean value of indices and WD annotated on the figure. The GNN $d-\theta$ spaces for 2.0LT and 3.0LT are not plotted since the model crashed and produced NAN results.}
    \label{si-fig:Fig3_NMAE}
\end{figure}

\begin{figure}[H]
    \centering
    \includegraphics[width=0.8\linewidth]{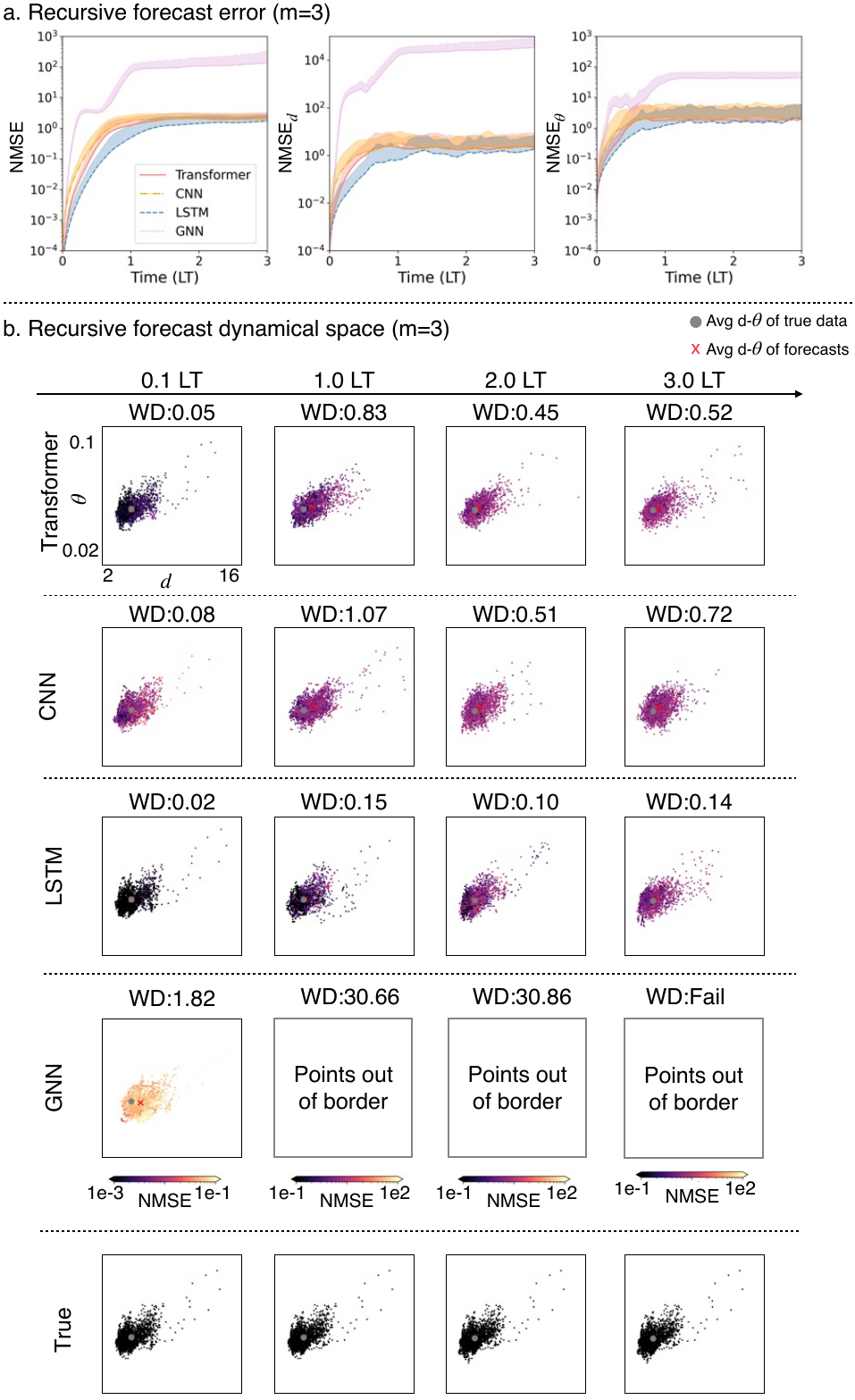}
    \caption{\textbf{NMSE and dynamical space of KS recursive forecast.} Panel (a): Forecast error vs recursive forecast time in terms of Lyapunov time (LT). The shaded area represents the standard deviation of forecasts starting from 2000 initial states. Panel (b): $d-\theta$ space of the 2000 trajectories at forecast time 0.1 LT, 1.0 LT, 2.0 LT and 3.0 LT. The horizontal and vertical coordinates are $d$ and $\theta$, with the mean value of indices and WD annotated on the figure.}
    \label{si-fig:Fig3_ks_NMSE}
\end{figure}

\begin{figure}[H]
    \centering
    \includegraphics[width=0.8\linewidth]{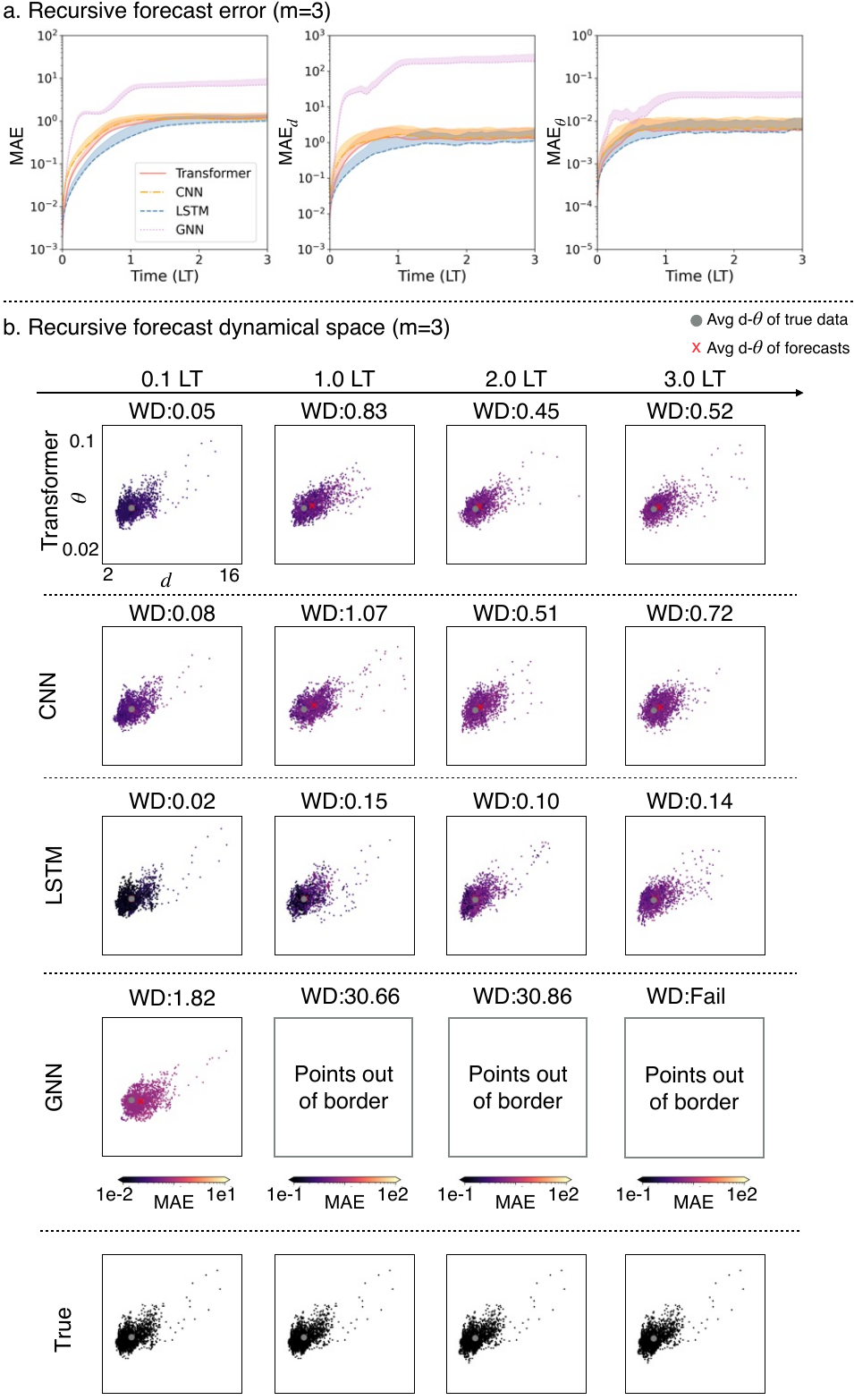}
    \caption{\textbf{MAE and dynamical space of KS recursive forecast.} Panel (a): Forecast error vs recursive forecast time in terms of Lyapunov time (LT). The shaded area represents the standard deviation of forecasts starting from 2000 initial states. Panel (b): $d-\theta$ space of the 2000 trajectories at forecast time 0.1 LT, 1.0 LT, 2.0 LT and 3.0 LT. The horizontal and vertical coordinates are $d$ and $\theta$, with the mean value of indices and WD annotated on the figure.}
    \label{si-fig:Fig3_ks_MAE}
\end{figure}

\begin{figure}[H]
    \centering
    \includegraphics[width=0.8\linewidth]{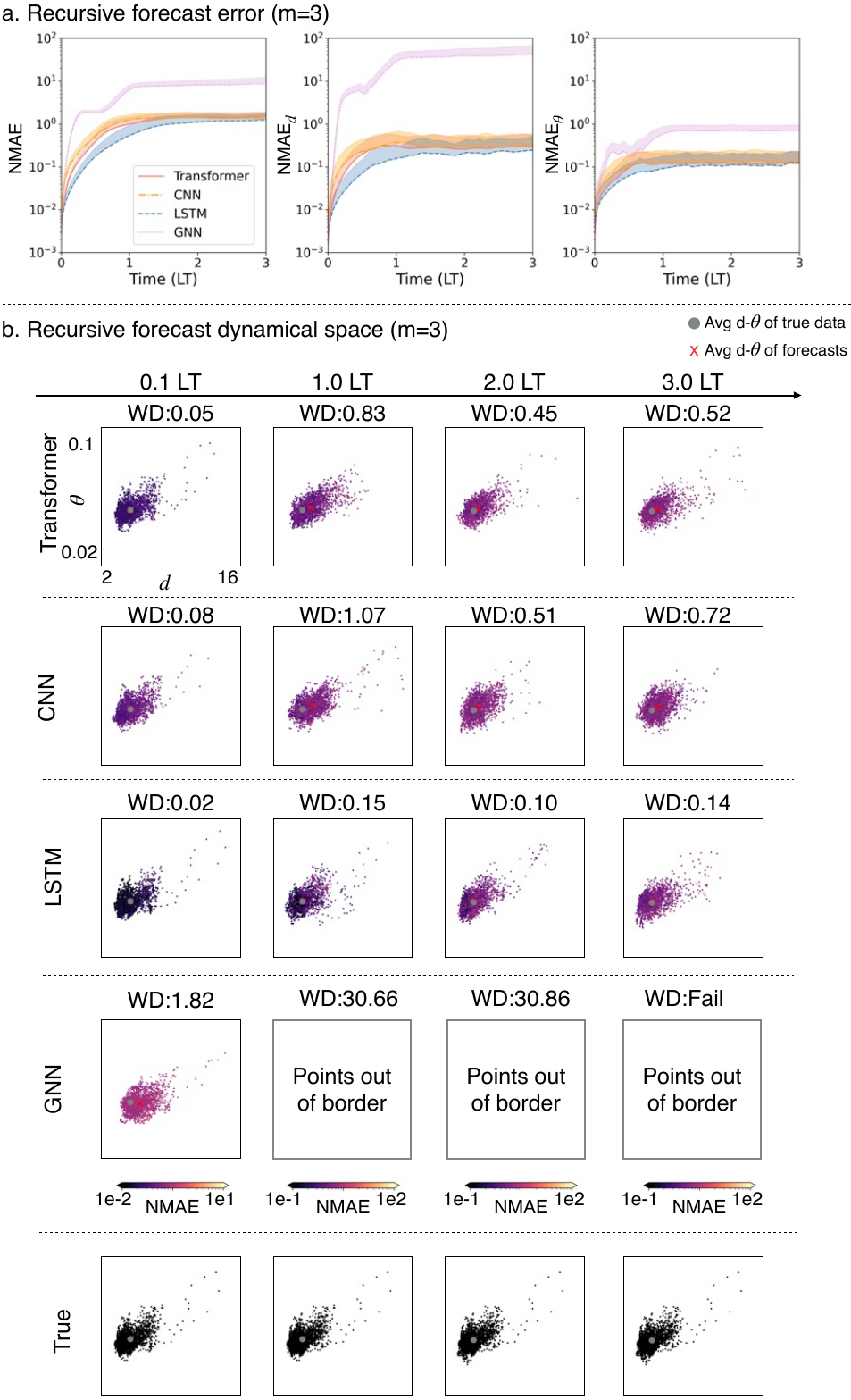}
    \caption{\textbf{NMAE and dynamical space of KS recursive forecast.} Panel (a): Forecast error vs recursive forecast time in terms of Lyapunov time (LT). The shaded area represents the standard deviation of forecasts starting from 2000 initial states. Panel (b): $d-\theta$ space of the 2000 trajectories at forecast time 0.1 LT, 1.0 LT, 2.0 LT and 3.0 LT. The horizontal and vertical coordinates are $d$ and $\theta$, with the mean value of indices and WD annotated on the figure.}
    \label{si-fig:Fig3_ks_NMAE}
\end{figure}

\begin{figure}[H]
    \centering
    \includegraphics[width=0.8\linewidth]{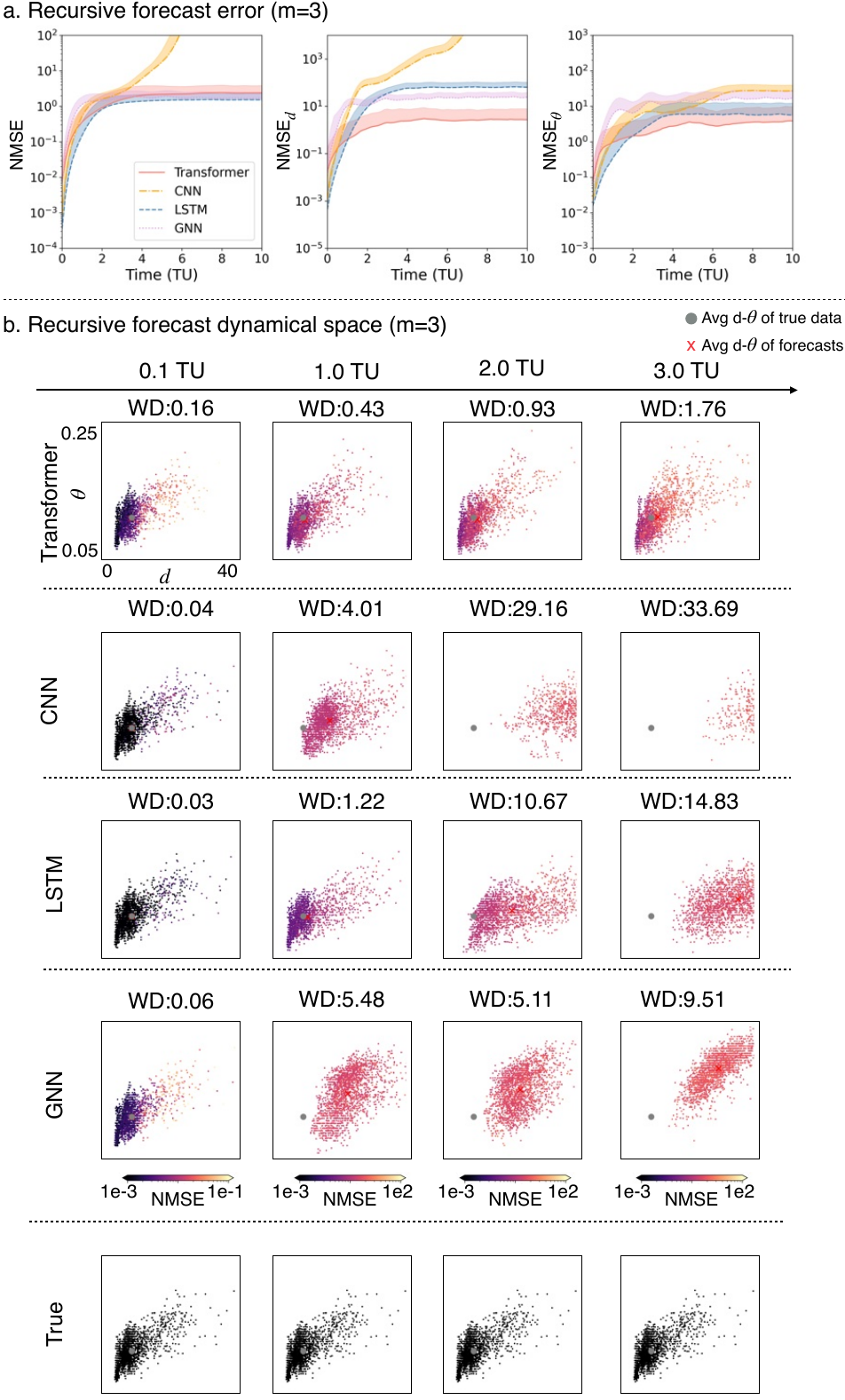}
    \caption{\textbf{NMSE and dynamical space of KF recursive forecast.} Panel (a): Forecast error vs recursive forecast time in terms of characteristic time units (TU). The shaded area represents the standard deviation of forecasts starting from 2000 initial states. Panel (b): $d-\theta$ space of the 2000 trajectories at forecast time 0.1 TU, 1.0 TU, 2.0 TU and 3.0 TU. The horizontal and vertical coordinates are $d$ and $\theta$, with the mean value of indices and WD annotated on the figure.}
    \label{si-fig:Fig3_kf_NMSE}
\end{figure}

\begin{figure}[H]
    \centering
    \includegraphics[width=0.8\linewidth]{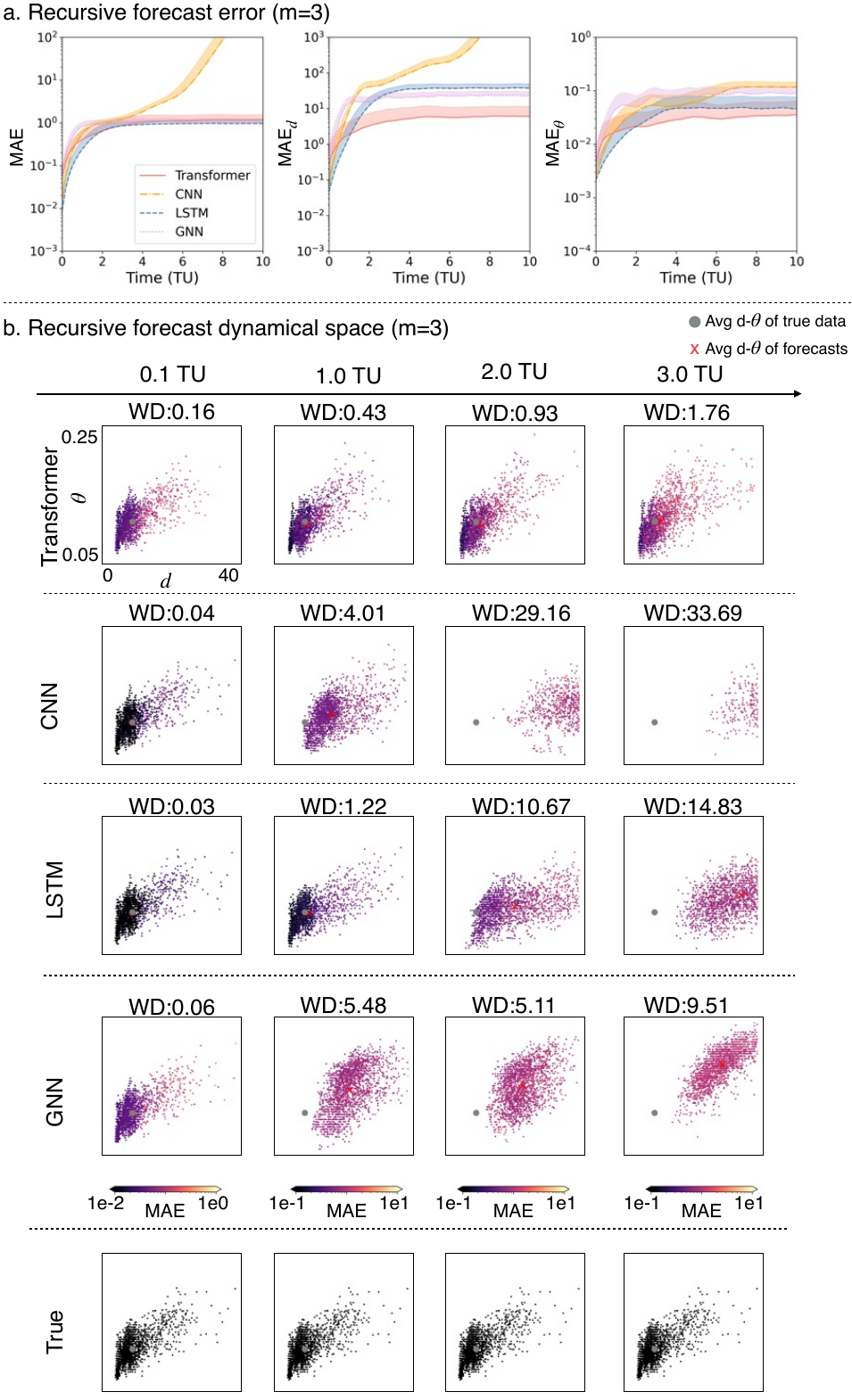}
    \caption{\textbf{MAE and dynamical space of KF recursive forecast.} Panel (a): Forecast error vs recursive forecast time in terms of characteristic time units (TU). The shaded area represents the standard deviation of forecasts starting from 2000 initial states. Panel (b): $d-\theta$ space of the 2000 trajectories at forecast time 0.1 TU, 1.0 TU, 2.0 TU and 3.0 TU. The horizontal and vertical coordinates are $d$ and $\theta$, with the mean value of indices and WD annotated on the figure.}
    \label{si-fig:Fig3_kf_MAE}
\end{figure}

\begin{figure}[H]
    \centering
    \includegraphics[width=0.8\linewidth]{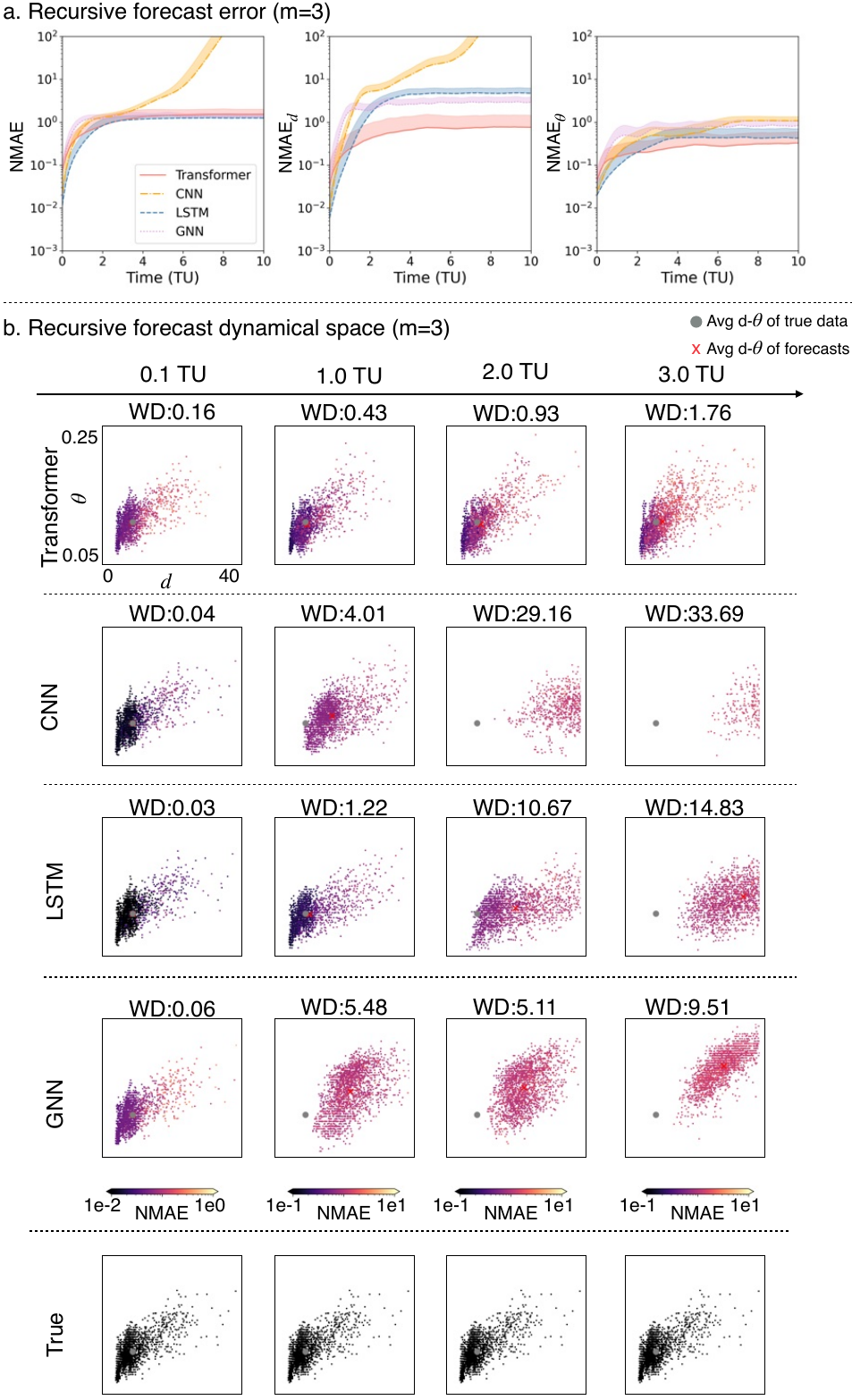}
    \caption{\textbf{NMAE and dynamical space of KF recursive forecast.} Panel (a): Forecast error vs recursive forecast time in terms of characteristic time units (TU). The shaded area represents the standard deviation of forecasts starting from 2000 initial states. Panel (b): $d-\theta$ space of the 2000 trajectories at forecast time 0.1 TU, 1.0 TU, 2.0 TU and 3.0 TU. The horizontal and vertical coordinates are $d$ and $\theta$, with the mean value of indices and WD annotated on the figure.}
    \label{si-fig:Fig3_kf_NMAE}
\end{figure}

Figs.~\ref{si-fig:Fig3_NMSE}--\ref{si-fig:Fig3_kf_NMAE} present the results of recursive forecasts for the three canonical datasets, for NMSE, MAE, and NMAE, respectively. 

\clearpage
\subsection{DID analysis of weather cyclone}
\label{si:cyclone_did}
\begin{figure}[H]
    \centering
    \includegraphics[width=0.8\linewidth]{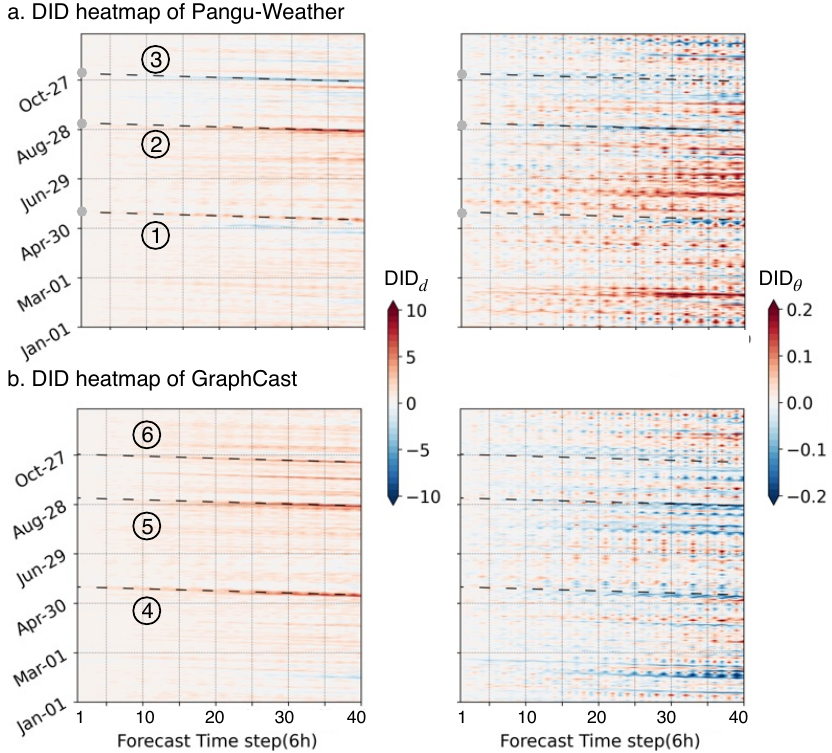}
    \caption{\textbf{DID high-error states in recursive weather forecast.} The $x$-axis in each panel represents recursive forecast step, and the $y$-axis denotes the forecast start date. Panel (a) shows the DID heat map of Pangu-Weather, and panel (b) for GraphCast.}
    \label{si-fig:cyclone_did_weather}
\end{figure}

Fig.~\ref{si-fig:cyclone_did_weather} shows the DID heatmap of the recursive weather forecast.

\clearpage
\subsection{Goodness of GPD fitting}
\begin{figure}[H]
    \centering
    \includegraphics[width=0.9\linewidth]{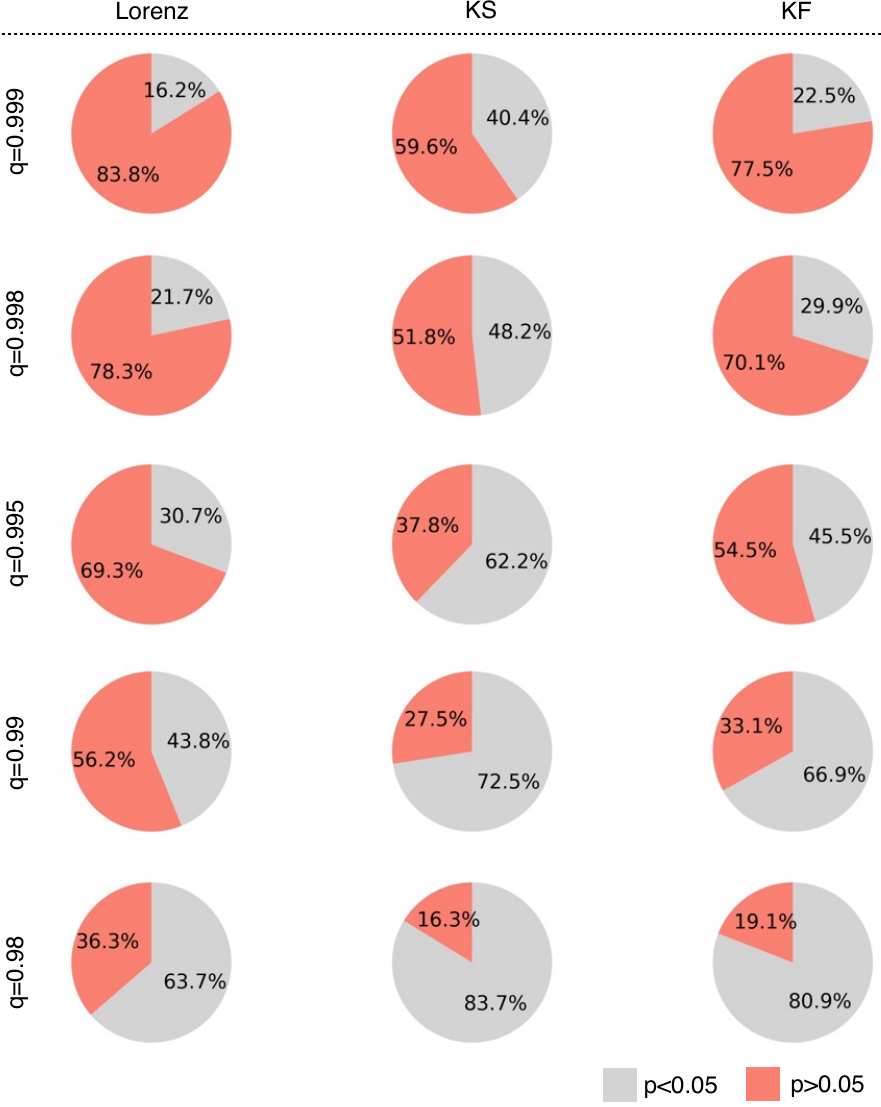}
    \caption{\textbf{Pareto distribution fitness test vs quantile $q$.} P\textgreater0.05: The null hypothesis is accepted, that the fitness is good; p\textless0.05: The null hypothesis is rejected, the data does not fit to the distribution}
    \label{si-fig:FigS11_q_sensitivity}
\end{figure}

\begin{figure}[H]
    \centering
    \includegraphics[width=0.9\linewidth]{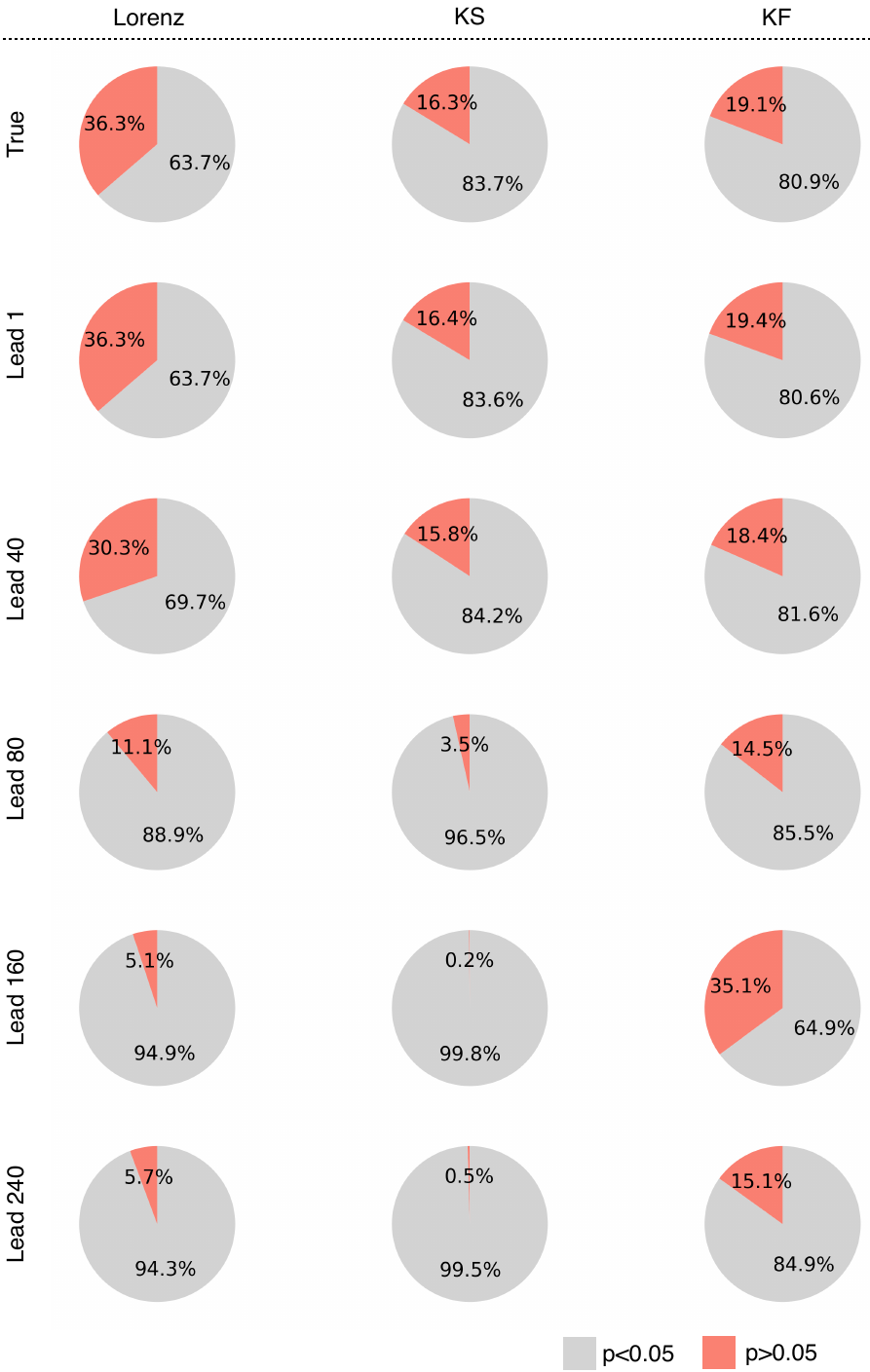}
    \caption{\textbf{Pareto distribution fitness of ML forecasts} P\textless 0.05: The null hypothesis is accepted, that the fitness is good; p\textgreater 0.05: The null hypothesis is rejected, the data does not fit to the distribution}
    \label{si-fig:FigS10_chi2}
\end{figure}

We test the goodness of GPD fitting with different value of $q$, as shown in Fig.~\ref{si-fig:FigS11_q_sensitivity}. Furthermore, the fit versus different forecast time is shown in Fig.~\ref{si-fig:FigS10_chi2}.
\label{si:fitness}

\subsection{Calculation of DI using ML output}
\label{si:method_illustration}

\begin{figure}[H]
    \centering
    \includegraphics[width=0.7\linewidth]{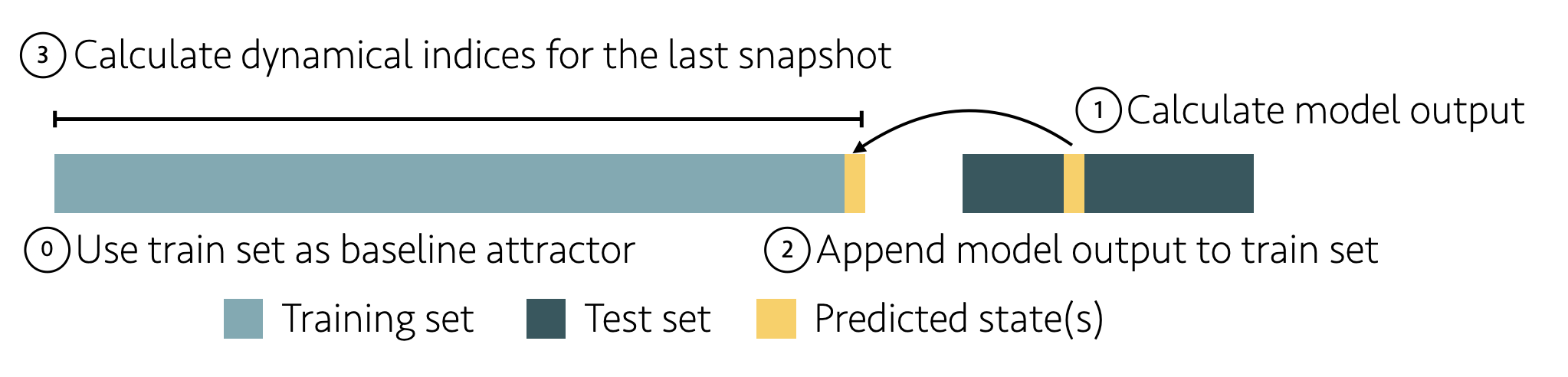}
    \caption{\textbf{Diagram of calculating dynamical indices using model output.} The training set is used as the reference attractor, and the dynamical indices for the ML forecasts are estimated based on the historical data.}
    \label{si-fig:calculate_di}
\end{figure}

\begin{figure}[H]
    \centering
    \includegraphics[width=0.5\linewidth]{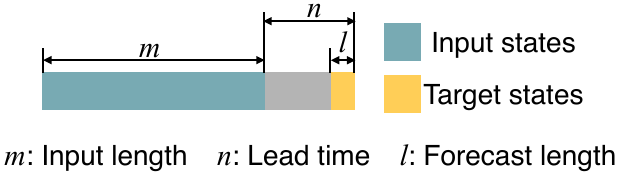}
    \caption{\textbf{General ML forecast task.} $m$: length of input feature; $n$: the forecast lead time; $l$: the length of prediction.}
    \label{si-fig:task}
\end{figure}

In Fig.~\ref{si-fig:calculate_di}, we show the workflow of calculating dynamical indices based on ML model output, as introduced in section~\ref{sec:methods-indices}. In Fig.~\ref{si-fig:task}, we depict general machine learning forecasting tasks, as introduced in section~\ref{sec:ml_task}.

\clearpage
\subsection{Dynamical indices of normalized and raw data}\label{normalization}
\begin{figure}[H]
    \centering
    \includegraphics[width=0.95\linewidth]{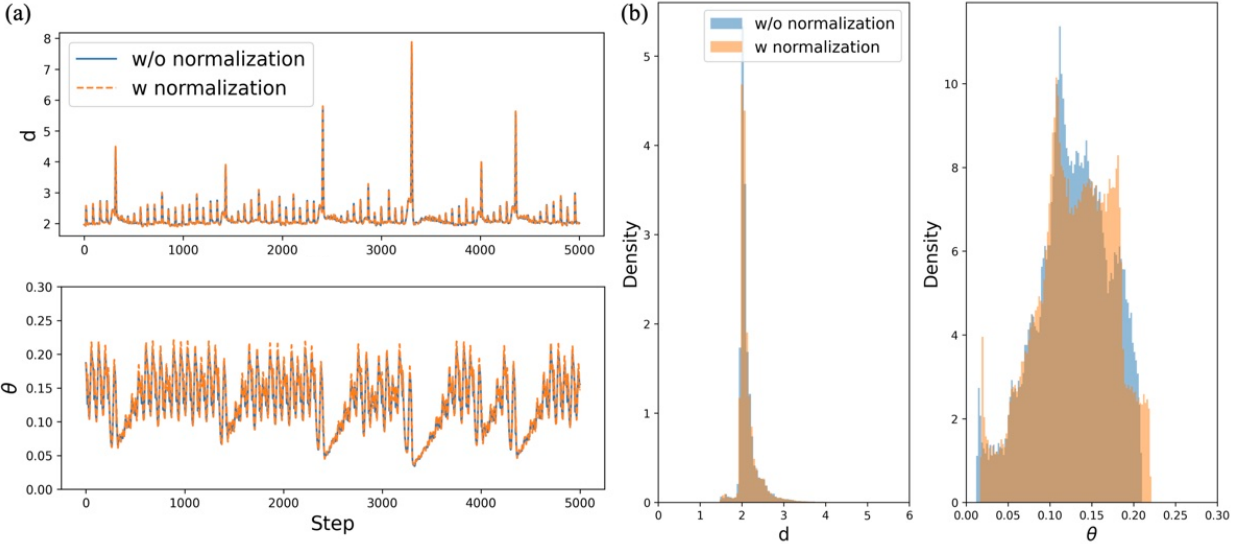}
    \caption{\textbf{Comparison between normalized and non-normalized data.} Dynamical indices time series and distribution of raw and normalized data}
    \label{si-fig:normalization}
\end{figure}

Fig.~\ref{si-fig:normalization} presents a comparison of the dynamical indices computed from both the original and normalized data on the Lorenz dataset. 
The time series of the indices display nearly identical behavior in both cases, indicating that normalization does not significantly affect their temporal evolution. 
However, differences emerge in the statistical distributions of the indices. 
For practical applications, we recommend computing dynamical indices on the original data scale, as it enables more physically meaningful interpretation of the results.

\end{document}